\documentclass[10pt,twocolumn,letterpaper]{article}

\usepackage{iccv}
\usepackage{times}
\usepackage{epsfig}
\usepackage{graphicx}
\usepackage{amsmath}
\usepackage{amssymb}
\usepackage{multirow}
\usepackage{color,soul}
\usepackage{xcolor,colortbl}
\usepackage{caption}
\usepackage{overpic}
\usepackage{booktabs}

\newcommand{\nickname}{Sketch-A-Shape\xspace}

\usepackage[breaklinks=true,bookmarks=false]{hyperref}

\iccvfinalcopy % *** Uncomment this line for the final submission

\def\iccvPaperID{12255} % *** Enter the ICCV Paper ID here

\ificcvfinal\fi

\begin{document}

%%%%%%%%% TITLE
\title{ \nickname: Zero-Shot Sketch-to-3D Shape Generation}

\author{Aditya Sanghi \qquad Pradeep Kumar Jayaraman\footnotemark[3]  \qquad Arianna Rampini\footnotemark[3] \qquad Joseph Lambourne \qquad \\ Hooman Shayani\qquad Evan Atherton \qquad 
Saeid Asgari Taghanaki \\
\text{\normalsize Autodesk Research}\\
% {\color{magenta} \url{https://ivl.cs.brown.edu/\#/projects/clip-sculptor}}
}

% \maketitle
\ificcvfinal\thispagestyle{empty}\fi

%%%%%%%%% ABSTRACT

%%%%%%%%% BODY TEXT
% \section{Introduction}
% \footnote{\footnotemark[3] Equal contribution}

\twocolumn[{%
\renewcommand\twocolumn[1][]{#1}%
\maketitle
\begin{center}
\centering
\captionsetup{type=figure}
\setlength{\tabcolsep}{0pt}
        \includegraphics[width=\textwidth]{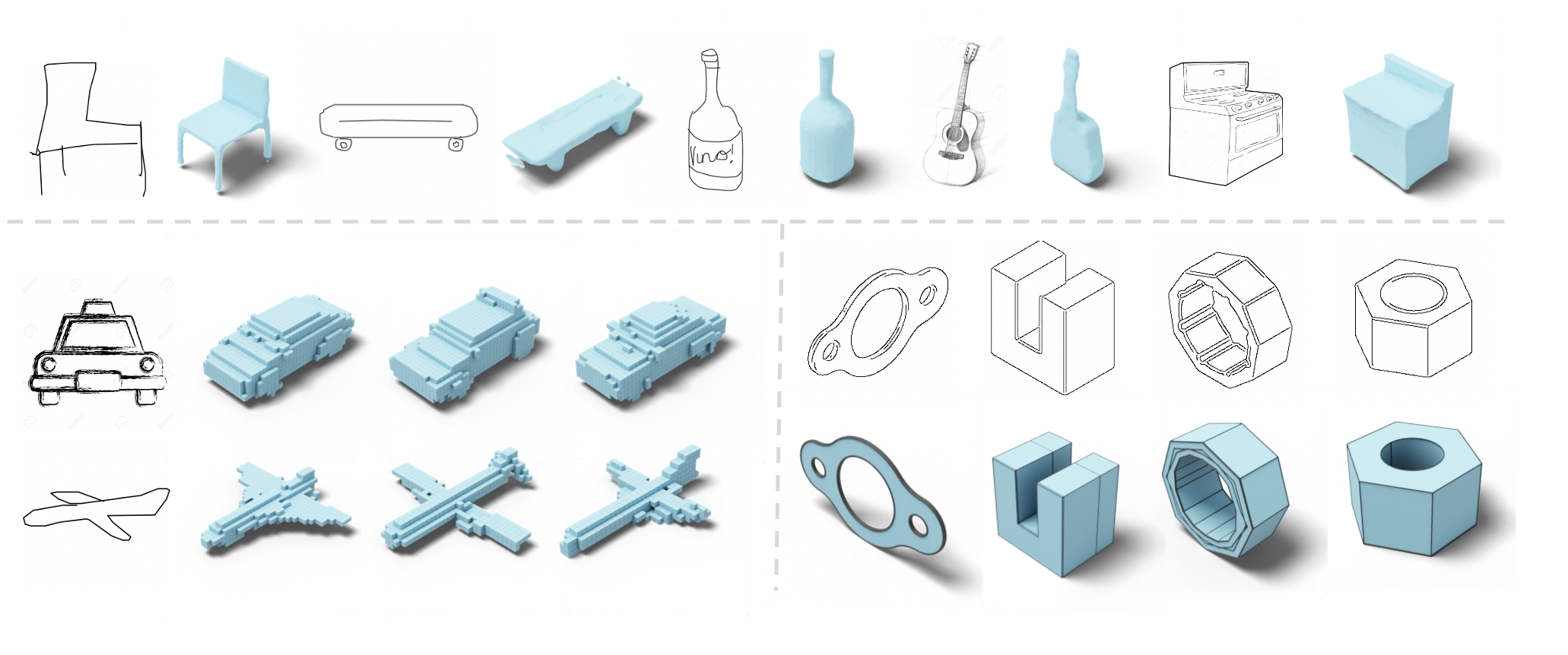}
        \caption{\nickname is a zero-shot sketch-to-3D generative model. Here we show how our method can generalize across voxel, implicit, and CAD representations and synthesize consistent 3D shapes from a variety of inputs ranging from casual doodles to professional sketches with different levels of \textit{ambiguity}.
        }
        \label{fig:teaser}
        \vspace{-0.05in}
\end{center}%
}]

\begin{abstract}
Significant progress has recently been made in creative applications of large pre-trained models for downstream tasks in 3D vision, such as text-to-shape generation. 
This motivates our investigation of how these pre-trained models can be used effectively to generate 3D shapes from sketches, which has largely remained an open challenge due to the limited sketch-shape paired datasets and the varying level of abstraction in the sketches. 
We discover that conditioning a 3D generative model on the features (obtained from a frozen large pre-trained vision model) of synthetic renderings during training enables us to effectively generate 3D shapes from sketches at inference time. 
This suggests that the large pre-trained vision model features carry semantic signals that are resilient to domain shifts, i.e., allowing us to use only RGB renderings, but generalizing to sketches at inference time. 
We conduct a comprehensive set of experiments investigating different design factors and demonstrate the effectiveness of our straightforward approach for generation of multiple 3D shapes per each input sketch regardless of their level of abstraction without requiring any paired datasets during training.   

\end{abstract}

\section{Introduction}

Throughout history, humans have used drawings and other visual representations to communicate complex ideas, concepts, and information. 

 As hand-drawn sketches have a high level of abstraction, they allow unskilled artists or even young children to convey semantic information about 3D objects \cite{ha2018a}, while providing trained professionals with a way to quickly express important geometric and stylistic details of a 3D design.
 The ability to create 3D models which can capture the essence of simple doodles while accurately reproducing 3D shapes described by concept design sketches, will make 3D modelling more accessible to the general public, while allowing designers to rapidly explore many different design ideas and create virtual models that more accurately reflect the shape, size, and characteristics of real-world objects and environments. 

Previous studies have endeavored to employ deep learning techniques in generating 3D shapes from sketches \cite{lun20173d, guillard2021sketch2mesh, gao2022sketchsampler, kong2022diffusion}, yet there are several limitations that hinder their widespread application. 
Firstly, there is a lack of (Sketch, 3D shape) paired data at a large scale which forces most methods to be trained on synthetic datasets or data collected on only few categories. 
Even when a small number of categories of paired sketch-shape data has been collected  \cite{kong2022diffusion}, current methods fail to generalize to different levels of abstractions in the sketches, ranging from casual doodles to detailed professional drawings. 
Finally, most of the present methods incorporate strong inductive biases, such as view information \cite{zhang2021sketch2model},  differentiable rendering \cite{guillard2021sketch2mesh} and depth estimation \cite{lun20173d, gao2022sketchsampler}, thereby constraining their generalizability across 3D representations.

To overcome the challenge of limited availability of paired data, a potential solution is to use prior knowledge encoded in large pre-trained image-text models. Recently, these large pre-trained models have been successfully applied to the 3D domain in creative ways, such as guiding the optimization of differentiable 3D representations \cite{jain2022zero, poole2022dreamfusion, lin2022magic3d} or to generate 3D shapes from text prompts using interchangeability of text-image embeddings \cite{sanghi2022clip,sanghi2022textcraft}, or using them for representation learning \cite{zhang2022pointclip, Schlachter2022}.

In this paper, we introduce a straightforward yet effective approach called \nickname, for generating 3D shapes from sketches in a zero-shot setting using pre-trained vision models. Our method is based on the idea that 3D shape rendering features derived from large-scale pre-trained models (such as CLIP \cite{radford2021learning} and DINOv2 \cite{oquab2023dinov2}) possess robust local semantic signals that can withstand domain shifts from renderings to sketches. 
In \nickname,  we first train a VQ-VAE  to acquire shape embeddings. Following this, a masked transformer is trained to model the distribution of shape embeddings conditioned on local semantic features from an image encoder that is pre-trained and frozen. During inference, the masked transformer is conditioned on local semantic features of the sketch instead, in order to produce the 3D shape. Our findings suggest that with some architectural design choices, this straightforward method enables us to generate several 3D shapes that can generalize across sketches of varying complexities.

To sum up, we make the following contributions:
\begin{itemize}
    \item We propose \nickname, the first zero-shot approach for sketch-to-3D generation, leveraging large-scale pre-trained models to outdo the need of paired sketch-3D dataset.
    \item We experimentally show the generalization capability of our method among various datasets (\autoref{sec:experiments}) with different levels of sketch abstraction, going from simple doodles to professional sketches.
    \item We conduct thorough experiments to examine the different components of \nickname that contribute to the success of zero-shot shape generation via sketch.
\end{itemize}

\section{Related Work}
\label{sec:related}

\noindent \textbf{3D Generative Models.}  
Significant progress has been made in the field of generative models for the creation of 3D shapes in various formats such as voxels \cite{3dgan, choy20163d, tatarchenko2017octree, jimenez2016unsupervised}, CAD \cite{wu2021deepcad, jayaraman2022solidgen, lambourne2022reconstructing, xu2022skexgen}, implicit representations \cite{OccupancyNetworks2019, chen2019learning, Park_2019_CVPR}, meshes \cite{nash2020polygen, groueix2018papier}, and point clouds \cite{nash2017shape, achlioptas2018learning, yang2018foldingnet,li2018point, yang2019pointflow, Ko2021RPGLR}. Recent research on 3D generative models has focused primarily on the development of generative models based on VQ-VAE \cite{mittal2022autosdf, cheng2022autoregressive, yan2022shapeformer, zhang20223dilg, sanghi2022textcraft, xu2022skexgen}, GAN\cite{ chan2022efficient, schwarz2020graf, gao2022get3d}, or diffusion models \cite{zhou20213d, luo2021diffusion, nichol2022point, hui2022neural}. 
The present study concentrates on connecting the sketch modality with 3D shapes across three different 3D representations: voxels, CAD, and implicit representation. Although our approach is based on VQ-VAE, it can be easily extended to GAN or diffusion-based generative models. 

\noindent \textbf{3D Zero-Shot Learning.} Large pre-trained language and 2D vision models have been creatively used in several downstream 3D vision tasks. Initial works focused on using vision-text models such as CLIP \cite{radford2021learning} for 3D shape generation using text \cite{sanghi2022clip},  optimizing nerfs \cite{jain2022zero}, deforming meshes \cite{michel2022text2mesh}, stylizing meshes \cite{mishra2022clip} and animating avatars \cite{hong2022avatarclip} . More recently, text-to-image models such as Stable Diffusion \cite{rombach2022high} and Imagen \cite{saharia2022photorealistic}, have been used for text-to-shape generation \cite{poole2022dreamfusion, lin2022magic3d}, single-view reconstruction \cite{ye2021shelf}, and adding texture to 3D shapes \cite{metzer2022latent}. To the best of our knowledge, our work is the first to explore zero-shot 3D shape generation from sketches by leveraging a pre-trained model.

\noindent \textbf{3D Shape Generation from Sketch.}
Several supervised learning methods have been  used to generate 3D shapes from sketches. Works such as \cite{lun20173d} use a neural net to estimate depth and normals from a set of viewpoints for a given sketch, which are then integrated into a 3D point cloud. \cite{delanoy20183d} proposes to use a CNN to predict the initial shape and then refine the shape using novel viewpoints using another neural network.  Another work \cite{jin2020contour} represent the 3D shape and its occluding contours in a joint VAE latent space during training which enables them to retrieve a sketch during inference and generate a 3D shape. Sketch2Mesh \cite{guillard2021sketch2mesh} uses an encoder-decoder architecture to represent and refine a 3D shape to match the target external contour using a differentiable render. Methods such as \cite{wang2018unsupervised, zhang2021sketch2model} employ a domain adaption network between unpaired sketch and rendered image data to boost performance on abstract hand-drawn sketches. To address the ambiguity problem of sketches, \cite{zhang2021sketch2model} introduces an additional encoder-decoder to extract separate view and shape sketch features, while \cite{gao2022sketchsampler} proposes a sketch translator module to fully exploit the spatial information in a sketch and generate suitable features for 3D shape prediction.
Recently, \cite{kong2022diffusion} trains a diffusion model for generation of 3D point clouds conditioned on sketches using a multi-stage training, and fine-tuning technique. However, we take the novel approach of not training on paired shape-sketch data at all and instead rely on the robustness of the local semantic features from a frozen large pre-trained image encoder such as CLIP.

\begin{figure*}
\centering
\includegraphics[width=0.99\textwidth]{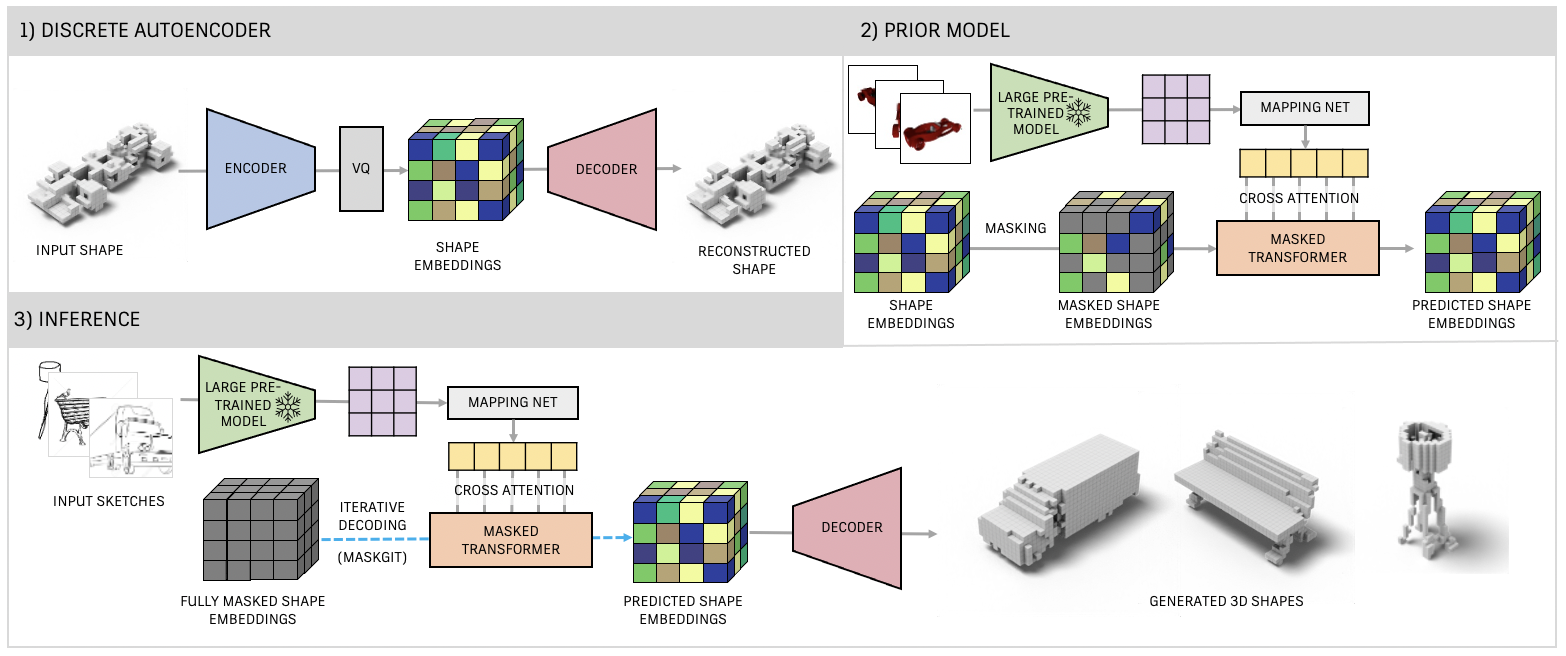}

\caption{An overview of the \nickname method. The top row shows the two training stages. The bottom row shows how 3D shape is generated via sketch at inference time.}
\label{fig:arch}
\end{figure*}

\section{Method}\label{sec:method}

Our approach strives to generate 3D shapes from sketches of different complexities, solely employing synthetic renderings, and without the requirement of a paired dataset of sketches and shapes. The training data consists of 3D shapes, each denoted by $\mathbf{S}$, which can be represented as a voxel grid, implicit (e.g. occupancy), or CAD, and their $\mathbf{r}$ multi-view renderings ($\mathbf{I}^{1:r}$).
% collectively represented as $\mathcal{D} = \{(\mathbf{I}^{1:r}_n, \mathbf{  S}_n )\}_{n=1}^{N}$. 
Our approach involves two training stages. In the first stage, the shapes are transformed into discrete sequences of indices (shape embeddings), denoted by $\mathbf{Z}$, using a discrete auto-encoder framework \cite{oord2017neural}. In the second stage, the distribution of these indices is modeled using a transformer-based generative model that is conditioned on features of shape renderings obtained from a frozen pre-trained model. These shape rendering features are a grid of local features from the frozen pre-trained model  which are converted into a sequence of local features then conditioned to the transformer through a cross-attention mechanism.
During inference, we use an iterative decoding scheme \cite{chang2022maskgit} to generate the shape indices iteratively based on features of the sketch. Once the shape indices are generated, we can use the decoder of the auto-encoder to generate the 3D shape. The overall process is illustrated in Figure \ref{fig:arch} .

\subsection{Stage 1: Training Discrete Auto-encoder}
In the first stage, we use an auto-encoder framework to capture the shape distribution into a compressed sequence of discrete indices (shape embeddings) among various modalities. To achieve this, we opt for the Vector Quantized Variational Auto-encoder (VQ-VAE) architecture \cite{oord2017neural} which efficiently models the 3D shape in a compressed latent space, circumventing posterior collapse and enabling the generation of high-quality 3D shapes. The dataset of 3D shapes $\mathbf{S}$, are transformed using an encoder, $E(.)$, into a sequence of discrete indices $\mathbf{Z}$, pointing to a shape dictionary, which their distributions are then modeled in stage 2 using a transformer-based generative model. This is shown below:
\begin{align}
\begin{split}
\mathbf{Z} = VQ(E(\mathbf{S})), \quad \mathbf{S}^{'} = D(\mathbf{Z})
\end{split}
\label{equa:noise_reg}
\end{align}

 The shape $\mathbf{S}^{'}$ is then generated from $\mathbf{Z}$ using a decoder, $D(.)$, with a reconstruction loss $L_{rec}(S, S^\prime)$. We also use the commitment loss \cite{oord2017neural} to encourage encoder output $E(.)$ commits to
an embedding in the shape dictionary, and exponential moving average strategy \cite{razavi2019generating,robust_ema} to encourage dictionary enteries to gradually be pulled toward the encoded features.
 When dealing with voxel representation, we leverage a 3D convolution based on the ResNet architecture \cite{he2016deep} for both the encoder and decoder network. Whereas with implicit representation, we rely on a ResNet-based encoder and an up-sampling process for the decoder that generates a higher resolution volume, which is then queried locally to obtain the final occupancy \cite{OccupancyNetworks2019, peng2020convolutional}. In the case of CAD representation, we use the SkexGen VQ-VAE architecture \cite{xu2022skexgen}. More details of the architectures are provided in the supplementary material.

\subsection{Stage 2: Masked Transformer}
The goal of stage 2 is to train a prior model which can effectively generate shape indices conditioned on a sketch at  inference time. 
We achieve this by modelling the sequence of discrete indices (shape embedding $\mathbf{Z}$), produced from stage 1, using a conditional generative model. We use a bi-directional transformer \cite{chang2022maskgit} based network which is conditioned on the features of the synthetic 3D renderings using a cross-attention mechanism. 
During training, we randomly mask a fraction of shape indices with a special mask token \cite{chang2022maskgit} to produce $\mathbf{Z}^{msk}$. The training objective then becomes how to fully unmask the masked indices using the help of the provided conditional information. The training objective is to minimize:
\begin{equation}
 L_{mask}= - \mathop{\mathbb{E}}_{Z,C \in D} [\log p(\mathbf{Z}|\mathbf{Z}^{msk}, \mathbf{C})] 
\end{equation}

Here,  $\mathbf{C}$ represents the conditional information from a given shape $\mathbf{S}$ which are obtained from the multi-view image renderings of the 3D shape. At each iteration, we randomly sample a view to render an image of the 3D shape, which is then converted to local features using a locked pre-trained model. 
The choice of pre-trained model is an important design criteria which we investigate thoroughly in Section \ref{pre-trained_sec}, and find that using large models trained on diverse data produces the most robust semantic local features which allow domain shift from synthetic  renderings to sketches.   

The local features sequence can be obtained from different parts of the pre-trained network, which we investigate in Section \ref{layer_sec}. Our findings indicate that utilizing the feature grid output of the deeper layers in the pre-trained models yields better results. This is because deeper layers generate more semantic features, and the grid structure of the feature preserves its local properties. We convert this grid into a sequence using a mapping network comprising of several MLP layers. Finally, we take features obtained and add  learnable positional encoding before applying cross-attention with the shape indices' features at each transformer layer. The choice of conditioning is also an important design choice which we discuss in Section \ref{cond_sec}. Additionally, we replace the local features sequence with a null embedding sequence 5\% of the time to allow for classifier-free guidance during inference.

\subsection{Inference} 
During the generation phase, we first convert the sketch into a sequence of local features using the same frozen pre-trained model utilized during training. These local features are semantically robust and serve as the conditioning query for the transformer. We employ an iterative decoding scheme with a cosine schedule, similar to the one proposed in Mask-GIT \cite{chang2022maskgit}. The process begins with a completely masked set of indices, which are gradually unmasked in parallel using the conditional information provided by the local features sequence from the sketch. At each time step, the transformer predicts the complete unmasked shape sequence, of which a specific fraction of the highest confidence masked tokens are accepted. These selected tokens are designated as unmasked for the remaining steps, while the rest of the tokens are reset to masked, except for the already unmasked tokens from the previous steps. For each time step, we also apply classifier-free guidance \cite{ho2022classifier} with a guidance scale of 3. This process continues until all the tokens are unmasked. Finally, the completely unmasked tokens are converted into the 3D object using the shape decoder trained in stage 1. It is worth noting that we can restart the same process multiple times  to generate different 3D shapes for the same sketch query.

\section{Experiments}\label{sec:experiments}
In this section, we present the results of our experiments evaluating the accuracy and fidelity of the generated output produced by our model.  We conducted each experiment three times for each metric and reported the average result for each.  The experimental setup details are provided in the supplementary material with additional results that may be of interest.

\noindent \textbf{Training Dataset.} 
Our experimentation utilizes two subsets of the ShapeNet(v2) dataset \cite{chang2015shapenet}. The first subset, ShapeNet13, consists of 13 categories from ShapeNet, which were also employed in previous studies \cite{choy20163d,OccupancyNetworks2019}. In line with Sketch2Model  \cite{zhang2021sketch2model}, we adopt the same train/val/test partition. The second subset, ShapeNet55, includes all 55 categories of ShapeNet and we follow the same split as \cite{sanghi2022textcraft}.  We use the DeepCAD \cite{wu2021deepcad} dataset to train our CAD model. 
\vspace{1pt}

\noindent \textbf{Evaluation Sketch Dataset.}  One advantage of our method is that it's not trained on paired (shape, sketch) datasets. Therefore, to comprehensively evaluate its performance, we tested it on various sketch datasets that range from professional to non-expert sketches. Specifically, we utilized the ShapeNet-Sketch dataset \cite{zhang2021sketch2model}, which comprises 1300 free-hand sketches across ShapeNet13.  In addition, we employed the ImageNet-Sketch dataset \cite{wang2019learning}, which contains 50 sketch images for 1000 ImageNet classes obtained from Google, encompassing a range of professional to non-expert sketches. Moreover, we utilized the TU-Berlin Sketch dataset \cite{eitz2012hdhso}, which includes 20,000 non-expert sketches of 250 object categories. Lastly, QuickDraw Dataset \cite{quickdraw-data, ha2018a} is a collection of $50$ million drawings across 345 categories, contributed by players of the game \emph{Quick, Draw!} \cite{quickdraw-game}.
ImageNet-Sketch, TU-Berlin Sketch, and QuickDraw datasets also lack ground-truth 3D models, and we only utilized the categories of ShapeNet for these datasets. To evaluate our CAD model we use synthetic edge map sketches but don't train the model using  edge maps as augmentation.

\vspace{3pt}
\noindent \textbf{Evaluation Metrics.} To evaluate our method on different sketch datasets we use two metrics: classification accuracy and human evaluation which are outlined below.

\begin{enumerate}
\itemsep0em 
    
    \item \textbf{Classifier Accuracy.} As we are dealing with sketch data that lacks ground-truth 3D models, we use the Accuracy (Acc) metric to ensure that the generated shape for a given sketch corresponds to its category. To achieve this, we employ a pre-trained shape classifier, as implemented in \cite{sanghi2022clip, sanghi2022textcraft}. We use this metric for all datasets: ImageNet-Sketch \cite{wang2019learning}, TU-Berlin \cite{eitz2012hdhso}, ShapeNet-Sketch \cite{zhang2021sketch2model}, and QuickDraw \cite{quickdraw-data}.  We refer to this metric as IS-Acc, TU-Acc, SS-Acc, and QD-Acc, respectively. 
    %for the ImageNet-Sketch \cite{wang2019learning} and TU-Berlin \cite{eitz2012hdhso} datasets, respectively. 
    As our method can generate multiple shape per sketch query, we report the mean across 5 sampled shapes for a given sketch query.

    \item \textbf{Human Perceptual Evaluation.} We also use Amazon SageMaker Ground Truth and crowd workers from the Mechanical Turk workforce ~\cite{mishra_2019} to evaluate how well our generated 3D models preserve important geometric and stylistic details from the sketches.

\end{enumerate}

\subsection{Qualitative Results}

\begin{figure*}
\centering
\includegraphics[width=0.24\textwidth]{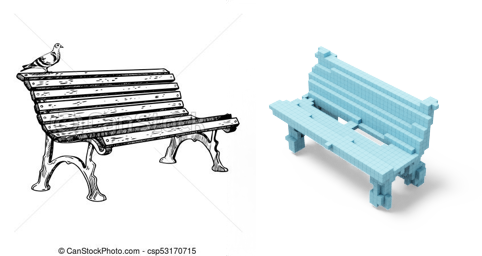}
\includegraphics[width=0.24\textwidth]{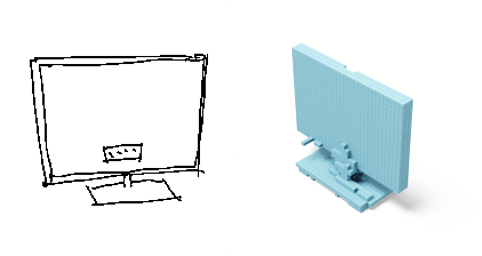}
\includegraphics[width=0.24\textwidth]{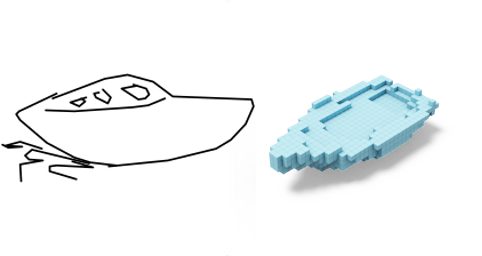}
\includegraphics[width=0.24\textwidth]{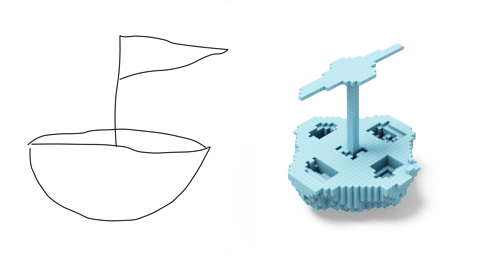}\\
\includegraphics[width=0.24\textwidth]{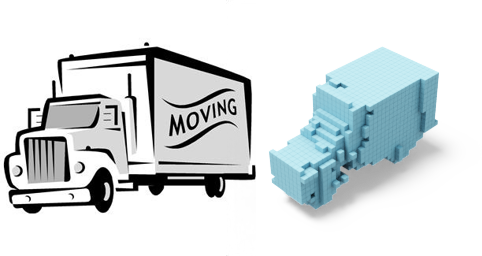}
\includegraphics[width=0.24\textwidth]{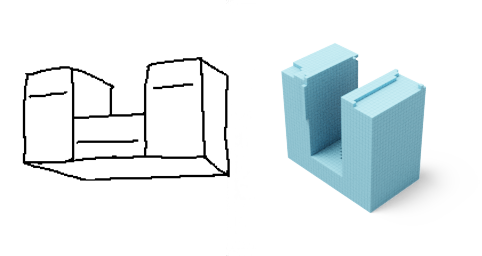}
\includegraphics[width=0.24\textwidth]{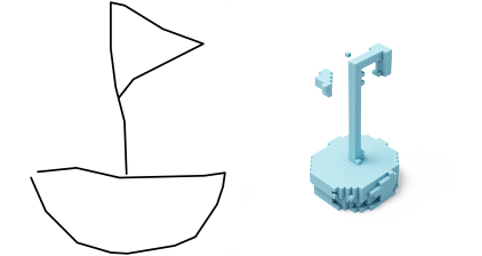}
\includegraphics[width=0.24\textwidth]{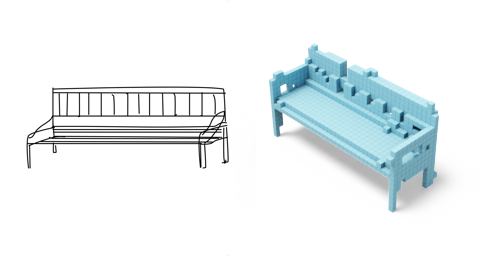}\\
\includegraphics[width=0.24\textwidth]{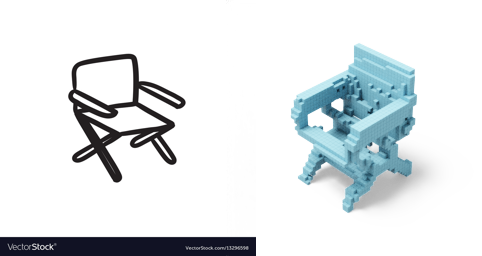}
\includegraphics[width=0.24\textwidth]{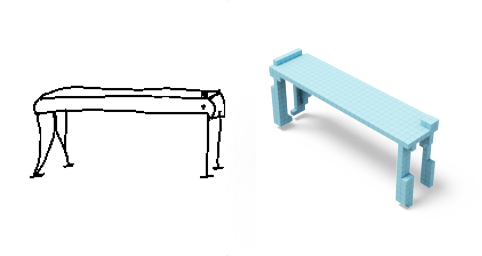}
\includegraphics[width=0.24\textwidth]{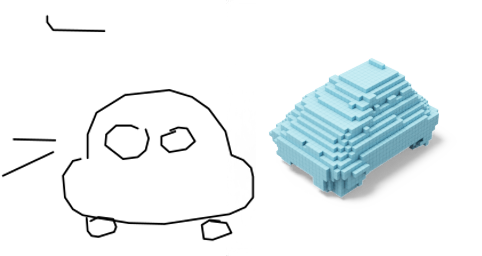}
\includegraphics[width=0.24\textwidth]{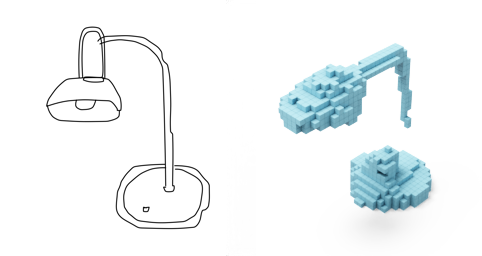}\\
%% Remove
\includegraphics[width=0.24\textwidth]{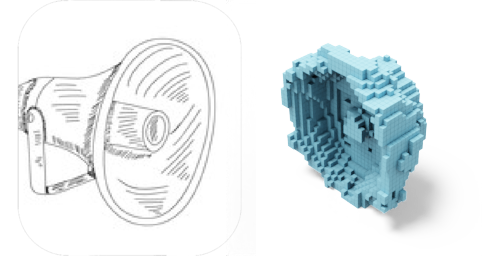}
\includegraphics[width=0.24\textwidth]{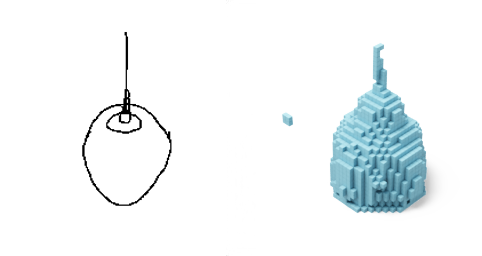}
\includegraphics[width=0.24\textwidth]{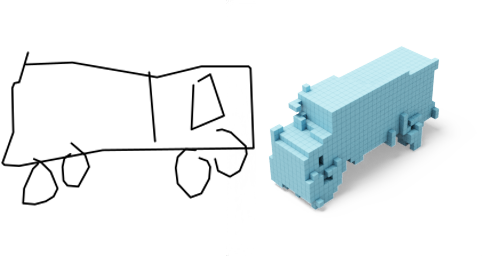}
\includegraphics[width=0.24\textwidth]{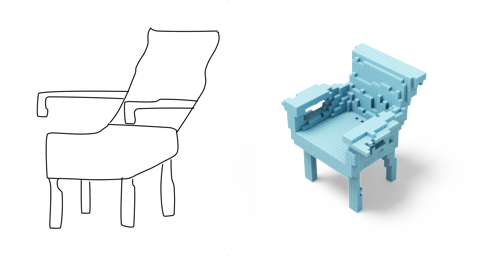}\\
\includegraphics[width=0.24\textwidth]{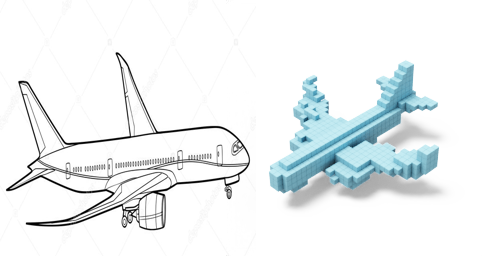}
\includegraphics[width=0.24\textwidth]{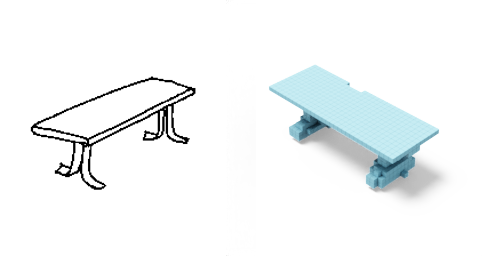}
\includegraphics[width=0.24\textwidth]{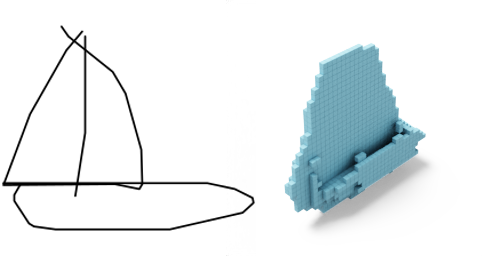}
\includegraphics[width=0.24\textwidth]{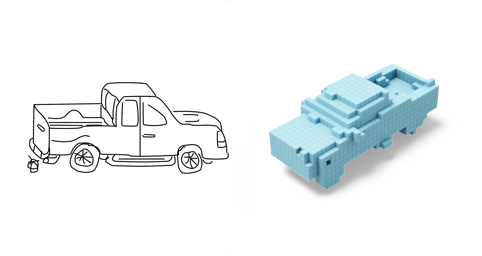}\\
\includegraphics[width=0.24\textwidth]{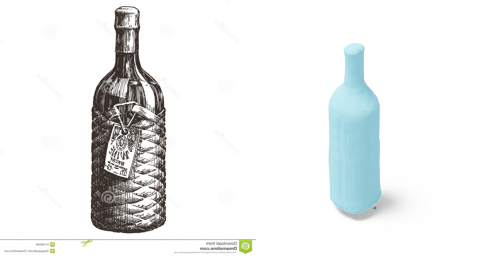}
\includegraphics[width=0.24\textwidth]{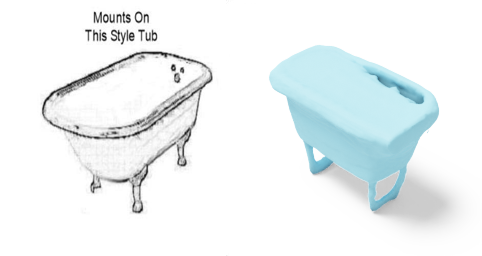}
\includegraphics[width=0.24\textwidth]{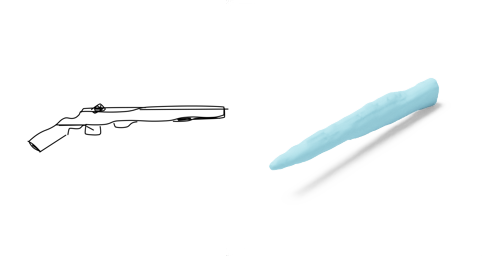}
\includegraphics[width=0.24\textwidth]{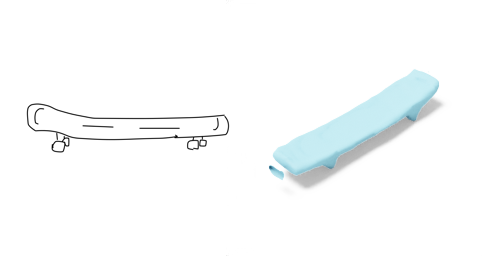}
%%
% {\color{lightgray}\hrule}
\includegraphics[width=0.49\textwidth]{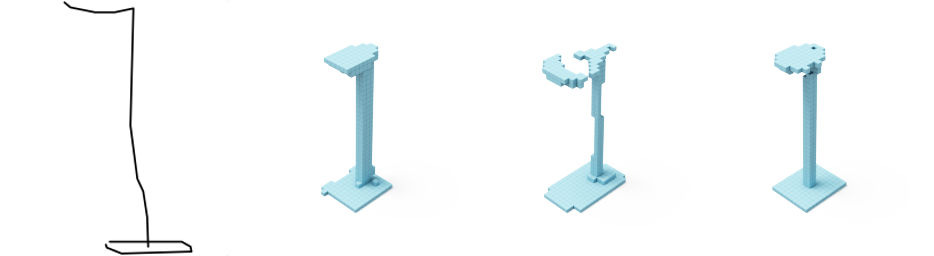}
\includegraphics[width=0.49\textwidth]{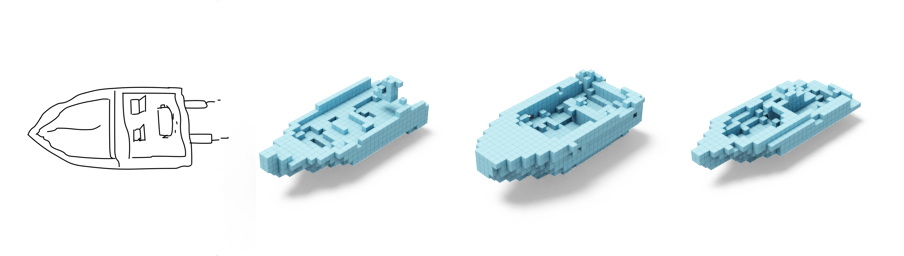}\\
\includegraphics[width=0.49\textwidth]{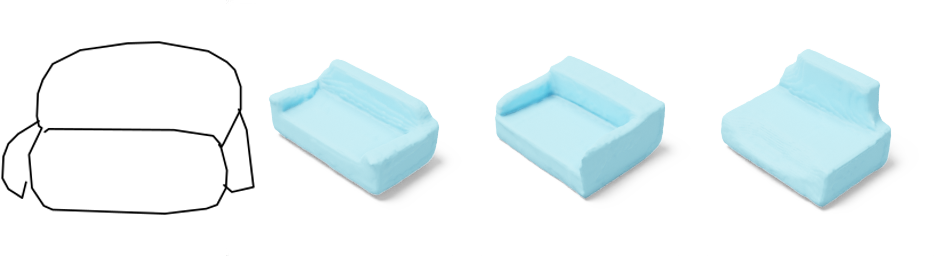}
\includegraphics[width=0.49\textwidth]{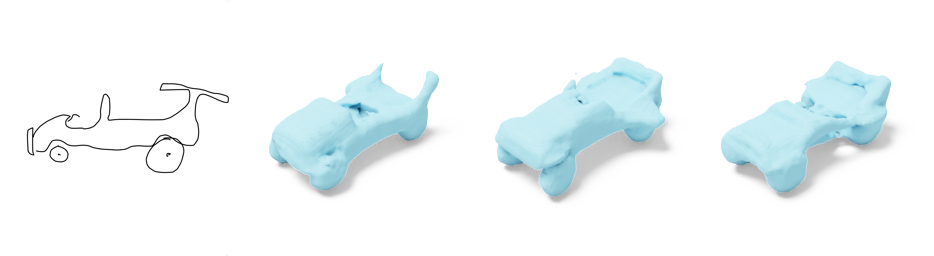}\\

\includegraphics[width=0.49\textwidth]{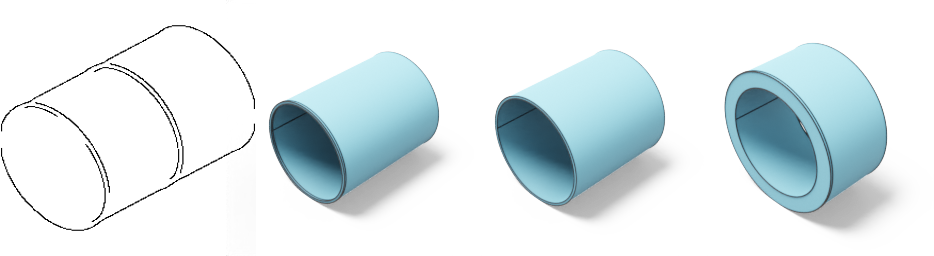}
\includegraphics[width=0.49\textwidth]{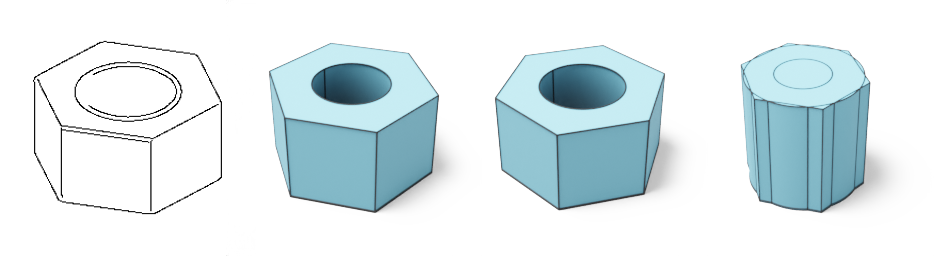}\\
\caption{Generated 3D shapes from sketches of different domains. Rows 1--5, and 6 refer to voxel, and implicit representations, respectively. The last three rows show different 3D samples generated per sketch in voxel, implicit, and CAD formats.}
\label{fig:qual_results}
\end{figure*}

In \autoref{fig:qual_results}, we visualize sample generated 3D shapes in different representations such as voxel, implicit, and CAD from sketches of different domains. As shown, our method performs reasonably well on different types of sketches (from simple to professional drawings), particularly when there is ambiguity (such as the view angle of drawings) given the nature of 2D sketches.

\subsection{Human Perceptual Evaluation}
\label{sec:human_perceptual_eval}
%%% 

\begin{figure}
    % 
    % \vspace{10pt}
    \centering    
    \includegraphics[width=0.4\textwidth]{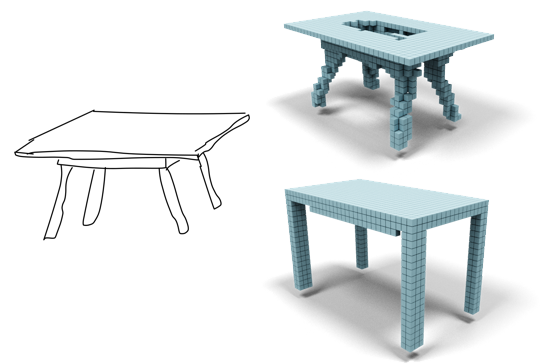}
    \caption{An example of the images shown to the Mechanical Turk crowd workers.  The question is ``Which of the 3D models on the right hand side best matches the sketch on the left hand side?" and the  options are ``Top" and ``Bottom".}
    \label{fig:groundtruth_setup}
\end{figure}

In addition to generating shapes in the same broad object category as abstract hand drawn sketches, our method is also able to incorporate geometric and stylistic details from a sketch or concept design into the final 3D model.   To demonstrate this quantitatively, we run a human perceptual evaluation using Amazon SageMaker Ground Truth and crowd workers from the Mechanical Turk workforce ~\cite{mishra_2019}.  We evaluate  691 generated models, conditioned on sketches from TU-Berlin \cite{eitz2012hdhso}, ShapeNet-Sketch  \cite{zhang2021sketch2model}, ImageNet-Sketch \cite{wang2019learning} and QuickDraw \cite{ha2018a}.   

The human evaluation is posed as a two-alternative forced choice study \cite{fechner1860elemente}.  The crowd workers are shown images with a sketch on the left hand side and two images of generated 3D models on the right hand side.  An example is shown in \autoref{fig:groundtruth_setup}.  One of the generated models was conditioned on the sketch shown, while the other was conditioned on a randomly selected sketch from the same object category.  The crowd workers are asked the question ``Which of the 3D models on the right hand side best matches the sketch on the left hand side?".  The study is designed to measure the extent to which humans perceive our generated 3D models as preserving the shape and stylistic details presented in the sketch, as opposed to simply creating a model from the same object category.

We show each image to 7 independent crowd workers and count the number of images for which 4 or more of them correctly identify the 3D model which was conditioned on the sketch.  The results are shown in \autoref{tab:human_eval_by_dataset}.   On average $71.1\%$ of our generated 3D models are correctly identified by a majority of the crowd workers.  We note that the sketches in TU-Berlin and ShapeNet-Sketch give rise to generations which were easier for the crowd workers to identify, with $74.9\%$ and $73.1\%$ being selected correctly.   While these sketches often have a high level of abstraction, they communicate enough detail about the shape for our method to create distinctive 3D models which humans can identify.      While ImageNet-Sketch contains superior artwork, often with shading, shadows and other cues to the 3D nature of the shapes, many of the pictures contain full scenes with backgrounds and additional superfluous details.   This makes the generation of single objects more challenging, which is reflected by the fact that only $68.1\%$ are correctly identified by the crowd workers.  We note qualitatively that in cases where shaded sketches do not contain backgrounds or additional clutter the generated results look better, indicating the utility of our method for quickly generating 3D models from concept designs. The sketches in the QuickDraw dataset are sourced from the from the online game \emph{Quick, Draw!} \cite{quickdraw-game}, in which contributors are asked to drawn a shape in less than 20 seconds.  QuickDraw is the most abstract and noisy sketch dataset, with many of the sketches being drawn with a computer mouse.  While our method typically generates 3D shapes of the correct category, only $67.9\%$ of the generations are correctly identified by the crowd workers.

\begin{table}
\centering
{\small
\begin{tabular}{lc}
\toprule
\textit{Dataset} & \% correctly identified \\
\midrule
All                & 71.1\% \\
TU-Berlin          & 74.9\% \\
ShapeNet-Sketch    & 73.1\% \\
ImageNet-Sketch    & 68.1\% \\
QuickDraw          & 67.9\% \\
\bottomrule
\end{tabular}
}
\vspace{10pt}
\caption{Results of the human perceptual evaluation by dataset.  We show the percentage of generated models for which the majority of the 7 crowd workers reviewing each image correctly identified the 3D model conditioned on the sketch.} % \caption
\label{tab:human_eval_by_dataset}
\end{table}

\subsection{Comparison with Supervised Models}

As there is currently a lack of zero-shot methodologies for generating shapes from sketches, we compared our results to those of a supervised approach called Sketch2Model \cite{zhang2021sketch2model}, which was trained on a dataset of paired sketch-3D shapes. We evaluated both methods using our classifier accuracy metric, and the results are presented in \autoref{tab:supervised-comparison}. Our model was not exposed to any sketch-3D pairings during training, but it displays superior generation capabilities compared to Sketch2Model across different datasets.
We attribute this difference in performance to several factors. Firstly, we believe that Sketch2Model may be more effective for single-category training rather than for the 13 categories in the ShapeNet dataset. Additionally, because Sketch2Model is a supervised method, it was not exposed to out-of-distribution sketches during training, which may have caused its performance to deteriorate. We provide further details and qualitative comparison with Sketch2Model and other supervised methods in the supplementary material.

\begin{table}
\centering

{\small
\begin{tabular}{c|cccc}
\toprule
\textit{Method} & \textbf{QD-Acc$\uparrow$} & \textbf{TU-Acc$\uparrow$} & \textbf{SS-Acc$\uparrow$}  & \textbf{IS-Acc$\uparrow$}  \\
\midrule
S2M \cite{zhang2021sketch2model} & 27.4 & 19.8 & 26.0 & 12.0 \\
\textbf{Ours} & \textbf{58.8} & \textbf{81.5} & \textbf{79.7} & \textbf{74.2} \\
\bottomrule
\end{tabular}
}
\vspace{10pt}
\caption{Classification accuracy comparison with the supervised method Sketch2Model.} % \caption
\label{tab:supervised-comparison}
\end{table}

\begin{table*}
\centering
\setlength{\tabcolsep}{13pt}
{\small
\begin{tabular}{c|c|c|c|cccc}
\toprule
\textit{Resolution} & \textit{CFG} & \textit{ Network} & \textit{Dataset} & \textbf{QD-Acc$\uparrow$} & \textbf{TU-Acc$\uparrow$} & \textbf{SS-Acc$\uparrow$}  & \textbf{IS-Acc$\uparrow$}  \\
\midrule
% 1 & $\times$ & B-32 & OpenAI data &  \\
1 x 512  & $\times$ & B-32 \cite{radford2021learning} & OpenAI \cite{radford2021learning} & 36.65 & 61.14 & 62.86 & 55.96\\
50 x 768  & $\times$ & B-32 \cite{radford2021learning} & OpenAI \cite{radford2021learning} & 37.85 & 63.25 & 63.78 & 52.79\\
50 x 768 & $\checkmark$ & B-32 \cite{radford2021learning} & OpenAI \cite{radford2021learning} & 38.86 & 65.86 & 67.36 & 49.19 \\
\midrule
197 x 768 & $\checkmark$ & B-16 \cite{radford2021learning} & OpenAI \cite{radford2021learning} &  38.47 & 71.66 & 70.72 & 61.10 \\
257 x 1024 & $\checkmark$ & L-14 \cite{radford2021learning} & OpenAI \cite{radford2021learning} & \textbf{55.45} & 77.15 & \textbf{74.53} & \textbf{69.06} \\

144 x 3072 & $\checkmark$ & RN50x16 \cite{radford2021learning} & OpenAI \cite{radford2021learning} &  34.61 & 70.81 & 
 58.82& 59.00 \\
196 x 4096 & $\checkmark$ & RN50x64 \cite{radford2021learning} & OpenAI \cite{radford2021learning} & 46.93 &  73.79 & 59.41 & 64.19\\

\midrule
257 x 1024  & $\checkmark$ & Open-L-14 \cite{ilharco_gabriel_2021_5143773} & LAION-2B \cite{schuhmann2022laion} & 54.63 & \textbf{77.60} & 69.03 & 68.35 \\

256 x 1024 & $\checkmark$ & DINO-L-14 \cite{oquab2023dinov2} & DINOv2  \cite{oquab2023dinov2} & 39.73 & 71.12 & 72.10 & 55.94   \\
197 x 1024 & $\checkmark$ & MAE-L \cite{he2022masked} & ImageNet \cite{deng2009imagenet} & 19.31 & 30.52 & 38.79 & 26.65  \\
257 x 1280 & $\checkmark$ & MAE-H \cite{he2022masked} & ImageNet \cite{deng2009imagenet} & 18.70 & 31.63 & 37.47& 31.42  \\

\bottomrule
\end{tabular}
}
\caption{Ablations on using different pre-trained models. Resolution indicates the size of the local features obtained from pre-trained model whereas CFG stands for classifier free guidance.} % \caption
\label{tab:pre_abl}

\end{table*}

\subsection{Investigating Pre-Trained Models}
\label{pre-trained_sec}

This section involves an extensive study of several pre-trained models that are open-sourced and trained on different datasets. The results are present in Table \ref{tab:pre_abl}. 
There are 3 major things we investigate through this experiment. 
First, we investigate the importance of  utilizing local grid features of pre-trained models. Typically, pre-trained models possess a global projection vector that is employed for downstream tasks like classification. We compare the efficacy of conditioning our generative model with the global projection vector (row 1) versus the local grid features (row 2). Our findings demonstrate that leveraging local grid features yields better performance compared to the global projection vector for most of the datasets. Furthermore, even from a visual standpoint, we observe that local grid features preserve more local details. It is worth noting that these accuracies are further improved by utilizing classifier-free guidance (CFG), as illustrated in row 3. 

Next, we investigate the role of size of pre-trained models and find that increasing the size of the same class of pre-trained model, despite being trained on the same data, results in better zero-shot performance. This phenomenon is evident in the case of the ViT-based \cite{dosovitskiy2020image} CLIP model, where upgrading from the B-32 model to the L-14 model yields a significant improvement in performance. This trend is also observed in the ResNet-based \cite{he2016deep} models. Interestingly, it is worth mentioning that the ResNet-based \cite{he2016deep} models perform worse than their corresponding ViT-based \cite{dosovitskiy2020image} CLIP models. This could be attributed to the ResNet models' emphasis on high-frequency, textural features \cite{ghiasi2022vision}. 

Finally,  we explore how different datasets impact the training of these models. Our findings indicate that the model's performance remains comparable when trained on extensive datasets such as LAION-2B \cite{schuhmann2022laion}, DINOv2 Dataset \cite{oquab2023dinov2} or OpenAI internal dataset \cite{radford2021learning}. However, when we reduce the dataset size significantly, such as in the case of the masked autoencoder \cite{he2022masked} trained on 400 times less data from ImageNet \cite{deng2009imagenet}, its performance significantly declines. Despite being trained on the reconstruction objective, we believe that the masked autoencoder's performance drop is primarily due to the significantly reduced dataset size, as it still performs reasonably well on this task. Additionally, it is important to highlight that language supervision is unnecessary to acquire resilient features from extensive pre-trained models, as demonstrated by the outcomes of DINOv2.

\subsection{Accuracy across Different Layers } %4 days 
\label{layer_sec}

In this experiment, we explore the optimal layer of the vision transformer (L-14 model) from which we  extract the local conditioning features. Table \ref{tab:layer_abl} summarizes our findings. We note that as we delve deeper into the vision transformer architecture, the features extracted from the deeper layers contain more significant semantic information leading to higher accuracy. Moreover, this indicates that the model maintains the positional correlation between patches instead of treating them as global information repositories as visually we can see local semantic generation.

\begin{table}
\centering

{\small
\begin{tabular}{c|cccc}
\toprule
\textit{L-14 Layer} & \textbf{QD-Acc$\uparrow$} & \textbf{TU-Acc$\uparrow$} & \textbf{SS-Acc$\uparrow$}  & \textbf{IS-Acc$\uparrow$}  \\
\midrule
1 & 17.16 &  22.21 & 24.16 & 11.13 \\
7 & 16.93 & 25.03 & 33.24 & 12.52 \\
13 & 29.43 & 51.50 & 59.64 & 40.98 \\
19 & 54.03 & 75.77 & 74.45 & 63.83 \\ 
24 &  \textbf{55.45} & \textbf{77.15} & \textbf{74.53} & \textbf{69.06} \\
\bottomrule
\end{tabular}
}
\caption{Accuracy across 24 layers of L-14 architecture.} % 
\label{tab:layer_abl}
\end{table}

\subsection{Design Choices for Conditioning} %4 days 
\label{cond_sec}
Table \ref{tab:att_abl} presents our investigation into the impact of the mapping network's size and the attention mechanism used for conditioning the image features to the transformer. Our results show that incorporating a  mapping layer does enhance the model's performance, with the optimal number of MLP layers being two. Furthermore, our findings suggest that cross-attention with a learnable positional embedding is the most effective conditioning mechanism, as evidenced by the deteriorated  performance on removing positional embedding or using self attention as shown in the last two rows of the table.

% In Table \ref{tab:att_abl}, we probe the size of mapping network as well as the attention mechanism used for conditioning the image features to the transformer.  We observe that having a common mapping layer does improve the performance slightly with 2 MLP layers being optimal. Moreover, we find that cross attention with learnable positional embedding is the most optimal attention mechanism. This can be observed from the last 2 rows.   

\begin{table}
\centering
{\small
\begin{tabular}{cc|cccc}
\toprule
\textit{M} & \textit{Att. type} & \textbf{QD-Acc$\uparrow$} & \textbf{TU-Acc$\uparrow$} & \textbf{SS-Acc$\uparrow$}  & \textbf{IS-Acc$\uparrow$}  \\
\midrule
0 & cross & 55.45 & 77.15 &74.53 & 69.06  \\
1 & cross& 54.69 & 77.91 & 77.19 & 69.17\\
2 & cross& \textbf{57.52} & \textbf{79.41} &  \textbf{78.18}& 70.53\\
3 &  cross & 52.10 & 76.85 & 76.85 & 68.03\\ 
\midrule
2 & no pos & 55.81 & 78.75  & 78.02 & \textbf{71.05}\\
2 & self & 49.33 & 76.97 & 75.42 & 69.07\\
\bottomrule
\end{tabular}
}
\caption{Ablation on design choices for conditioning. M is the number of MLP layers in mapping network (0 meaning just a linear layer). Note, self stands for self attention, cross for cross attention with positional embedding and no pos stands for cross attention with no positional embedding.} % \caption
\label{tab:att_abl}
\end{table}

\subsection{Effect of Augmentation}
% \subsubsection{Add Feature space noise}
In our final investigation, we explore whether the addition of data augmentation improves the accuracy of shape generation across datasets. The results are summarized in Table \ref{tab:aug_abl}. We make two noteworthy observations. Firstly, even without data augmentation, our method performs relatively well, indicating the robustness of pre-trained models. Secondly, different types of augmentations have a more significant impact on certain datasets than others. For instance, affine transformation significantly enhances the performance of QuickDraw and ImageNet Sketch, while canny edge augmentation is more effective for the ShapeNet Sketch dataset. Consequently, we decide to train a network with all augmentations and find that, on balance across datasets, it performs the best. 

\begin{table}
\centering

{\small
\begin{tabular}{c|cccc}
\toprule
\textit{Augmentation} & \textbf{QD-Acc$\uparrow$} & \textbf{TU-Acc$\uparrow$} & \textbf{SS-Acc$\uparrow$}  & \textbf{IS-Acc$\uparrow$}  \\
\midrule
No Aug & 57.52 & 79.41 & 78.18 & 70.53 \\
Affine (A)&  \textbf{61.52} & 78.96 & 77.95 & 74.19 \\
Color (C) &   58.37 & 78.03 & 77.15 & 71.24\\
%Blur (B) & 58.10 & 77.93 &76.54 & 70.18  \\
Canny (CN) &  53.15 & 79.04 & \textbf{80.48} & 68.81  \\
CN + A + C & 58.96 & \textbf{81.48} & 79.71 & \textbf{74.20}\\
\bottomrule
\end{tabular}
}
\caption{Effect of different augmentations on the shapes.} % \caption
\label{tab:aug_abl}
\end{table}

\section{Conclusion}
In this paper, we demonstrate how a 3D generative model conditioned on local features from a pre-trained large-scale image model such as CLIP can be used to generate 3D shapes from sketches. We show how this method can generate multiple shapes for different abstraction of sketches and can be applied to multiple 3D representations.
Future work will involve training on much larger and diverse 3D shape datasets and consequently testing on different styles of sketches and levels of details.

{\small
\bibliographystyle{ieee_fullname}
\bibliography{egbib}
}

\appendix

% --- PDF will be split by an editor (e.g. macOS preview), so need to restart from page 1

% --- repeat the title (AT: haven't found a more elegant way to do this...)
\twocolumn[
\centering
\Large
\vspace{0.5em}Supplementary Material \\
\vspace{1.0em}
] %< twocolumn
\appendix

% \section{Appendix}
\section{Human Perceptual Evaluation}
In Subsection 4.2 (main paper) we show the results of the  human perceptual study broken down according to the dataset which the target sketch came from.  In addition, the results can be broken down based on object category of the target sketch, as shown in \autoref{tab:human_eval_by_class}.   We see a wide range of performance across the different object categories, with ``lamps" being correctly identified $89.1\%$ of the time, while phones are identified just $52.5\%$ of the time, little better than random.   Categories which perform well, such as chair, table and sofa, tend to have distinctive shapes which are easy to communicate with sketches.  Categories like airplane and gun produce good models, but these are not distinctive enough for the human evaluators to distinguish the correct 3D model from a random model of in the same category.  Lower performance on these categories may also relate to the difficultly of drawing objects of these types. We believe having texture can further improve the human perceptual results.
\begin{table}
\centering
{\small
\begin{tabular}{lc}
\toprule
Category & \% correctly identified \\
\midrule
loudspeaker          & 68.0\% \\
airplane             & 58.5\% \\
bench                & 80.0\% \\
car                  & 69.4\% \\
boat                 & 64.3\% \\
monitor              & 70.0\% \\
lamp                 & 89.1\% \\
table                & 80.0\% \\ 
cabinet              & 83.3\% \\
sofa                 & 71.4\% \\
gun                  & 65.0\% \\
chair                & 83.6\% \\
phone                & 52.5\% \\
\bottomrule
\end{tabular}
}
\vspace{10pt}
\caption{Results of the human perceptual evaluation by class.  We show the percentage of generations the majority of the 7 crowd workers reviewing each image correctly identified the 3D model generated by conditioning on the sketch.} % \caption
\label{tab:human_eval_by_class}
\end{table}

As each generated model is rated by 7 individual crowd workers, we can count the number of raters who correctly identified the generated model, giving us a ``shape recognizably score" from 0 to 7.  In \autoref{fig:best_worst} we show examples from selected categories with the highest and lowest ``shape recognizably scores".  For ``airplane" category the least recognizable model appears to be in the wrong category,  due to the unusual orientation of the sketch.  The most and least recognizable sketch in the ``bench" category both come from the Imagenet-Sketch dataset.  The sketch for the most recognizable model contains a single bench while the sketch for the least recognizable model also contains background elements like trees and a lampost. For the ``gun" category the most recognizable model is actually from a sketch which looks nothing like a gun.  The least recognizable model is a generic gun which does not closely follow the shape of the sketch.   The Figure shows how the human evaluation is measure the ability of our method to generate distinctive shapes reflecting the geometry of the sketches as well as general generation quality.

\newcommand{\bestworstimgwidth}{0.10}
\begin{figure}
  \begin{center}
    \begin{tabular}{ cc cc }
    \multicolumn{2}{c}{Most recognizable} & 
\multicolumn{2}{c}{Least recognizable} \\
\includegraphics[width=\bestworstimgwidth\textwidth]{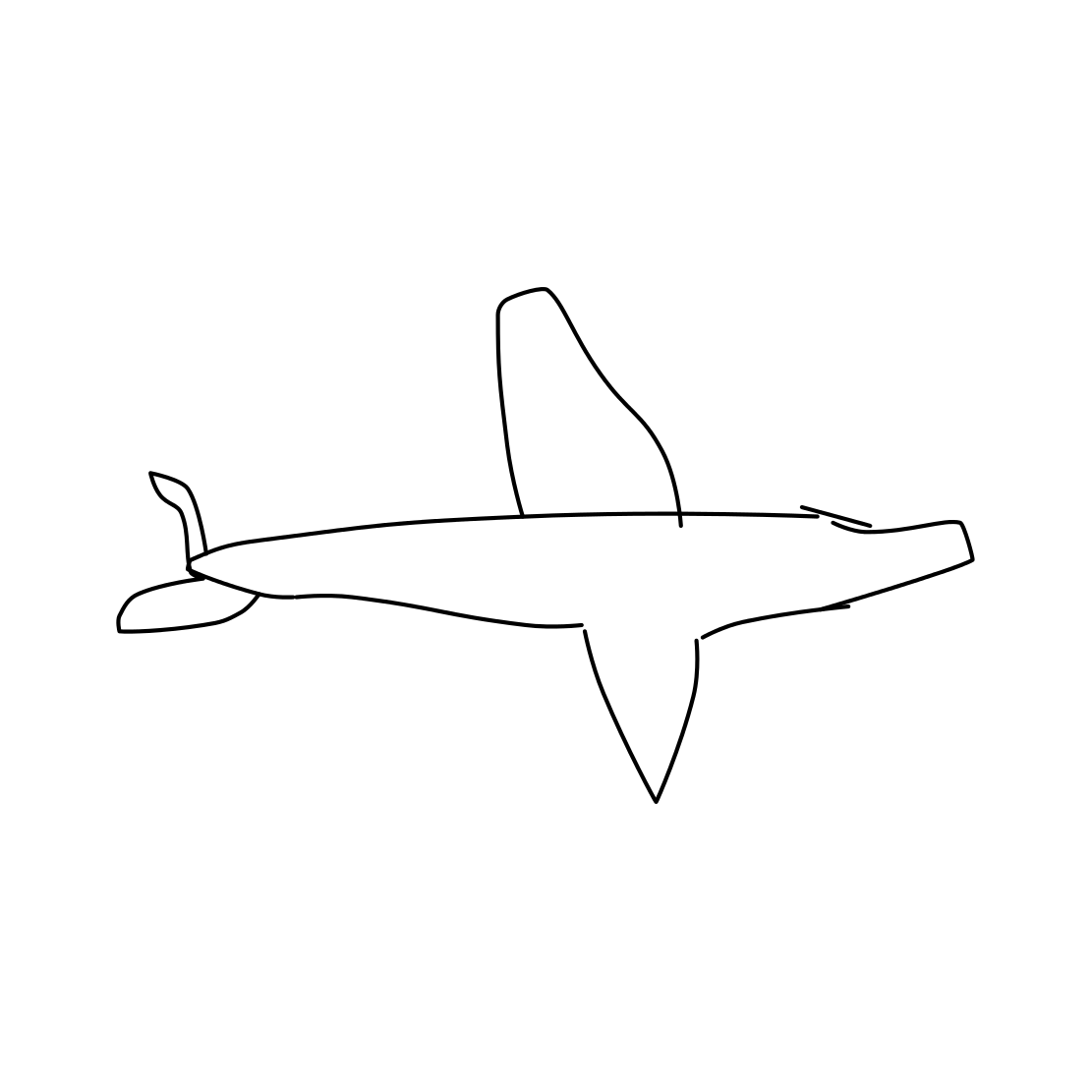} &
\includegraphics[width=\bestworstimgwidth\textwidth]{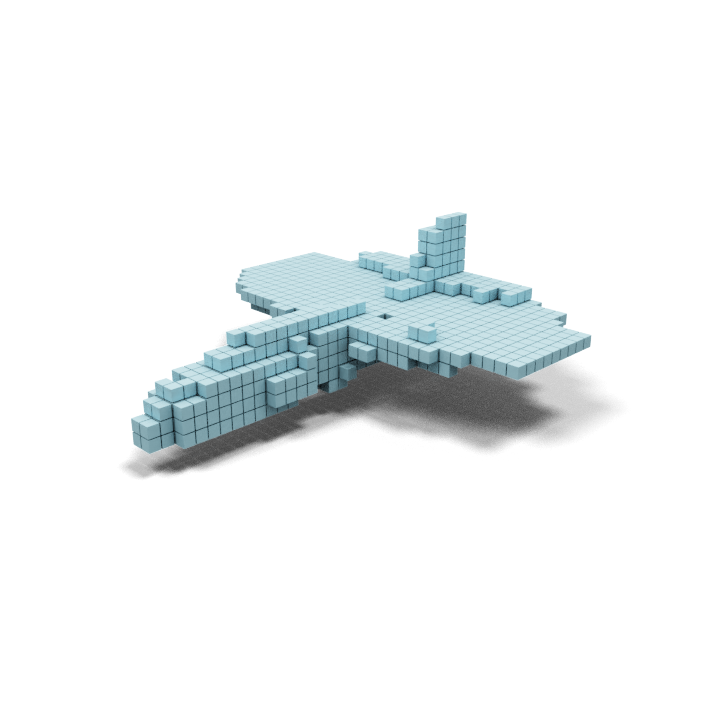} &
\includegraphics[width=\bestworstimgwidth\textwidth]{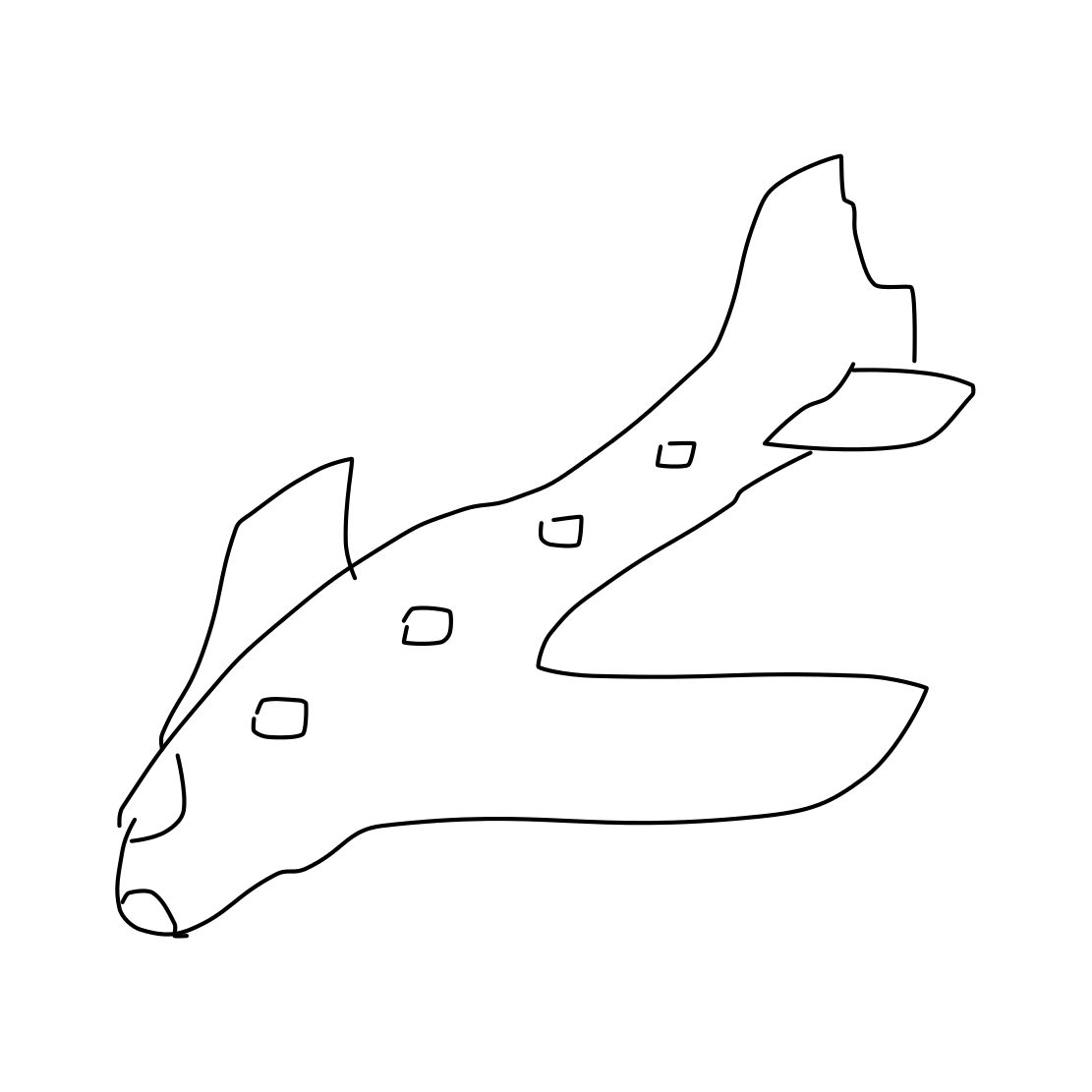} &
\includegraphics[width=\bestworstimgwidth\textwidth]{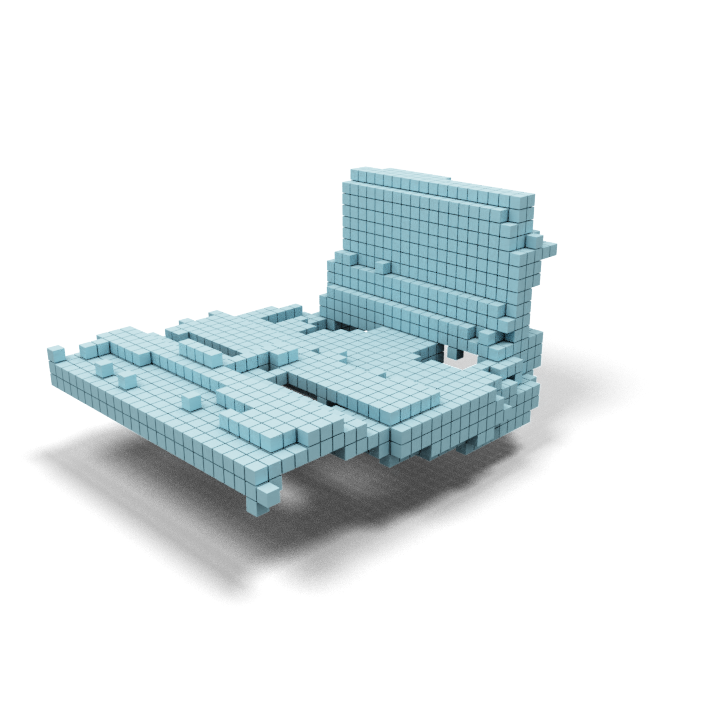} \\
% airplane & 2/7 & & 0/7 \\
% \vspace{10pt} \\

\includegraphics[width=\bestworstimgwidth\textwidth]{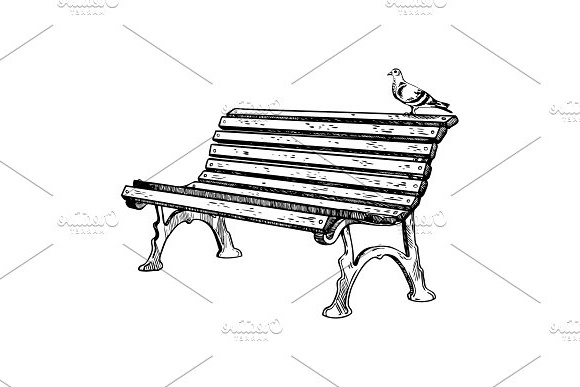} &
\includegraphics[width=\bestworstimgwidth\textwidth]{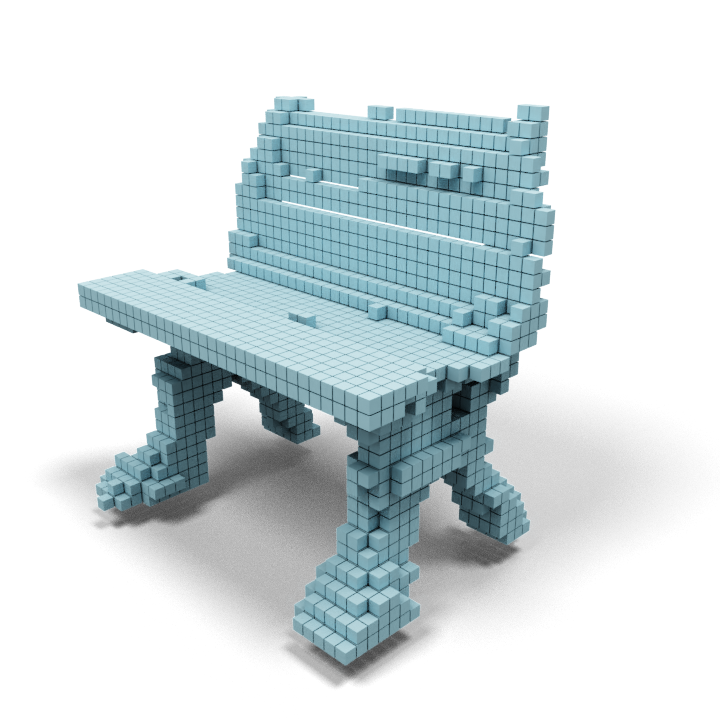} &
\includegraphics[width=\bestworstimgwidth\textwidth]{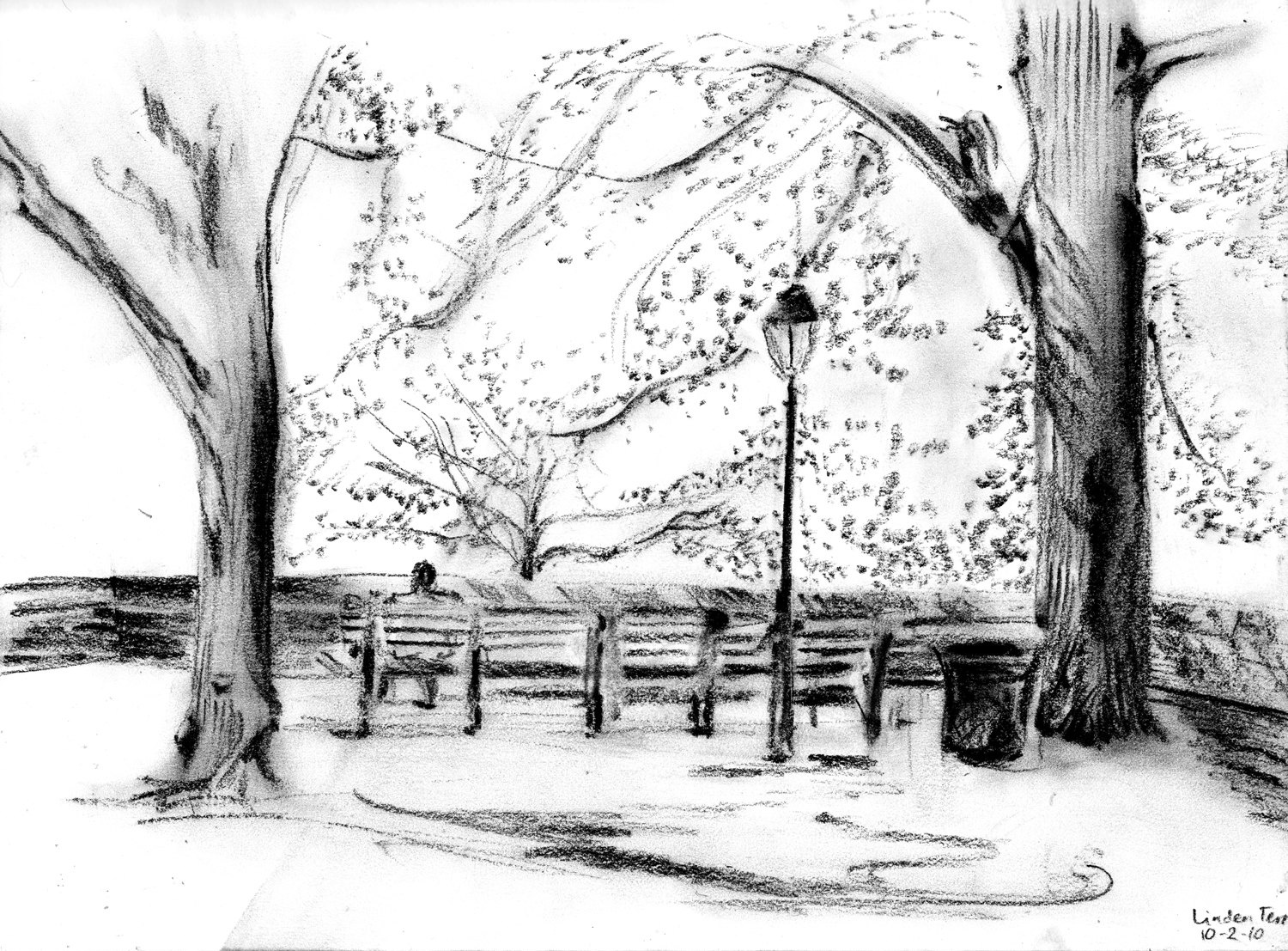} &
\includegraphics[width=\bestworstimgwidth\textwidth]{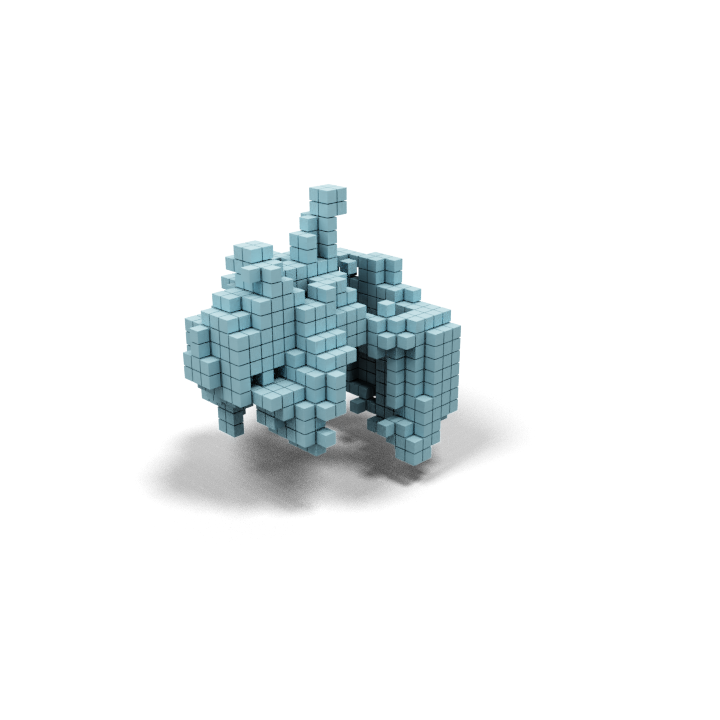} \\

\includegraphics[width=\bestworstimgwidth\textwidth]{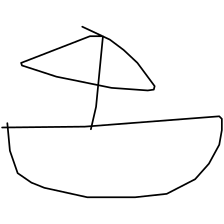} &
\includegraphics[width=\bestworstimgwidth\textwidth]{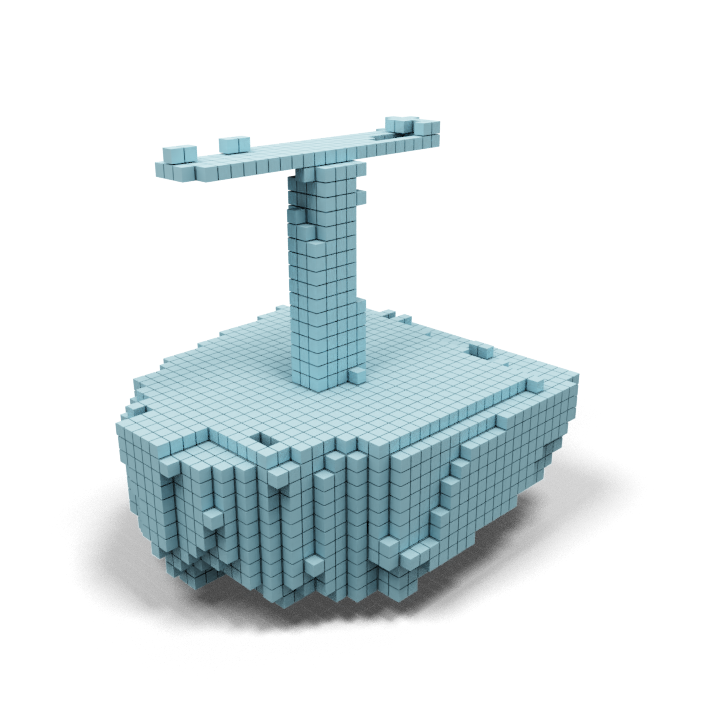} &
\includegraphics[width=\bestworstimgwidth\textwidth]{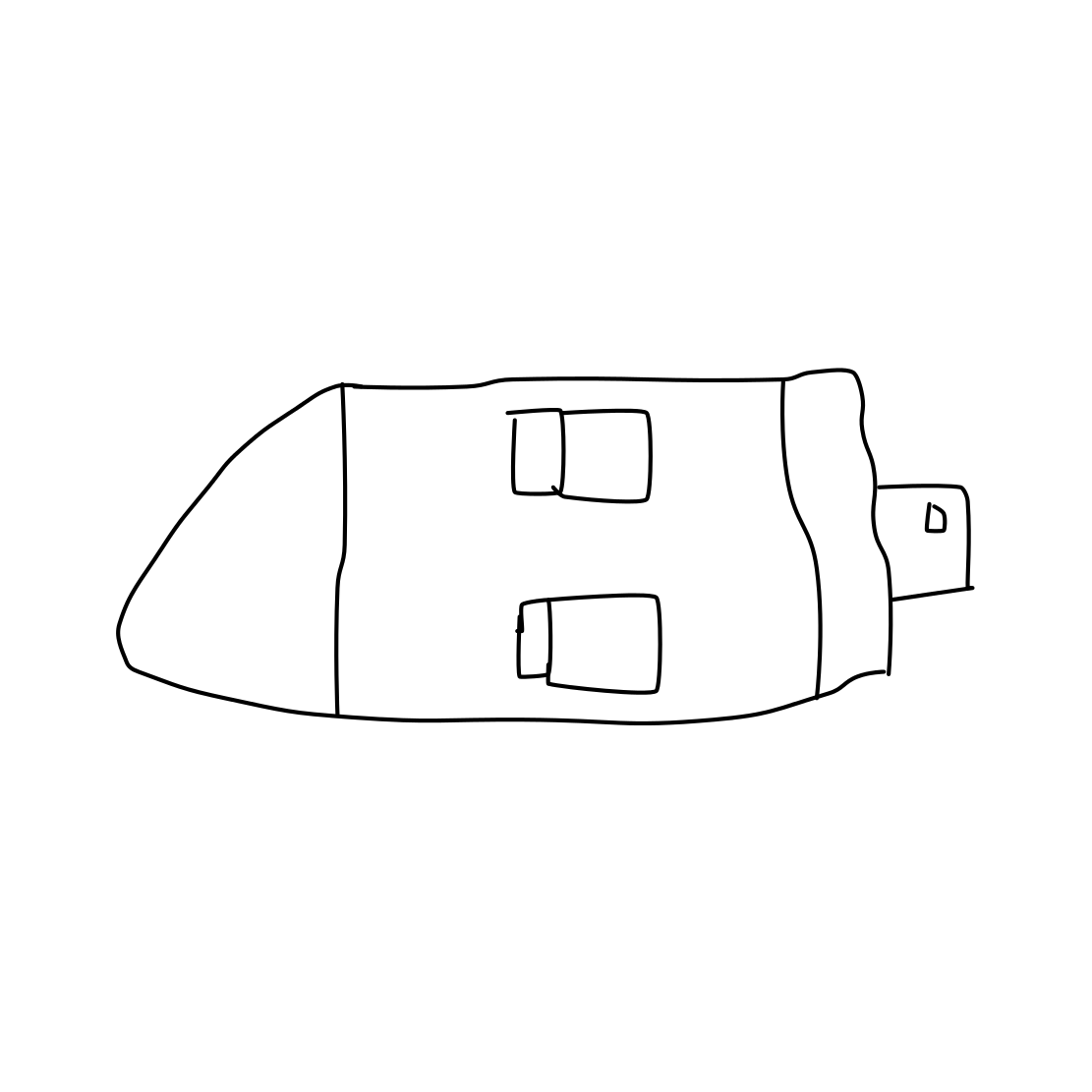} &
\includegraphics[width=\bestworstimgwidth\textwidth]{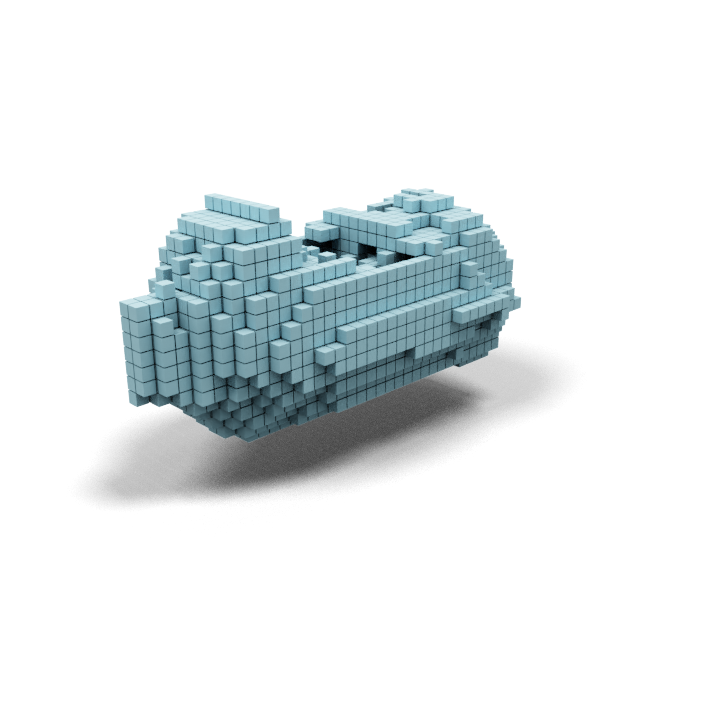} \\

\includegraphics[width=\bestworstimgwidth\textwidth]{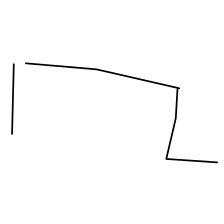} &
\includegraphics[width=\bestworstimgwidth\textwidth]{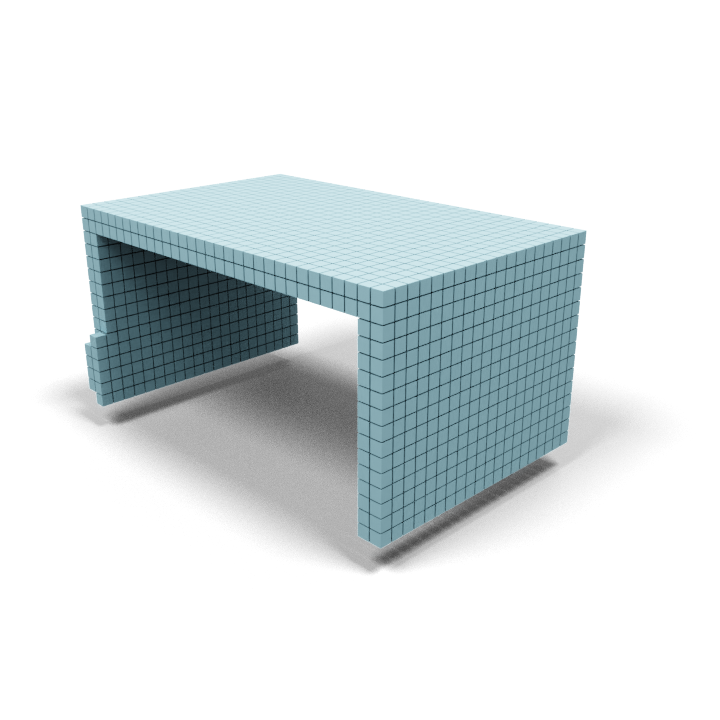} &
\includegraphics[width=\bestworstimgwidth\textwidth]{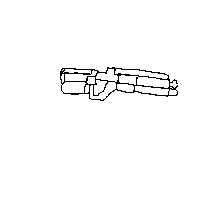} &
\includegraphics[width=\bestworstimgwidth\textwidth]{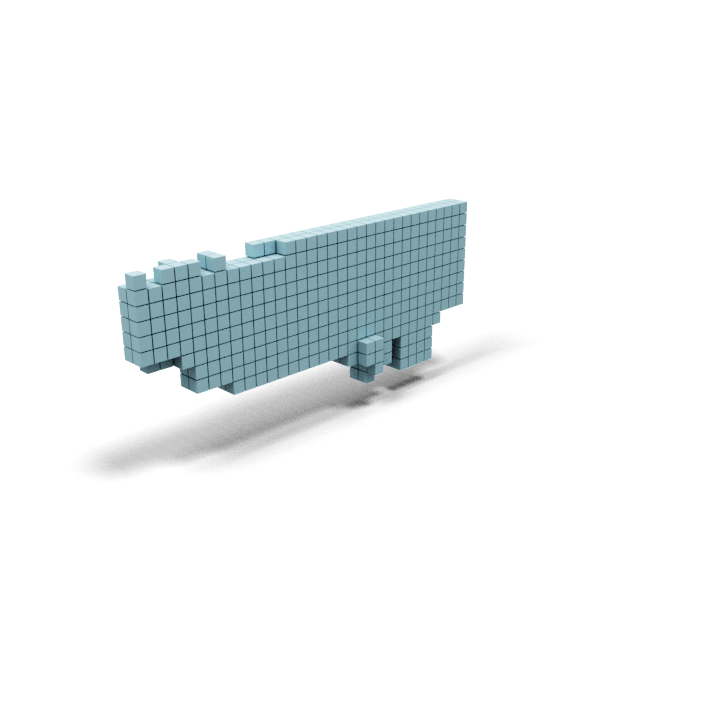} \\

\includegraphics[width=\bestworstimgwidth\textwidth]{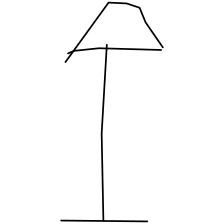} &
\includegraphics[width=\bestworstimgwidth\textwidth]{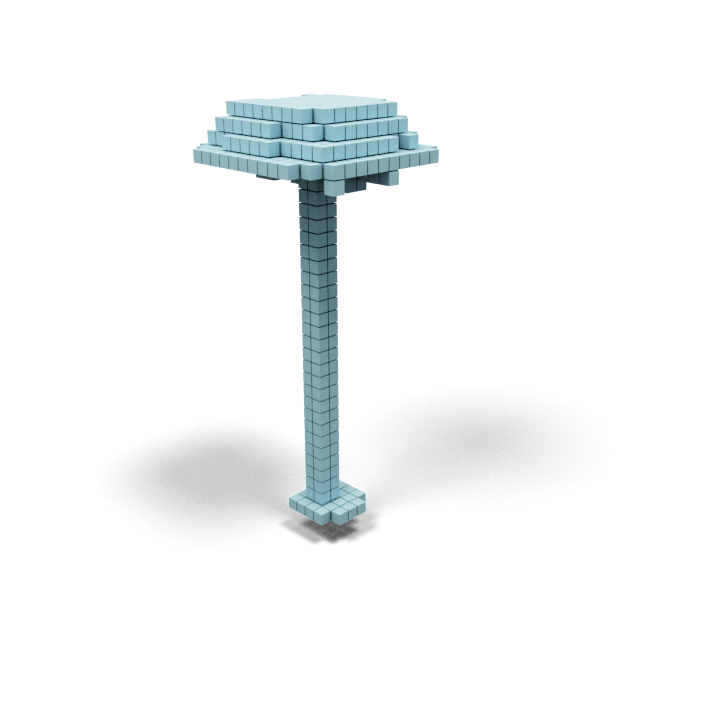} &
\includegraphics[width=\bestworstimgwidth\textwidth]{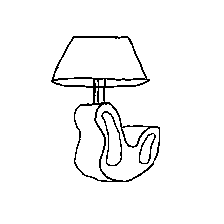} &
\includegraphics[width=\bestworstimgwidth\textwidth]{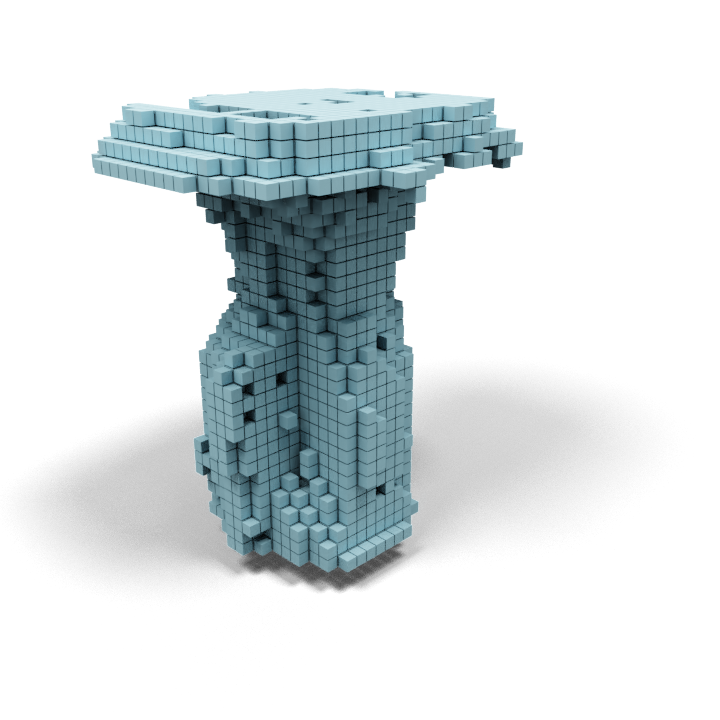} \\

\includegraphics[width=\bestworstimgwidth\textwidth]{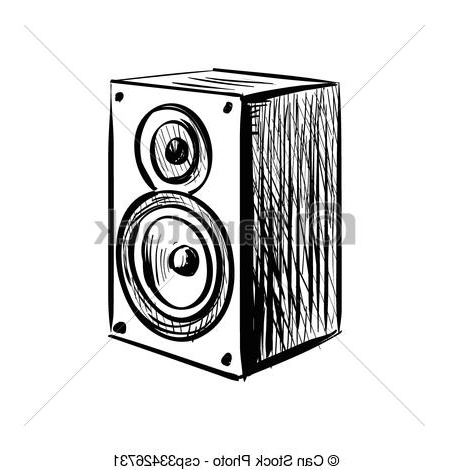} &
\includegraphics[width=\bestworstimgwidth\textwidth]{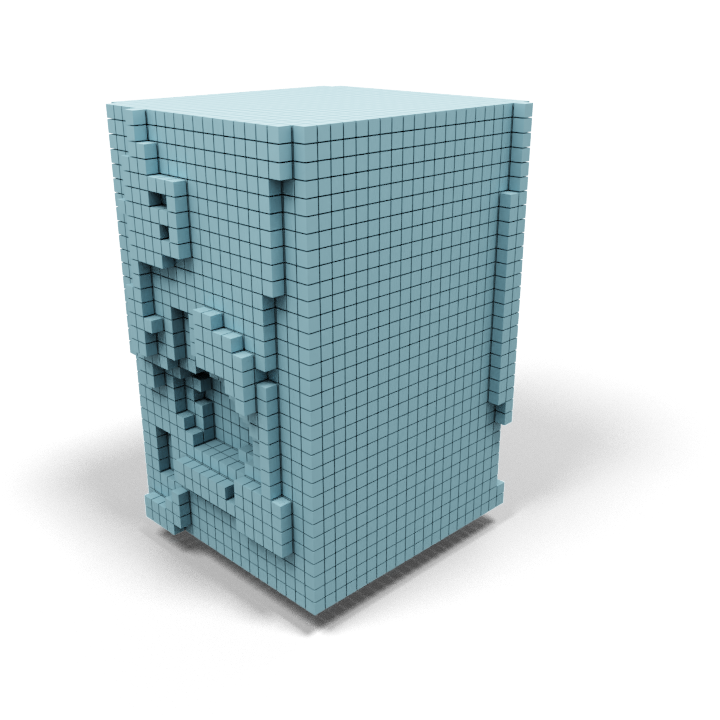} &
\includegraphics[width=\bestworstimgwidth\textwidth]{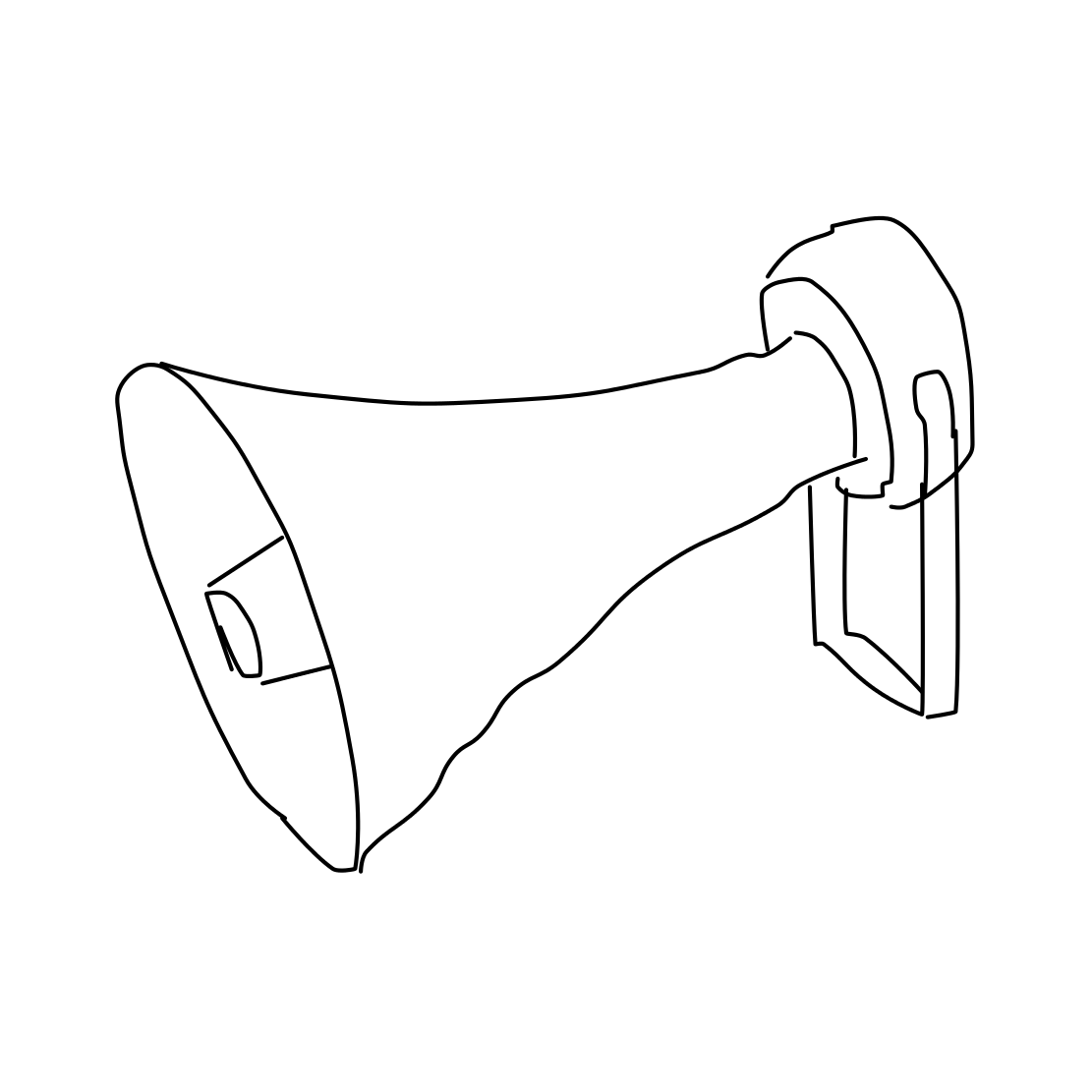} &
\includegraphics[width=\bestworstimgwidth\textwidth]{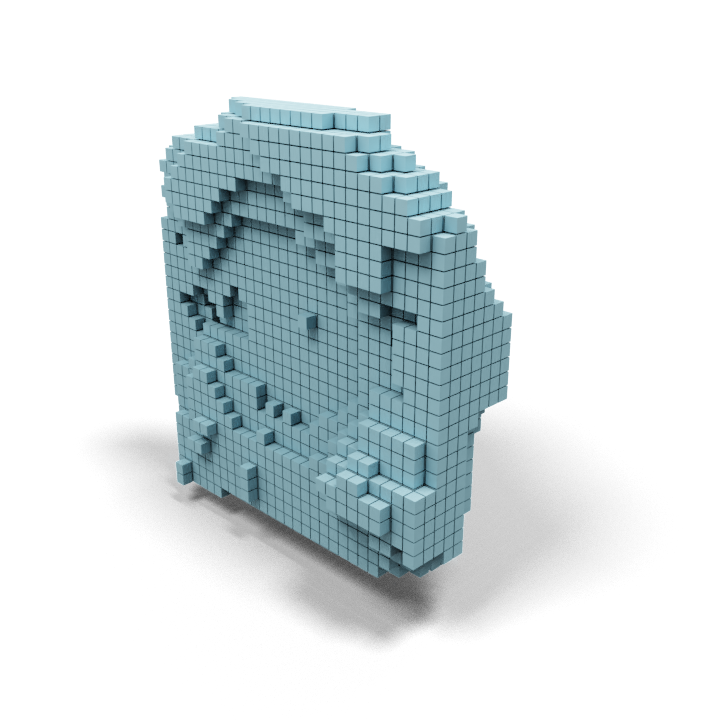} \\

\includegraphics[width=\bestworstimgwidth\textwidth]{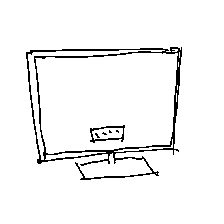} &
\includegraphics[width=\bestworstimgwidth\textwidth]{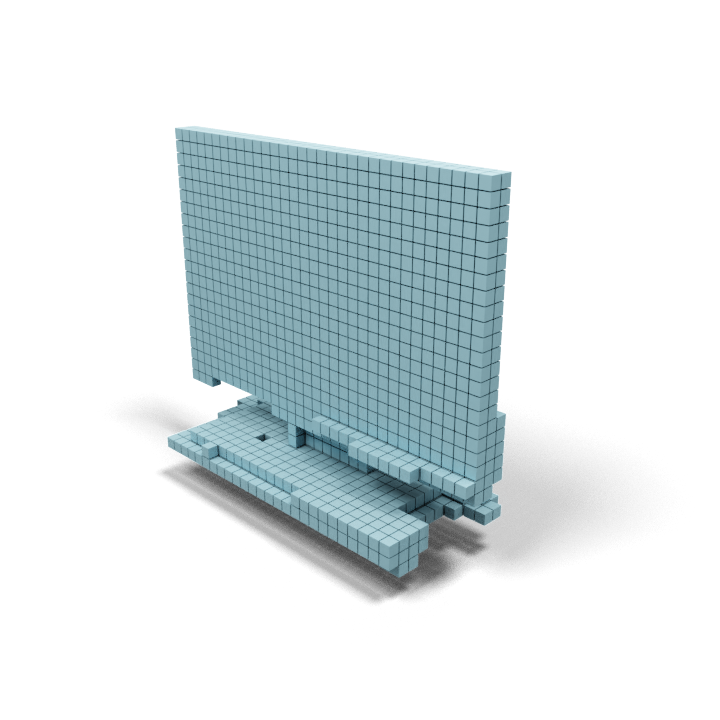} & 
\includegraphics[width=\bestworstimgwidth\textwidth]{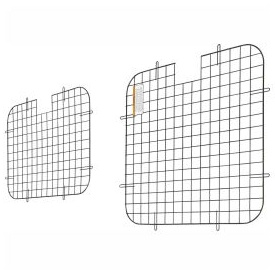} &
\includegraphics[width=\bestworstimgwidth\textwidth]{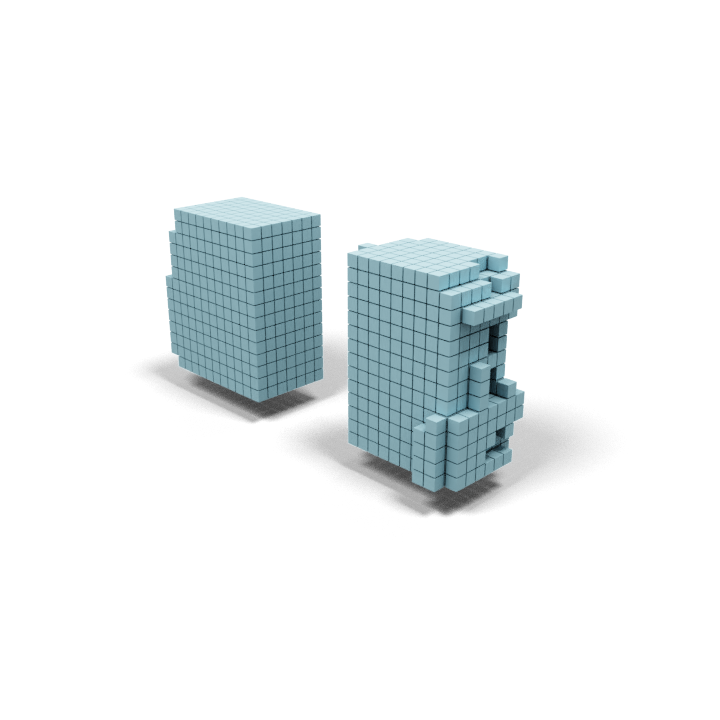} \\

\includegraphics[width=\bestworstimgwidth\textwidth]{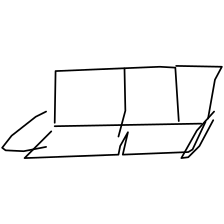} &
\includegraphics[width=\bestworstimgwidth\textwidth]{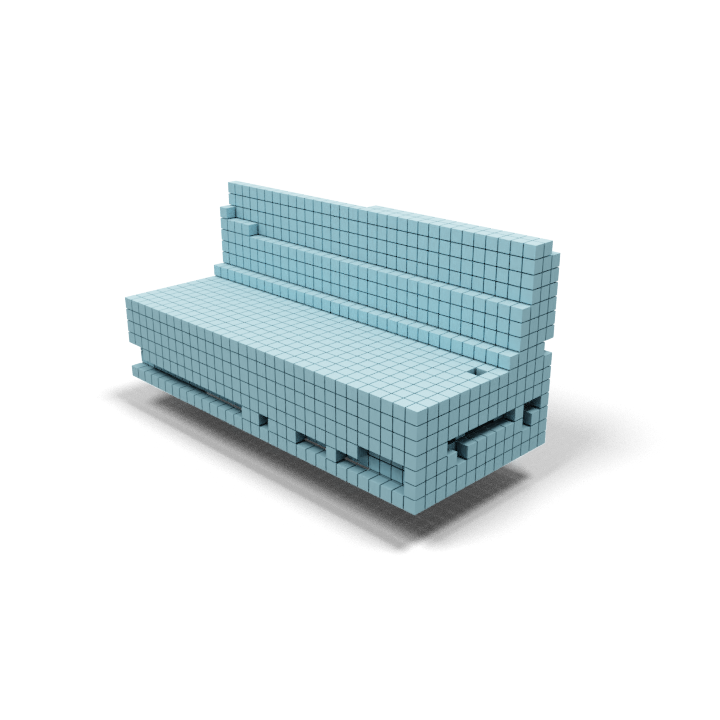} &
\includegraphics[width=\bestworstimgwidth\textwidth]{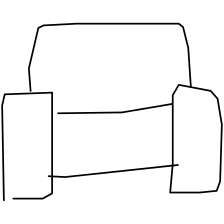} &
\includegraphics[width=\bestworstimgwidth\textwidth]{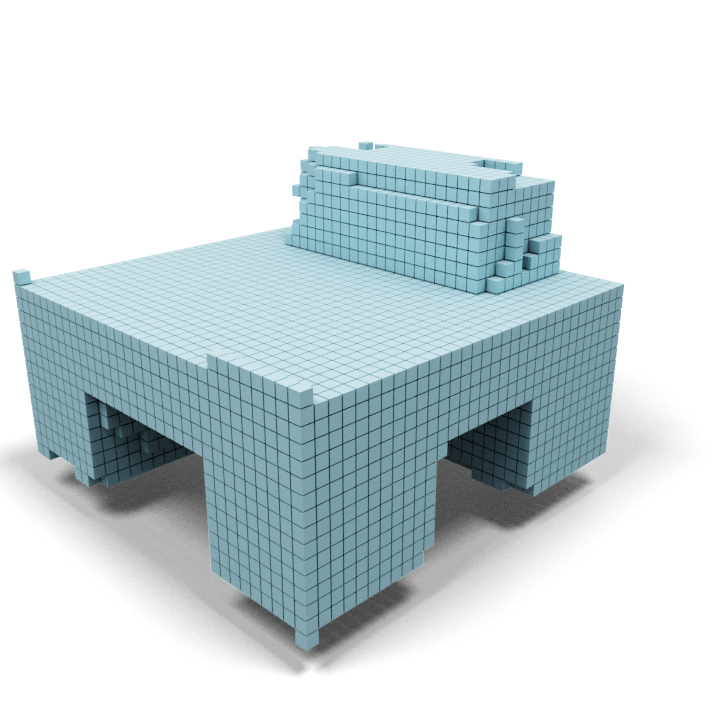} %\\
\end{tabular}
\end{center}
\caption{Generated results which were most and least recognizable for the human evaluators.}
\label{fig:best_worst}
\end{figure}

%%%%%%%%%%%%%%%%%%%

\section{Comparison with Supervised Methods}

\subsection{Quantitative comparison}

\begin{table}
\centering
{\small
\begin{tabular}{c|c|c}
\toprule
Method &  Type & IOU $\uparrow$\\
\midrule
Sketch2Mesh \cite{guillard2021sketch2mesh} & Supervised &  0.195 \\
Sketch2Model \cite{zhang2021sketch2model} & Supervised&  0.205 \\
Sketch2Point \cite{wang20203d}  & Supervised & 0.163 \\
SketchSampler \cite{gao2022sketchsampler} & Supervised &  0.244 \\
\midrule
ours & Zero-shot & 0.292 \\
\bottomrule
\end{tabular}
}
\vspace{10pt}
\caption{IOU Results Comparison. It can be seen that our method outperforms supervised methods. } % \caption
\label{tab:quan_voxel_iou}
\end{table}
We evaluate the quality of generated shapes on the ShapeNet-Sketch dataset \cite{zhang2021sketch2model} using Intersection over Union (IOU) with $32^3$ voxel shapes, as shown in \cite{maxseiner2023sketch}. This is the only dataset among the four we assessed that includes ground truth 3D voxels. We compare our results to those of other supervised methods presented in Table \ref{tab:quan_voxel_iou}, exactly as in \cite{maxseiner2023sketch}. Our generative model generates 5 shapes based on a given sketch query in the ShapeNet-Sketch dataset and averages the IOU results. Although our method is not trained on any sketch data, it outperforms the supervised baseline. This indicates that the pre-trained model's learned features are effective in enabling our method to generate 3D shapes using sketches in a zero-shot manner.

\subsection{Qualitative comparison}
\begin{figure*}
    \centering
    \begin{overpic}[trim=0cm 0cm 0cm 0cm,clip,width=0.125\linewidth]{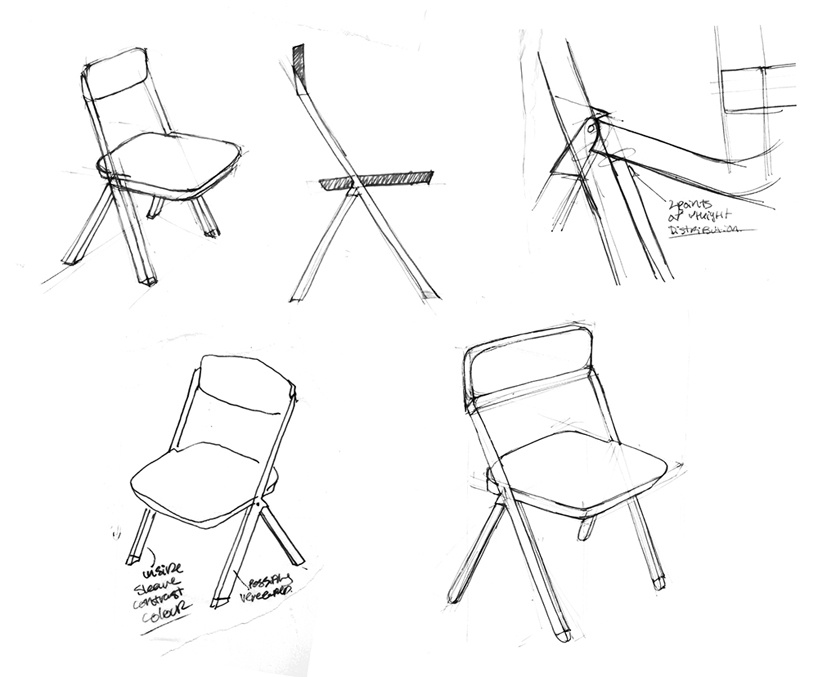}
     \put(95,100){\footnotesize{Sketch2Model \cite{zhang2021sketch2model}}}
     \put(210,100){\footnotesize{SketchSampler \cite{gao2022sketchsampler}}}
     \put(340,100){\footnotesize{\textbf{Ours}}}
    \end{overpic}%
    \includegraphics[width=0.125\textwidth]{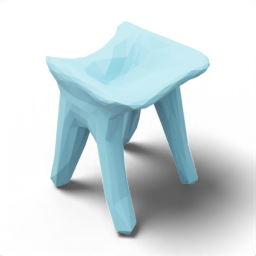}%
    \includegraphics[width=0.125\textwidth]{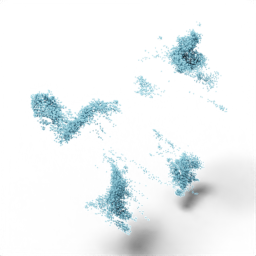}%
    \includegraphics[width=0.125\textwidth]{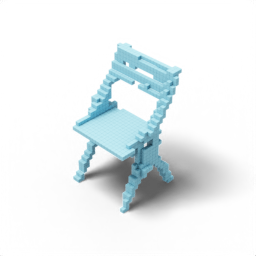}%
    \begin{overpic}[trim=0cm 0cm 0cm 0cm,clip,width=0.125\linewidth]{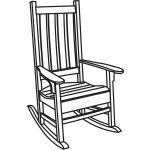}
     \put(95,100){\footnotesize{Sketch2Model \cite{zhang2021sketch2model}}}
     \put(210,100){\footnotesize{SketchSampler \cite{gao2022sketchsampler}}}
     \put(340,100){\footnotesize{\textbf{Ours}}}
    \end{overpic}%
    \includegraphics[width=0.125\textwidth]{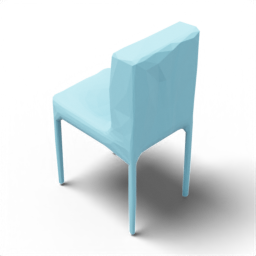}%
    \includegraphics[width=0.125\textwidth]{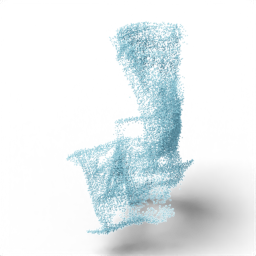}%
    \includegraphics[width=0.125\textwidth]{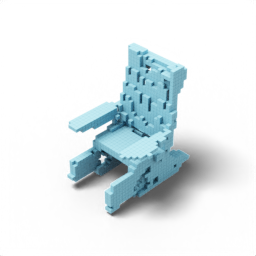}%
    
    \includegraphics[width=0.125\textwidth]{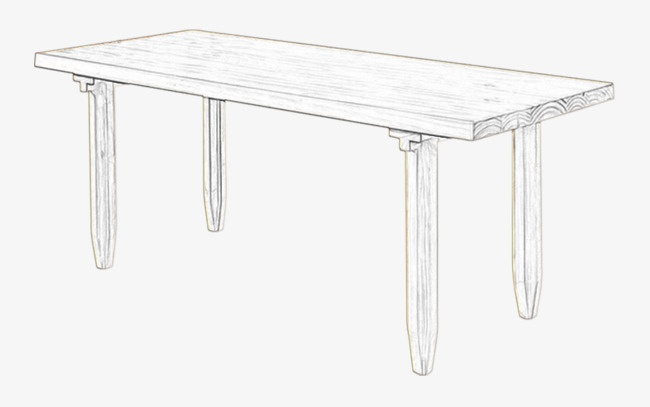}%
    \includegraphics[width=0.125\textwidth]{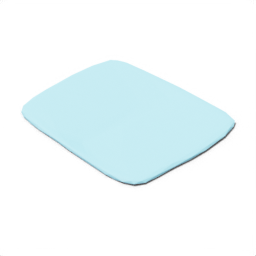}%
    \includegraphics[width=0.125\textwidth]{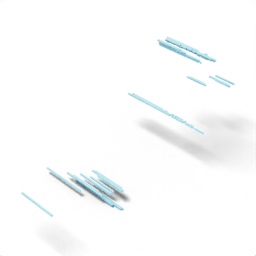}%
    \includegraphics[width=0.125\textwidth]{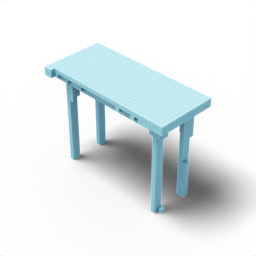}%
    \includegraphics[width=0.125\textwidth]{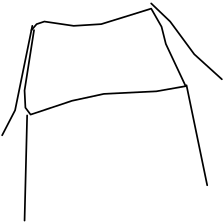}%
    \includegraphics[width=0.125\textwidth]{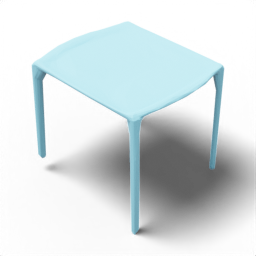}%
    \includegraphics[width=0.125\textwidth]{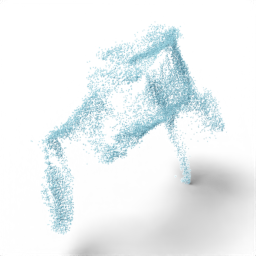}%
    \includegraphics[width=0.125\textwidth]{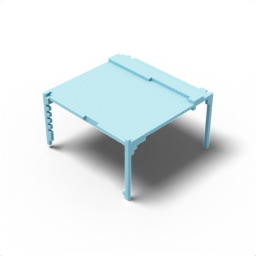}%

    \includegraphics[width=0.125\textwidth]{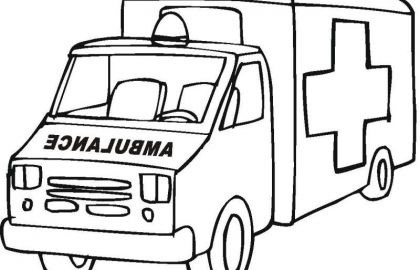}%
    \includegraphics[width=0.125\textwidth]{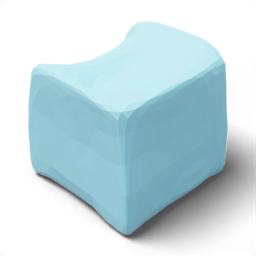}%
    \includegraphics[width=0.125\textwidth]{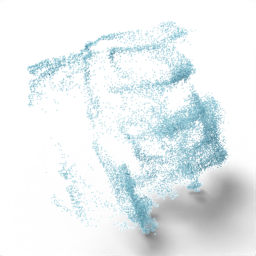}%
    \includegraphics[width=0.125\textwidth]{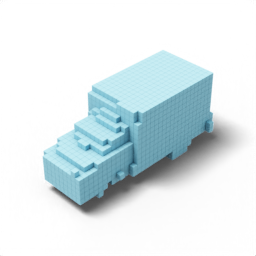}%
    \includegraphics[width=0.125\textwidth]{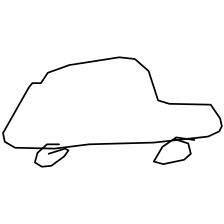}%
    \includegraphics[width=0.125\textwidth]{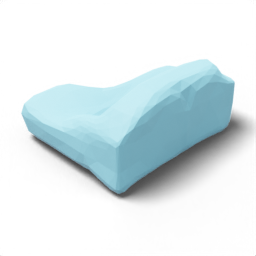}%
    \includegraphics[width=0.125\textwidth]{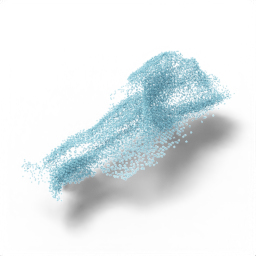}%
    \includegraphics[width=0.125\textwidth]{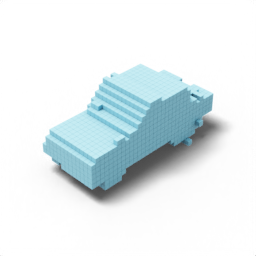}%
    
    \includegraphics[width=0.125\textwidth]{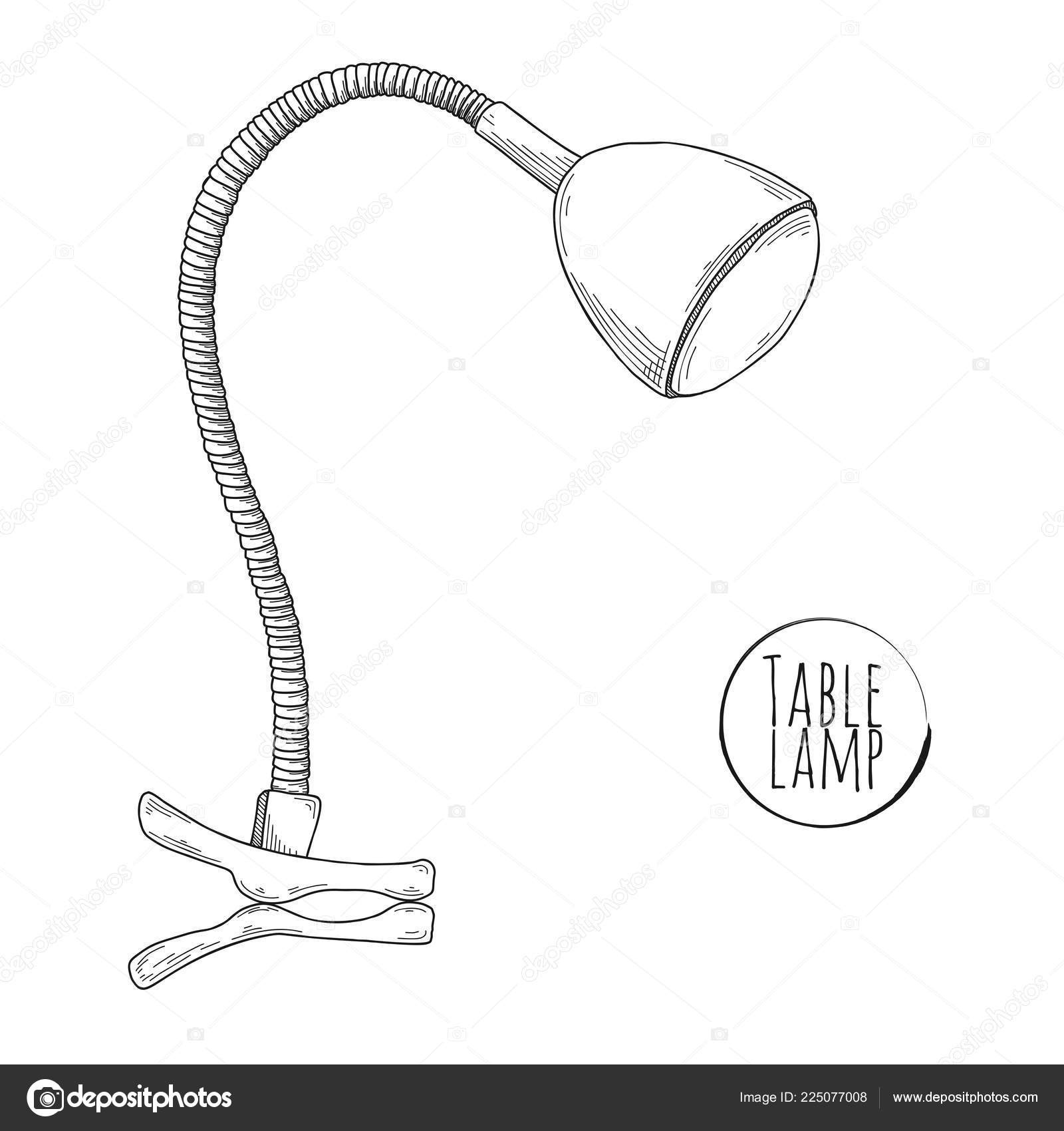}%
    \includegraphics[width=0.125\textwidth]{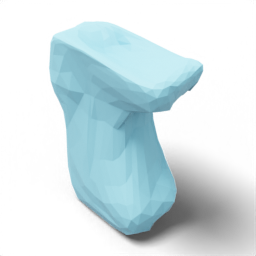}%
    \includegraphics[width=0.125\textwidth]{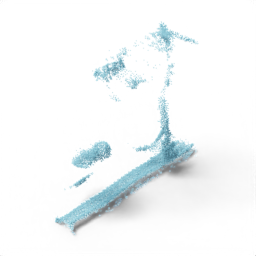}%
    \includegraphics[width=0.125\textwidth]{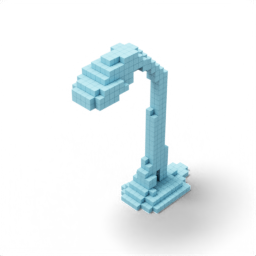}%
    \includegraphics[width=0.125\textwidth]{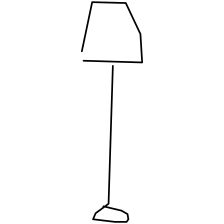}%
    \includegraphics[width=0.125\textwidth]{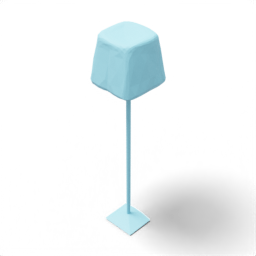}%
    \includegraphics[width=0.125\textwidth]{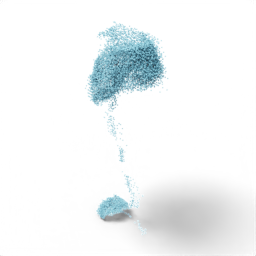}%
    \includegraphics[width=0.125\textwidth]{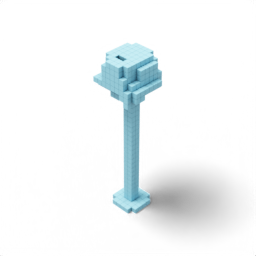}%
    
    \includegraphics[width=0.125\textwidth]{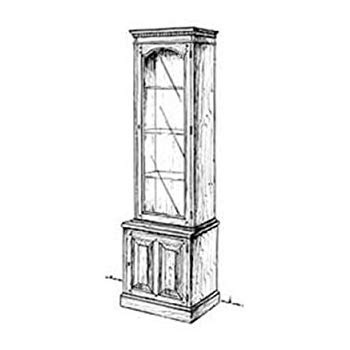}%
    \includegraphics[width=0.125\textwidth]{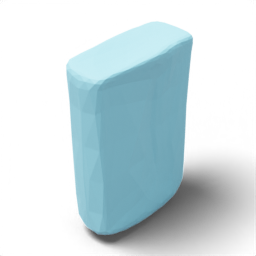}%
    \includegraphics[width=0.125\textwidth]{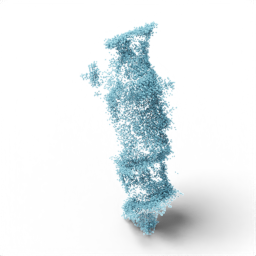}%
    \includegraphics[width=0.125\textwidth]{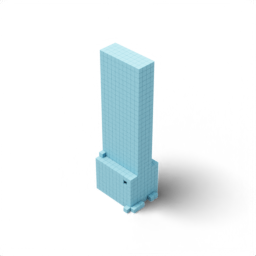}%
    \includegraphics[width=0.125\textwidth]{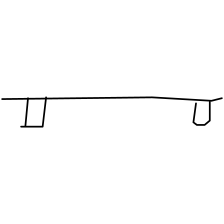}%
    \includegraphics[width=0.125\textwidth]{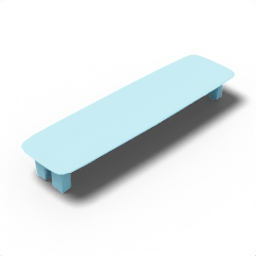}%
    \includegraphics[width=0.125\textwidth]{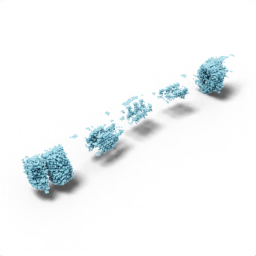}%
    \includegraphics[width=0.125\textwidth]{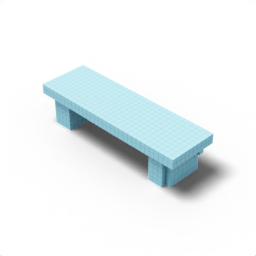}%

    \includegraphics[width=0.125\textwidth]{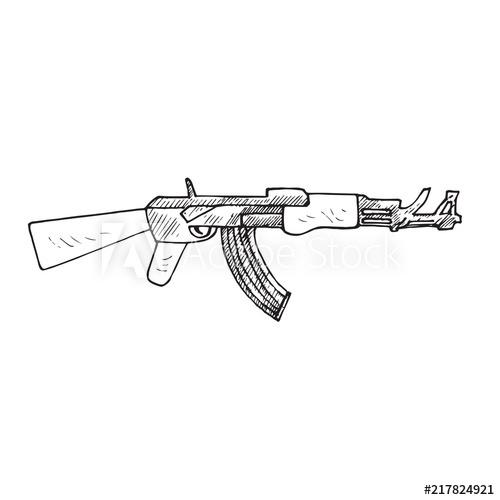}%
    \includegraphics[width=0.125\textwidth]{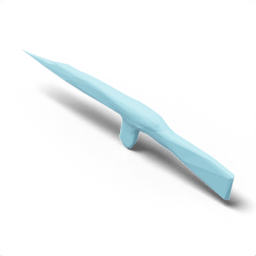}%
    \includegraphics[width=0.125\textwidth]{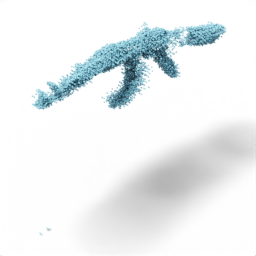}%
    \includegraphics[width=0.125\textwidth]{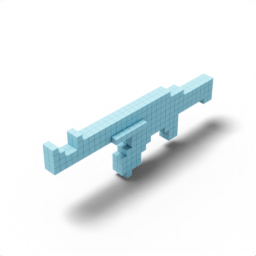}%
    \includegraphics[width=0.125\textwidth]{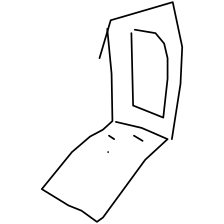}%
    \includegraphics[width=0.125\textwidth]{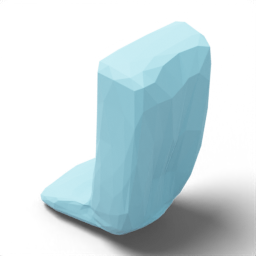}%
    \includegraphics[width=0.125\textwidth]{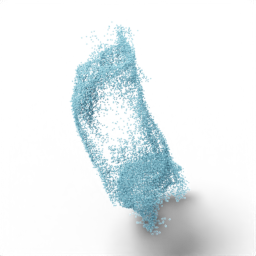}%
    \includegraphics[width=0.125\textwidth]{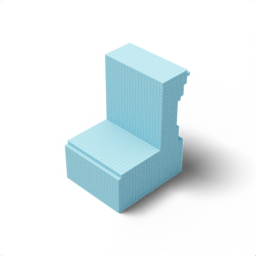}%

    \caption{Generated shapes with different methods. We can see that, compared to our method, Sketch2Model meshes lack geometric details and appear smoothed, while SketchSampler pointclouds present artifacts and outliers.}

    \label{fig:supervised}
\end{figure*}

We additionally provide a qualitative comparison with Sketch2Model \cite{zhang2021sketch2model} and SketchSampler \cite{gao2022sketchsampler}.
For this comparison, we considered diverse sketches with different levels of abstraction in the same classes of ShapeNet from four datasets: TU-Berlin \cite{eitz2012hdhso}, ShapeNet-Sketch  \cite{zhang2021sketch2model}, ImageNet-Sketch \cite{wang2019learning}, and QuickDraw \cite{quickdraw-data}. Implementation details can be found in \autoref{sec:implementation}. Results are in \autoref{fig:supervised}.

% comments
We can see that Sketch2Model reconstructed meshes often grasp the overall sketch shape, but they appear too smooth and lack geometric details. This method was originally intended for a single category scenario, as presented in the paper. However, this is often unpractical.

Similarly, SketchSampler fails to generalize to abstract or out-of-distribution sketches. The resulting point clouds present artifacts and outliers, especially in the direction of the point of view of the sketch (shapes proportion are only preserved when the point clouds are seen from this point of view). Unlike our approach, SketchSampler is designed for professional sketches only, with reliable shapes and fine-grained details. 
Thus, it cannot deal with sketches with significant deformation or only expressing conceptual ideas, like the ones in QuickDraw \cite{quickdraw-data}. 
%Note that shapes are not consistently oriented, as their orientation depends on the point of view of the sketch.  This is why we could not compare accuracy.

\section{ Architecture and Experiment Details}\label{sec:implementation}

\noindent  \textbf{Training Details.} We use the Adam Optimizer \cite{kingma2014adam} with a fixed learning rate of 1e-4 for training. The network is trained for 300 epochs during Stage 1 and for 250 epochs during Stage 2. We do not employ any learning rate scheduler during training. We train the $32^3$ voxel model solely on the ShapeNet13 dataset, while the Implicit model is trained on the ShapeNet55 subset. The CAD model is trained on the DeepCAD dataset~\cite{xu2022skexgen}. This is done to demonstrate the versatility and adaptability of our method to different datasets.

\noindent \textbf{Stage 1 Details.} For both the Implicit VQ-VAE and $32^3$ VQ-VAE we use a codebook size of 512, a grid size of $8^3$ and embedding dimensions of size 64. We employ the ResNet architecture for the $32^3$ VQ-VAE, for both the encoder and decoder. In the case of Implict VQ-VAE, we use the ResNet architecture for the encoder whereas we use a decoder that produces a  higher resolution volume, which is then queried locally to obtain the final occupancy \cite{OccupancyNetworks2019, peng2020convolutional, shapecrafter}.
The pretrained VQ-VAE from SkexGen~\cite{xu2022skexgen} is used for the CAD representation which is composed of three Transformer encoders and decoders for the topology, geometry and extrusions of a CAD model. The models output $4+2+4=10$ codes, with a total codebook size of $1000$.

\noindent \textbf{Stage 2 Details.} 
For Stage 2, we utilize a bidirectional Transformer with 8 attention blocks, 8 attention heads, and a token size of 256. We use 24 renderings \cite{choy20163d} for both the ShapeNet13 and ShapeNet55 experiments. During inference, we run the Transformer for 15 steps with classifier-free guidance, and the scale parameter is set to 3. The CLIP ViT-L/14 model is employed in all experiments, except in Table 3 of the main paper, where we conduct an ablation study over different pre-trained models. For all experiments, except Table 4, we incorporate cross-attention with learnable positional embedding and a mapping network consisting of 2 layers of MLP. We do not apply any augmentation for the quantitative experiments, except for the results presented in Table 6 and Table 2 of the main paper.
For the CAD results, we used a CLIP ViT-B/32 model.

 \noindent  \textbf{Sketch2Model.} 
The authors of Sketch2Model released ShapeNet-Synthetic as the training dataset \cite{zhang2021sketch2model}, which consists of synthetic sketches of objects from $13$ categories from ShapeNet. These objects have been rendered from $20$ different views. For training Sketch2Model, we used the official implementation provided in \cite{sketch2model-code}, along with the recommended hyperparameters. This implementation uses a step-type learning rate policy, beginning from $1e-4$ and decreasing by $0.3$ every $800$ epochs, and trains for 2000 epochs with the Adam optimizer.
We trained the model on all $13$ categories of ShapeNet-Synthetic using the same training/test split of the original paper.

 \noindent  \textbf{SketchSampler.} 
This method employs as training dataset Synthetic-LineDrawing \cite{gao2022sketchsampler}, a paired sketch-3D dataset based on 3D models from ShapeNet. In our experiments, we used the official implementation, cited in the original paper \cite{gao2022sketchsampler}. In particular, we used the pre-trained model released by the authors, and pre-processed the input sketches to be in the same format of Synthetic-LineDrawing ones.

% \section{Global versus local comparision}

\begin{figure}
    \centering

    \includegraphics[width=0.125\textwidth]{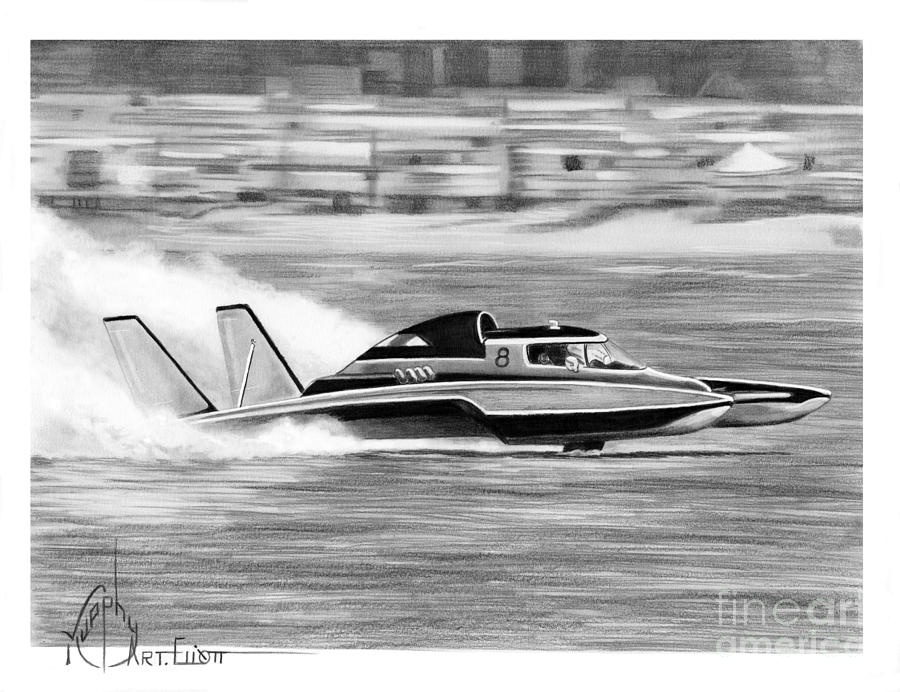}%
    \includegraphics[width=0.125\textwidth]{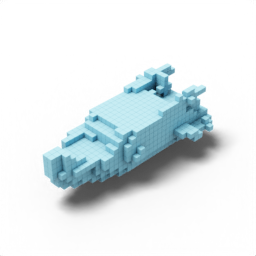}%
    \includegraphics[width=0.125\textwidth]{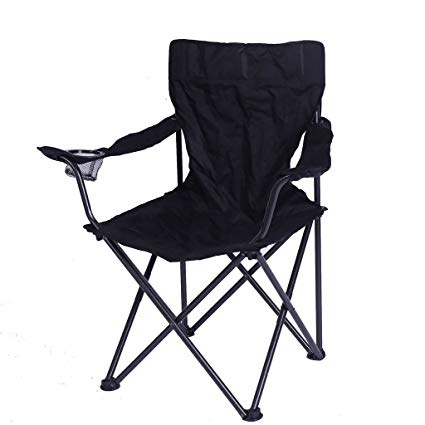}%
    \includegraphics[width=0.125\textwidth]{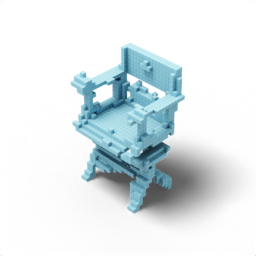}%

    \includegraphics[width=0.125\textwidth]{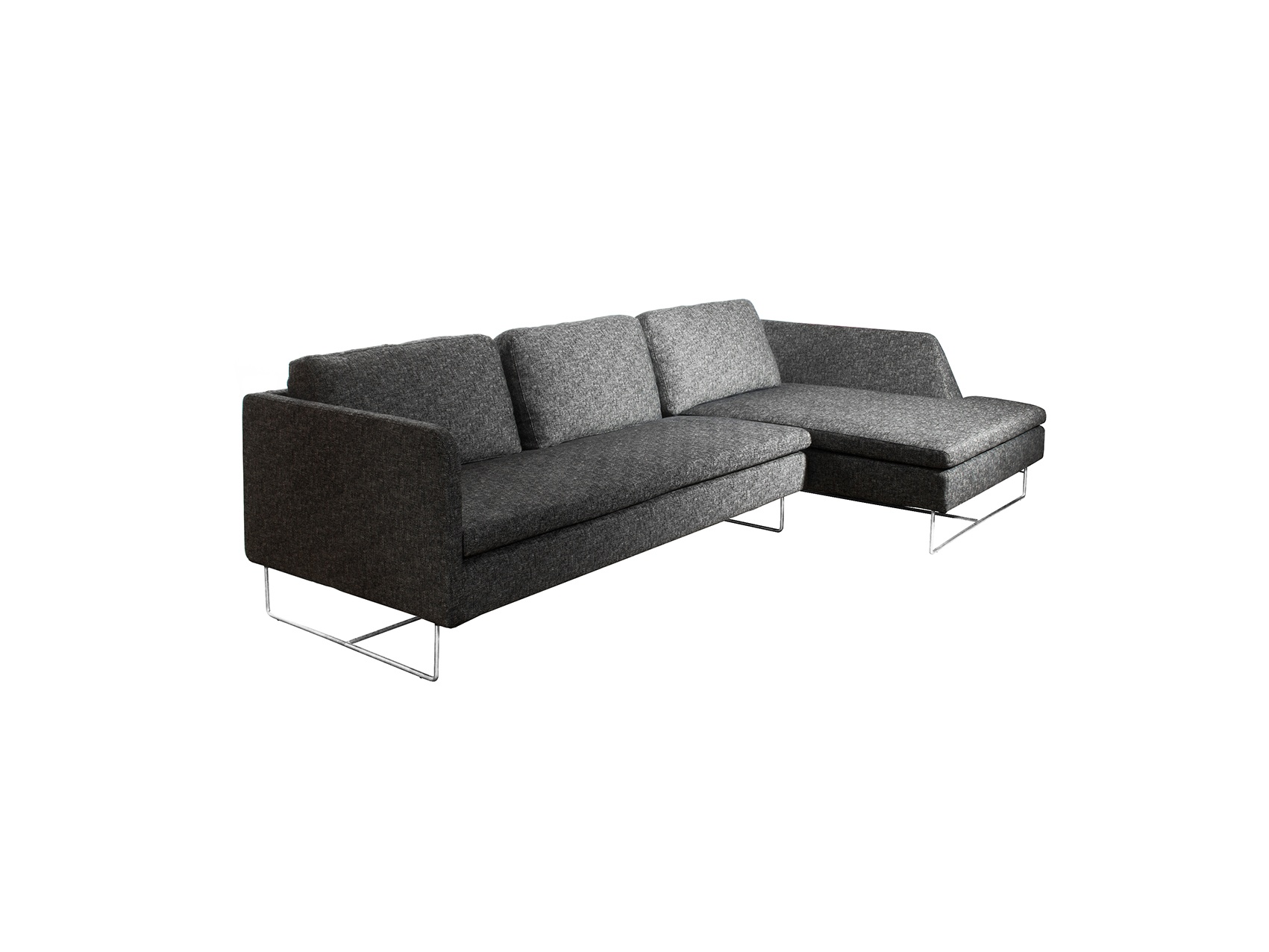}%
    \includegraphics[width=0.125\textwidth]{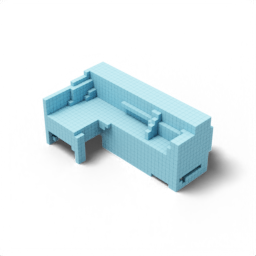}%
    \includegraphics[width=0.125\textwidth]{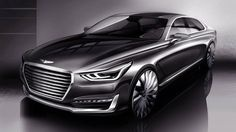}%
    \includegraphics[width=0.125\textwidth]{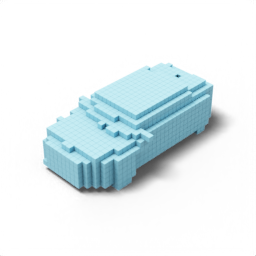}%

    \caption{Results of our method on natural images}

    \label{fig:Natural image}
\end{figure}

\section{Comparison with Point$\cdot$E}

Furthermore, we conducted a comparison between our work and Point$\cdot$E \cite{nichol2022pointea}, as illustrated in the table provided below (Row 1). The results clearly demonstrate the superior performance of our method, indicating the merit of our design choices.

\vspace{-8pt}
\begin{table}[h]
\centering
\resizebox{\linewidth}{!}{ %< auto-adjusts font size to fill line
\begin{tabular}{c|cccc}
\toprule
\textit{Method} & \textbf{QD-Acc$\uparrow$} & \textbf{TU-Acc$\uparrow$} & \textbf{SS-Acc$\uparrow$}  & \textbf{IS-Acc$\uparrow$}  \\
\midrule
Point$\cdot$E  & 12.6 & 40.1 & 43.2 & 18.9 \\
\textbf{Ours (CLIP)} & \textbf{58.8} & \textbf{81.5} & \textbf{79.7}  & \textbf{74.2} \\
\textbf{Ours (DINOv2)} & 39.7 & 71.1 & 72.1 & 55.9 \\

\bottomrule
\end{tabular}
} %< \resizebox
\vspace{-8pt}
% \caption{Parameter results}. % \caption
\label{tab:cf_comparison}
% \vspace{-8pt}
\end{table}

\section{Natural Images Results}
We explored the applicability of our method to natural images, as it is robust to domain shifts between renderings and sketches. The outcomes are depicted in Figure \ref{fig:Natural image}, indicating the efficacy of our method in generating natural images, including those with backgrounds. We believe that this discovery would be of interest to the Single View Reconstruction community.

\section{Failure Cases}
\begin{figure}
    \centering

    \includegraphics[width=0.125\textwidth]{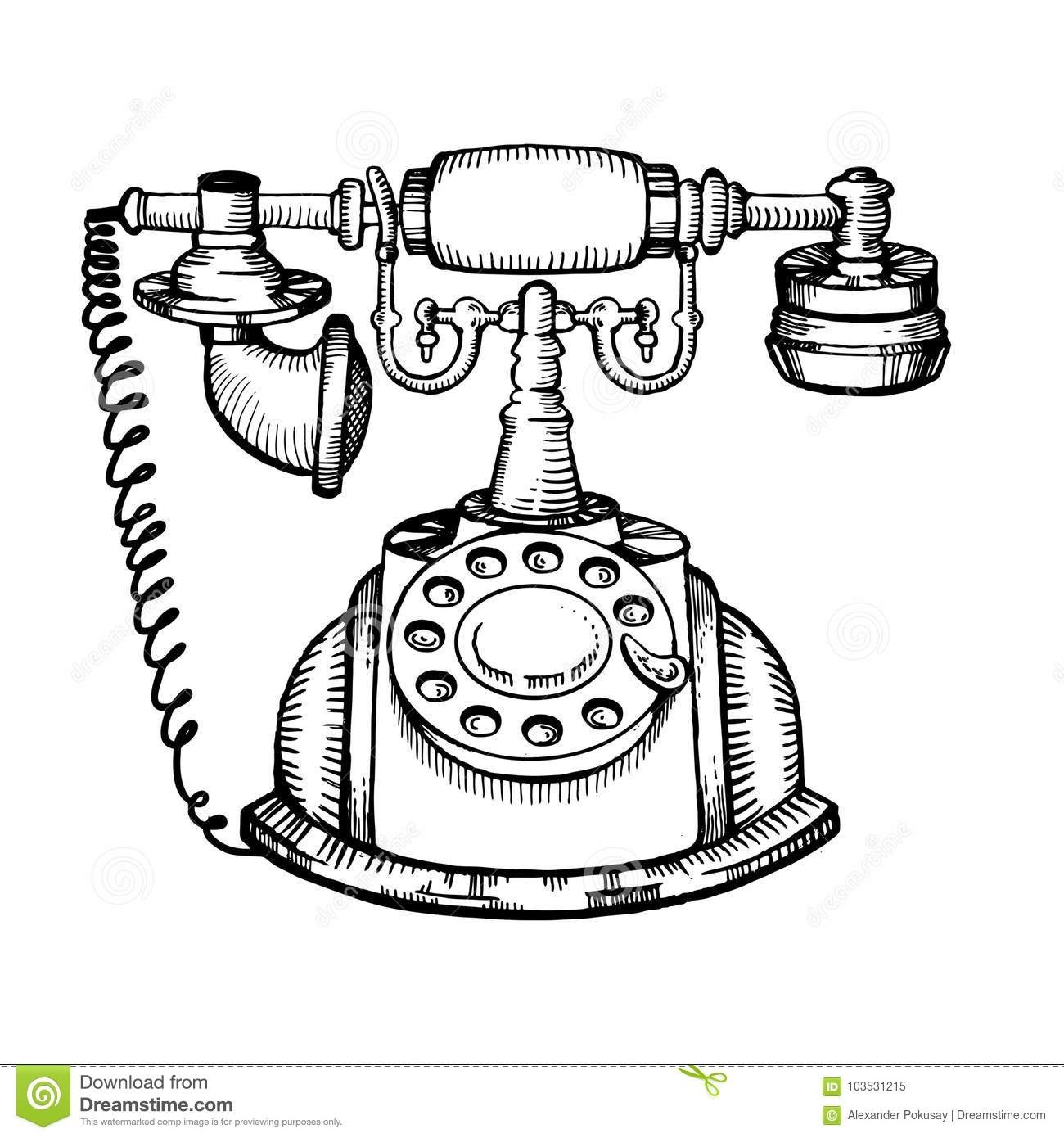}%
    \includegraphics[width=0.125\textwidth]{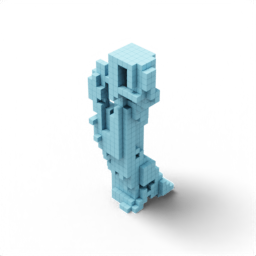}%
    \includegraphics[width=0.125\textwidth]{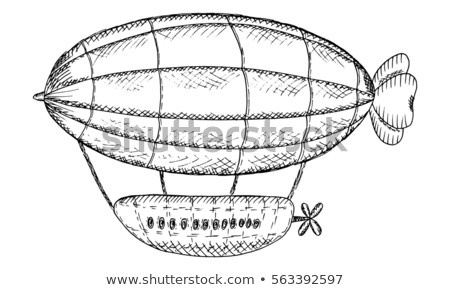}%
    \includegraphics[width=0.125\textwidth]{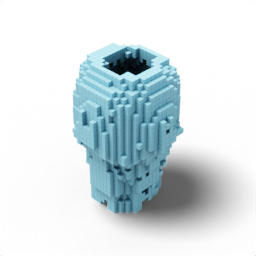}%

    \includegraphics[width=0.125\textwidth]{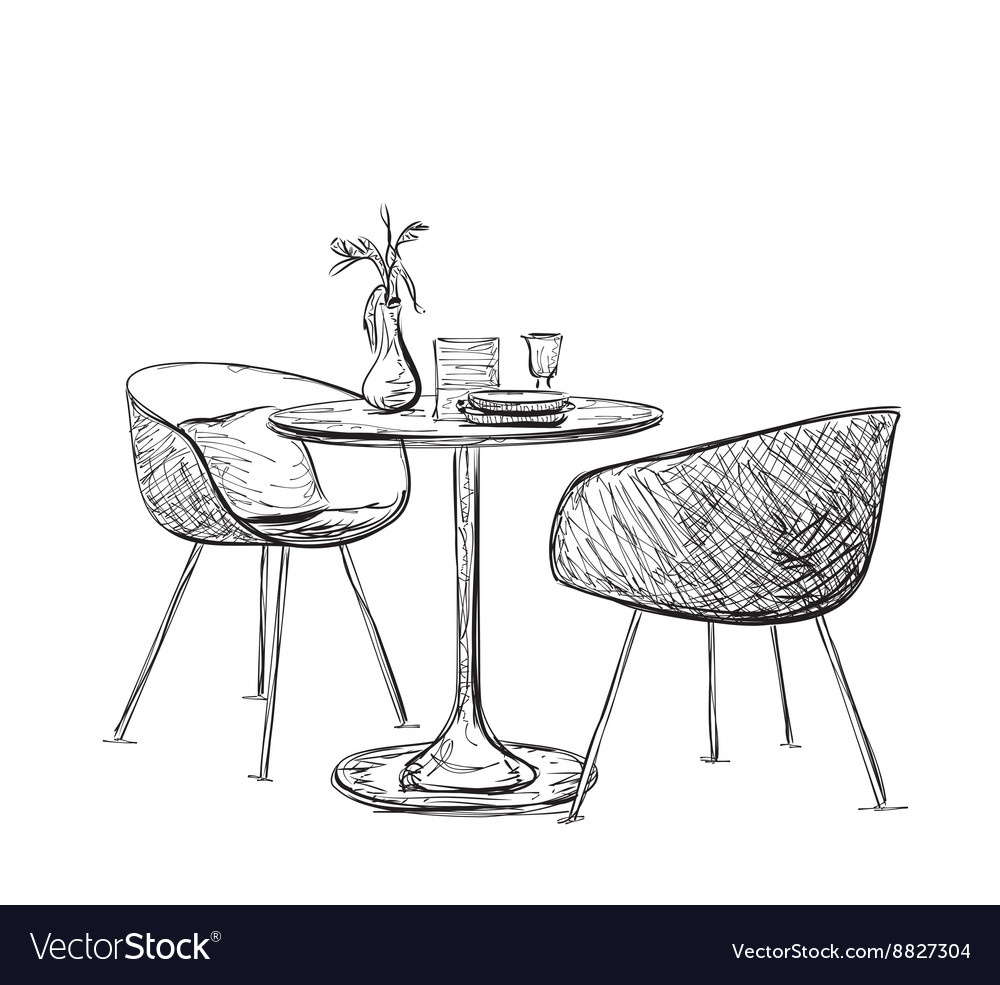}%
    \includegraphics[width=0.125\textwidth]{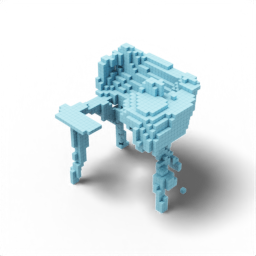}%
    \includegraphics[width=0.125\textwidth]{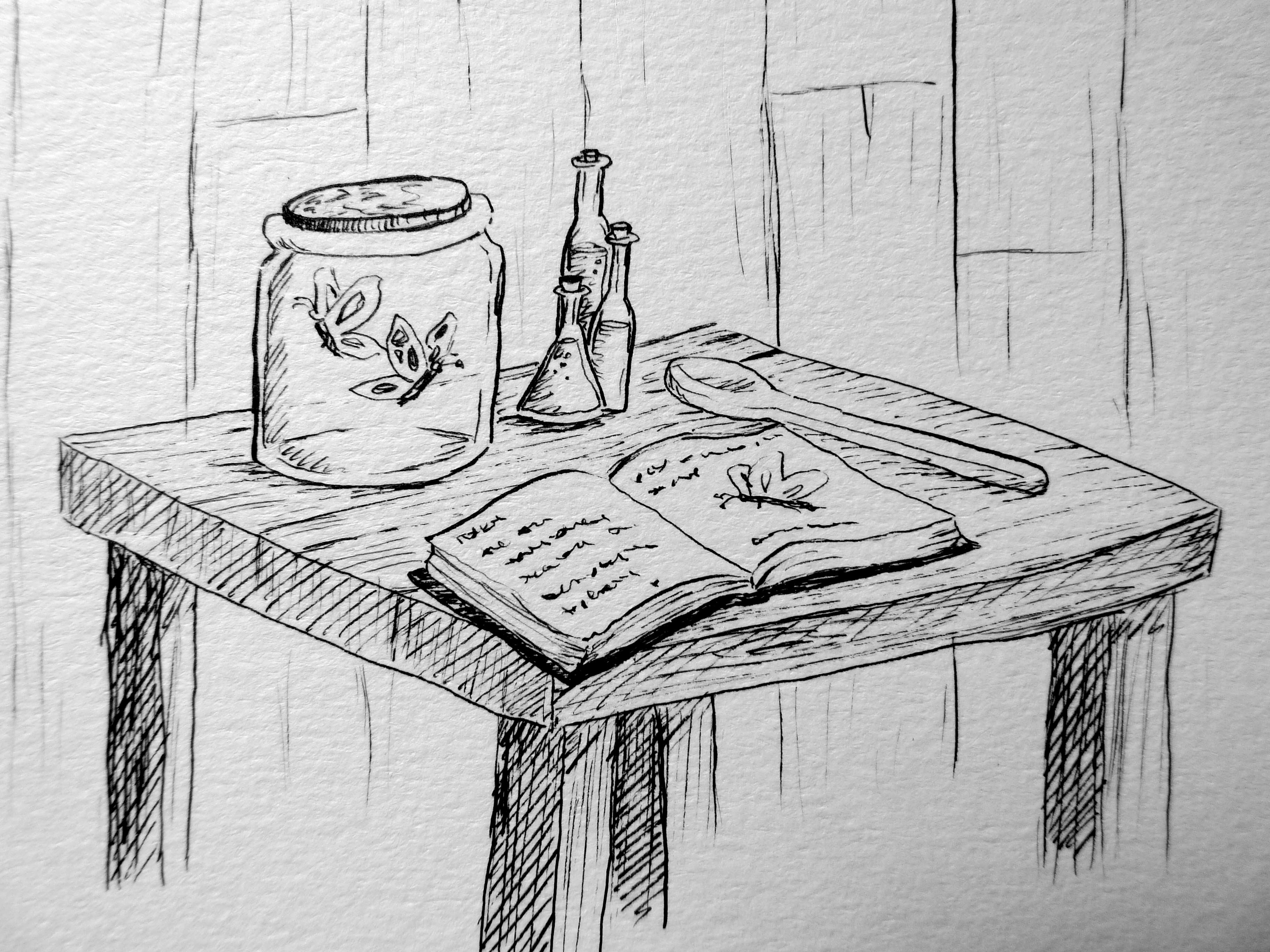}%
    \includegraphics[width=0.125\textwidth]{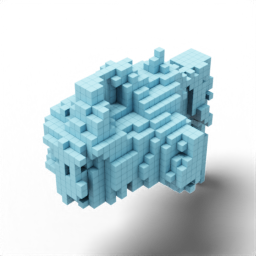}%

    \caption{Failure cases of our method}

    \label{fig:Failure}
\end{figure}

This section demonstrates the limitations of our method, as illustrated in Figure \ref{fig:Failure}. The outcomes reveal that our method encounters difficulties in generalizing to shapes that are beyond those present in ShapeNet13, as depicted in the first row. Furthermore, our method also faces challenges when dealing with sketches that depict multiple shapes, as shown in the second row. Lastly, our method experiences difficulties in accurately reproducing the local details of shapes, which we consider to be an intriguing direction for future work.

% \section{More Visual Results}

% \subsection{Multi-Results}

\section{Societal Impact}

The societal impact of Sketch-to-3D technology can be significant in various fields such as architecture, product design, gaming, and entertainment. With the help of Sketch-to-3D technology, designers and architects can create realistic 3D models quickly and efficiently, reducing the overall time and cost of the design process.
However, it is important to note that the widespread adoption of Sketch-to-3D technology could also lead to job displacement in certain industries. As with any technological advancement, it is crucial to consider the potential social and economic impacts and work towards ensuring a smooth transition for workers and communities affected by such changes.

\section{Future Work}
We aim to concentrate on expanding this method to handle bigger 3D datasets for our future work. Additionally, we think that enhancing the Stage 1 VQ-VAE can aid in preserving the local features of the 3D shape. Lastly, an intriguing avenue to explore would be to combine sketch with text conditioning, resulting in a more adaptable generative model.

\section{More Qualitative Results}

\begin{figure*}
\newcommand{\singlewidth}{0.24\textwidth}
\newcommand{\multiwidth}{0.49\textwidth}
    \includegraphics[width=0.12\textwidth]{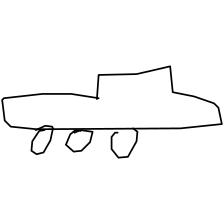}
    \includegraphics[width=0.12\textwidth]{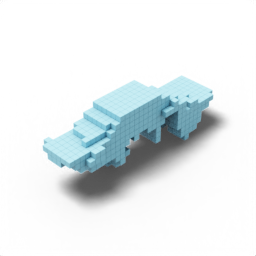}
    \includegraphics[width=0.12\textwidth]{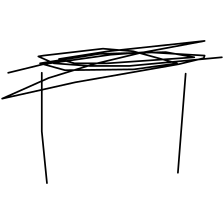}
    \includegraphics[width=0.12\textwidth]{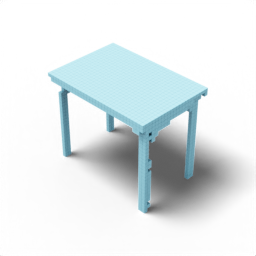}
    \includegraphics[width=0.12\textwidth]{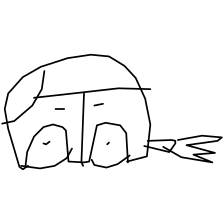}
    \includegraphics[width=0.12\textwidth]{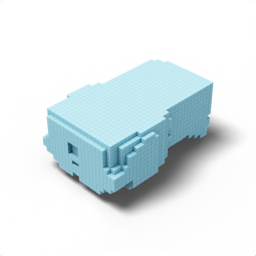}
    \includegraphics[width=0.12\textwidth]{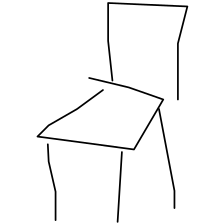}
    \includegraphics[width=0.12\textwidth]{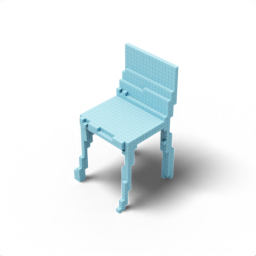}
    \\
    \includegraphics[width=0.12\textwidth]{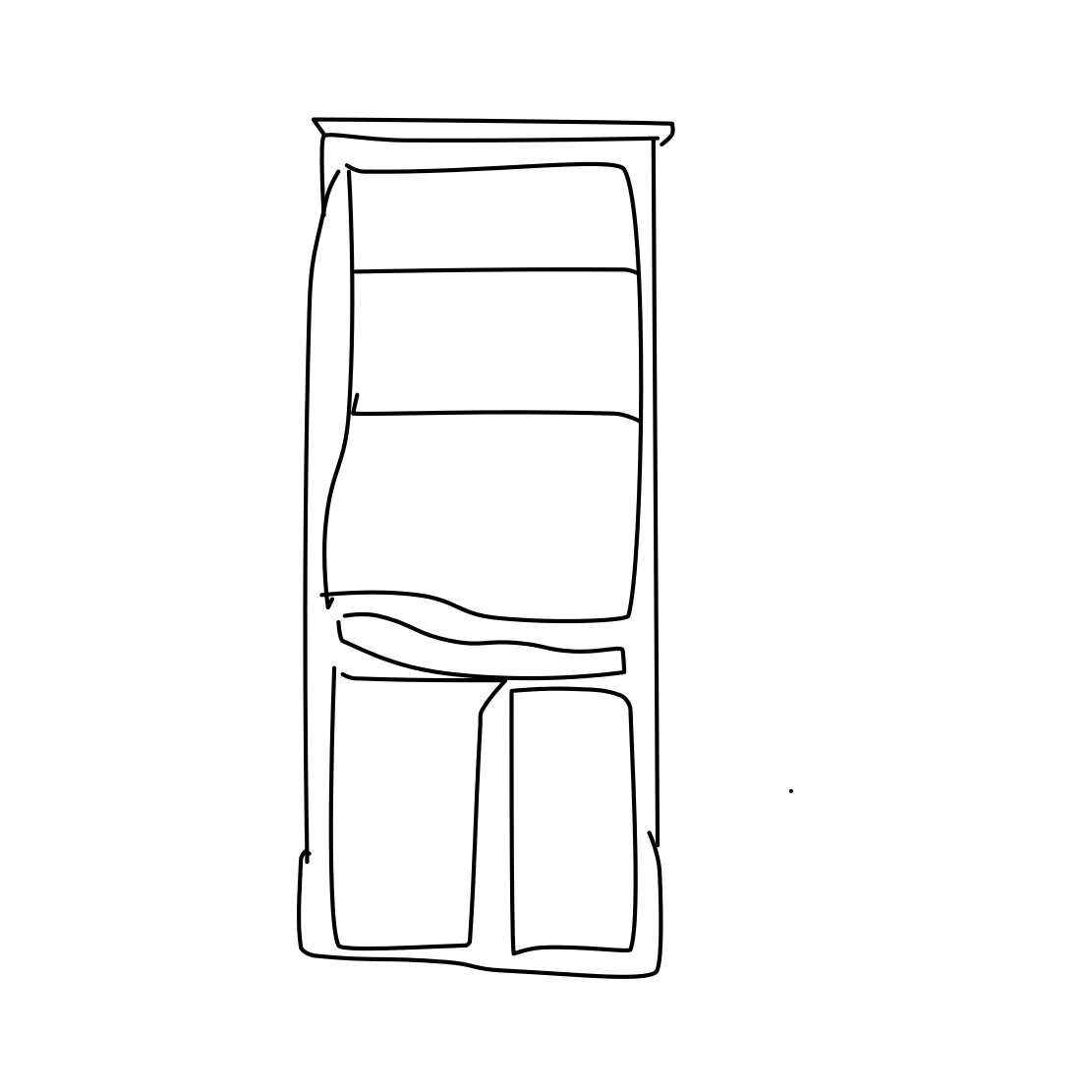}
    \includegraphics[width=0.12\textwidth]{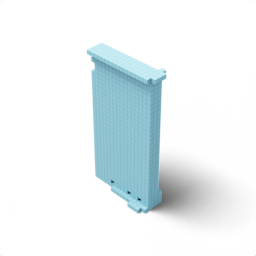}
    \includegraphics[width=0.12\textwidth]{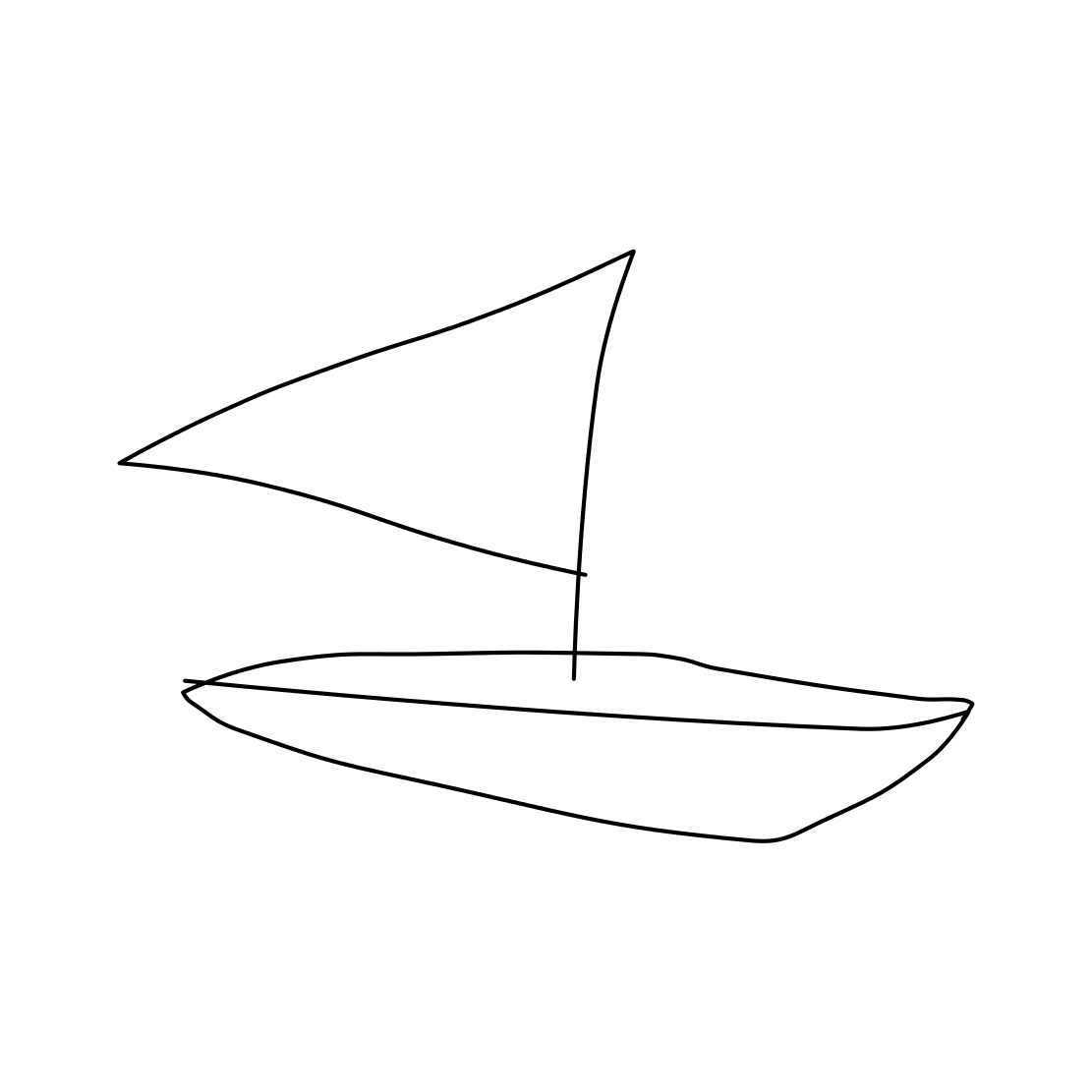}
    \includegraphics[width=0.12\textwidth]{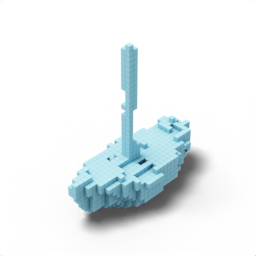}
    \includegraphics[width=0.12\textwidth]{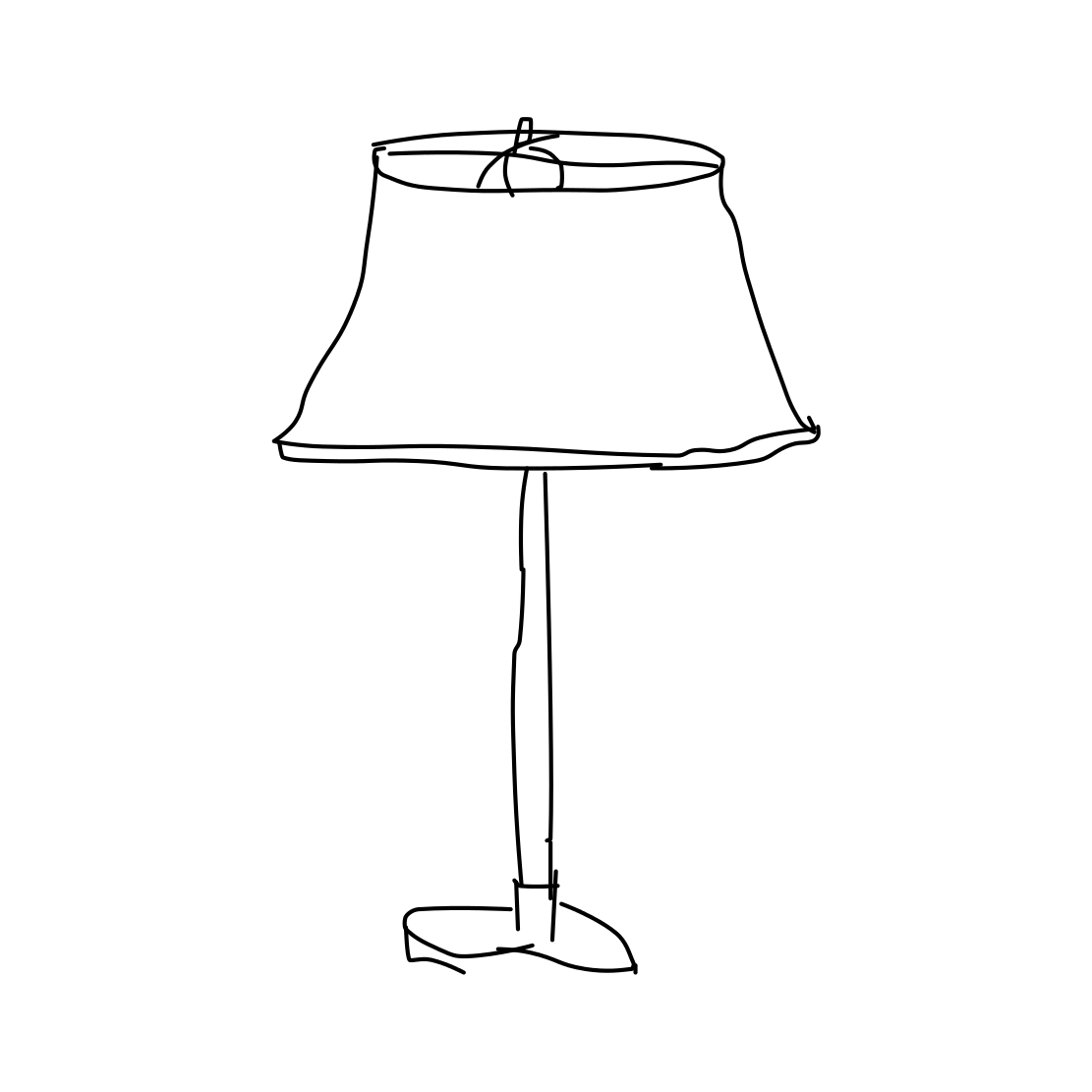}
    \includegraphics[width=0.12\textwidth]{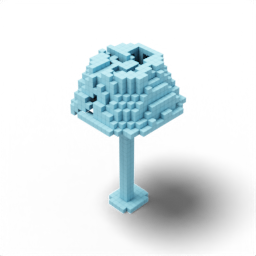}
    \includegraphics[width=0.12\textwidth]{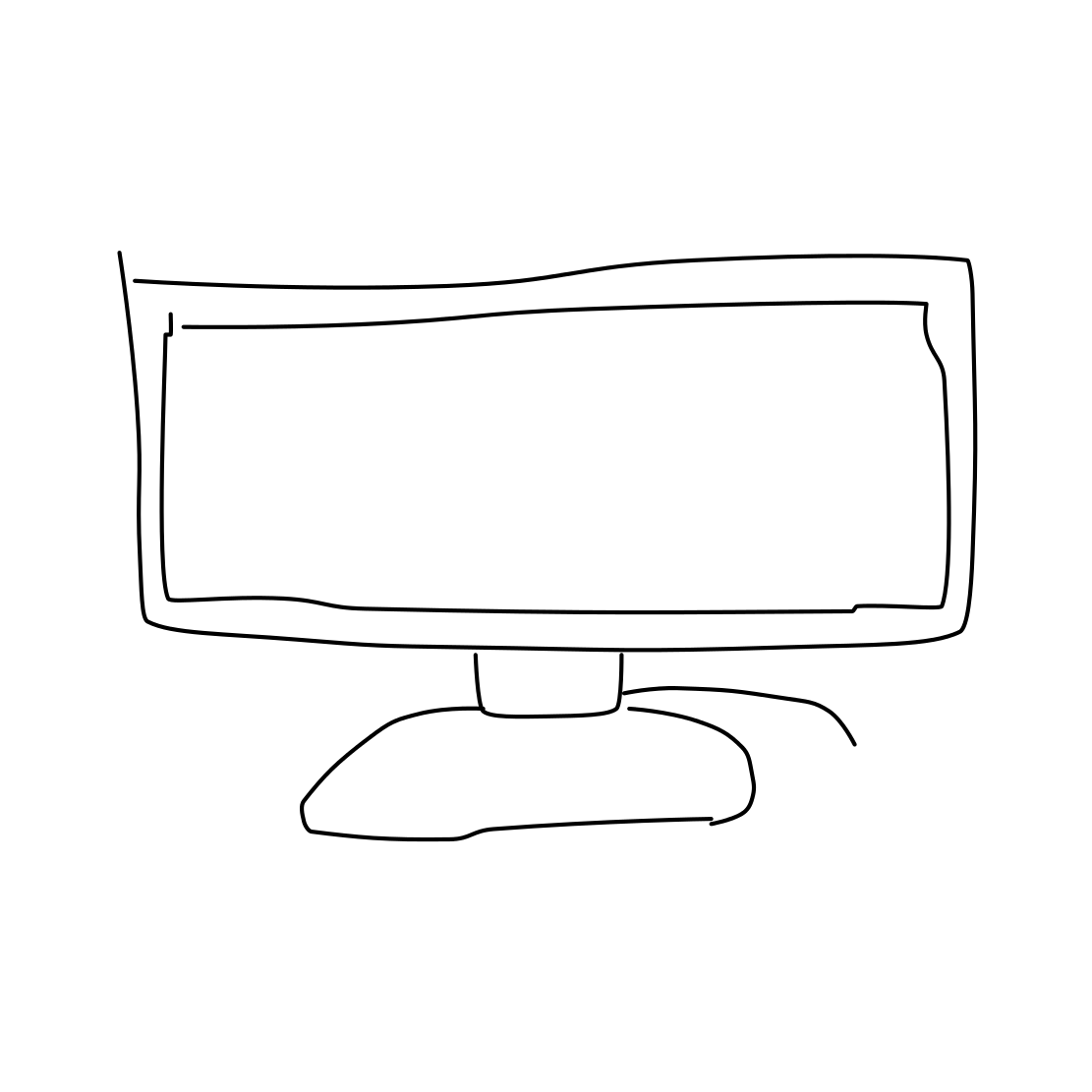}
    \includegraphics[width=0.12\textwidth]{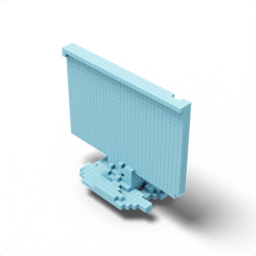}
    \\
    \includegraphics[width=0.12\textwidth]{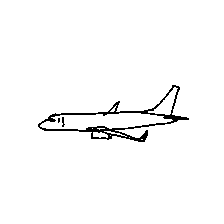}
    \includegraphics[width=0.12\textwidth]{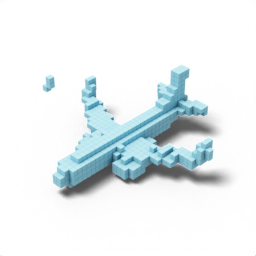}
    \includegraphics[width=0.12\textwidth]{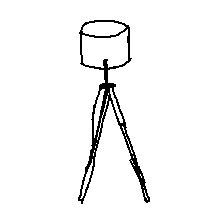}
    \includegraphics[width=0.12\textwidth]{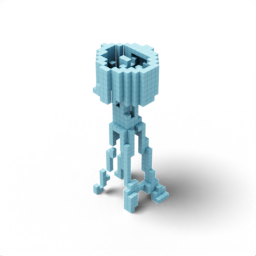}
    \includegraphics[width=0.12\textwidth]{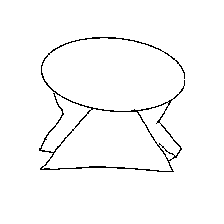}
    \includegraphics[width=0.12\textwidth]{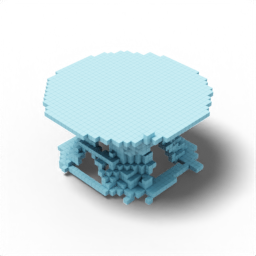} 
    \includegraphics[width=0.12\textwidth]{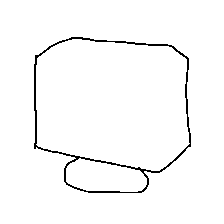}
    \includegraphics[width=0.12\textwidth]{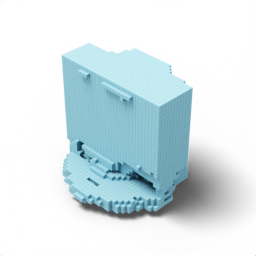}
    \\
    \includegraphics[width=0.12\textwidth]{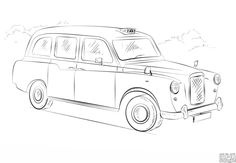}
    \includegraphics[width=0.12\textwidth]{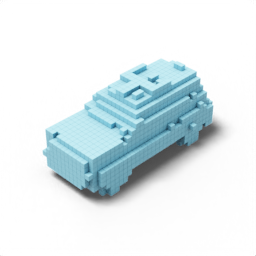}
    \includegraphics[width=0.12\textwidth]{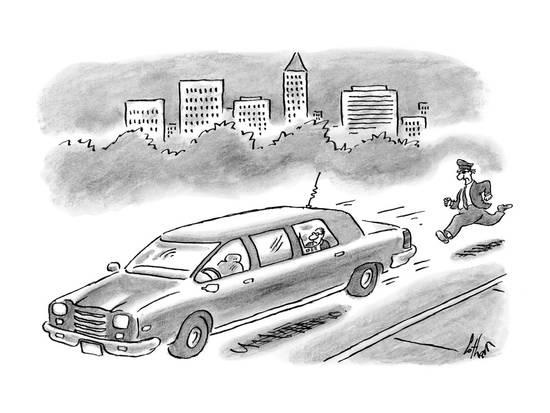}
    \includegraphics[width=0.12\textwidth]{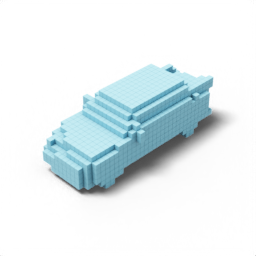}
    \includegraphics[width=0.12\textwidth]{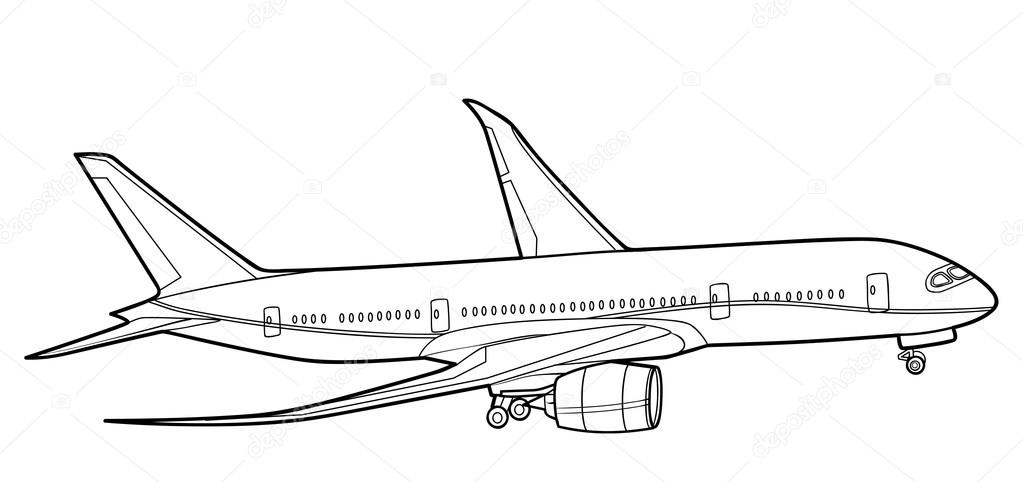}
    \includegraphics[width=0.12\textwidth]{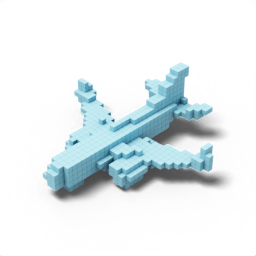}
    \includegraphics[width=0.12\textwidth]{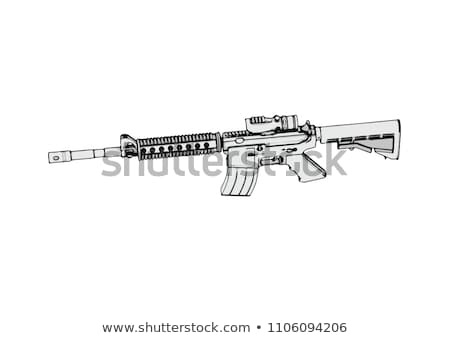}
    \includegraphics[width=0.12\textwidth]{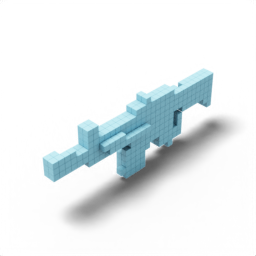}
    \\
      \includegraphics[width=0.12\textwidth]{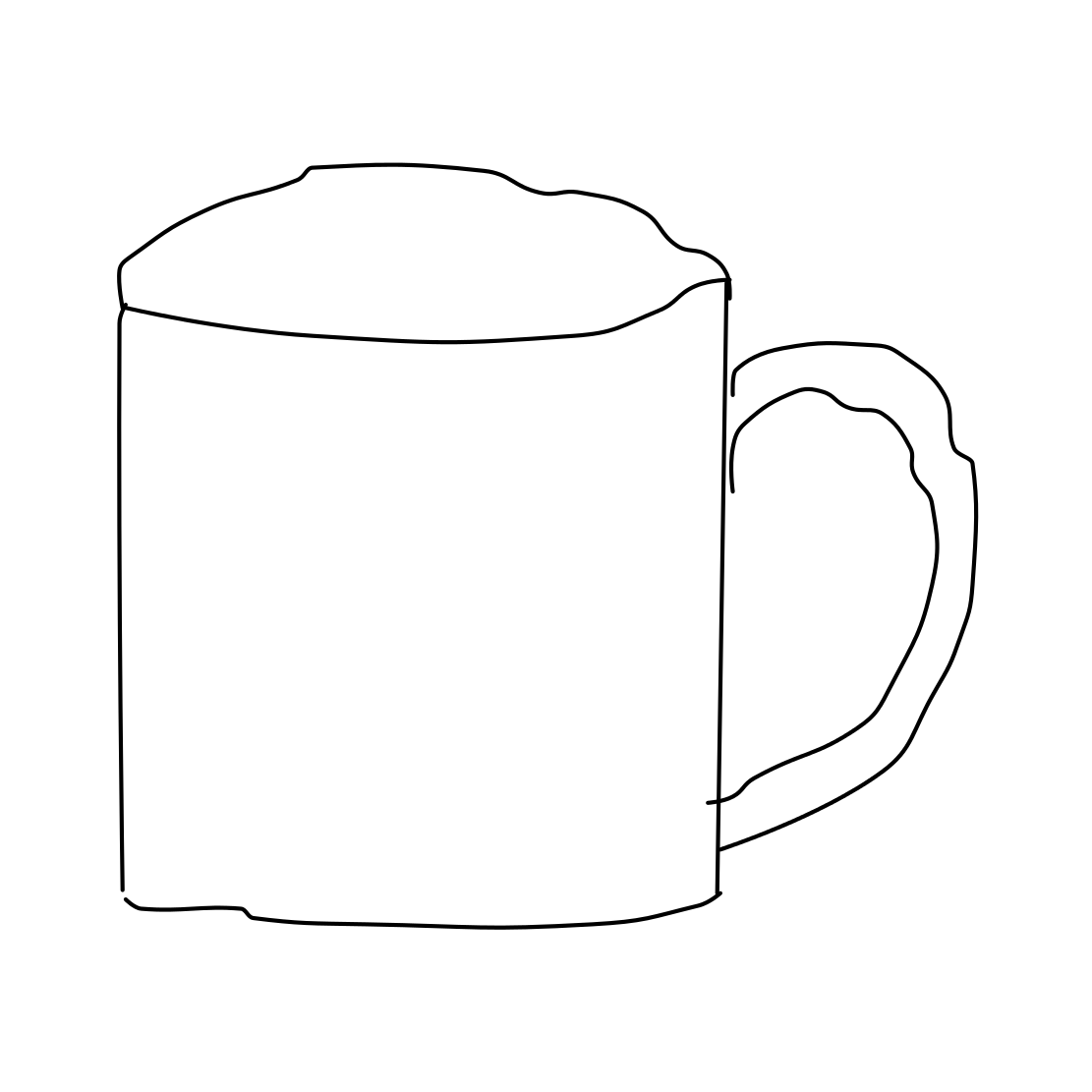}
    \includegraphics[width=0.12\textwidth]{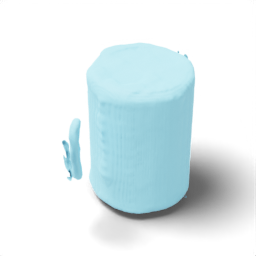}
    \includegraphics[width=0.12\textwidth]{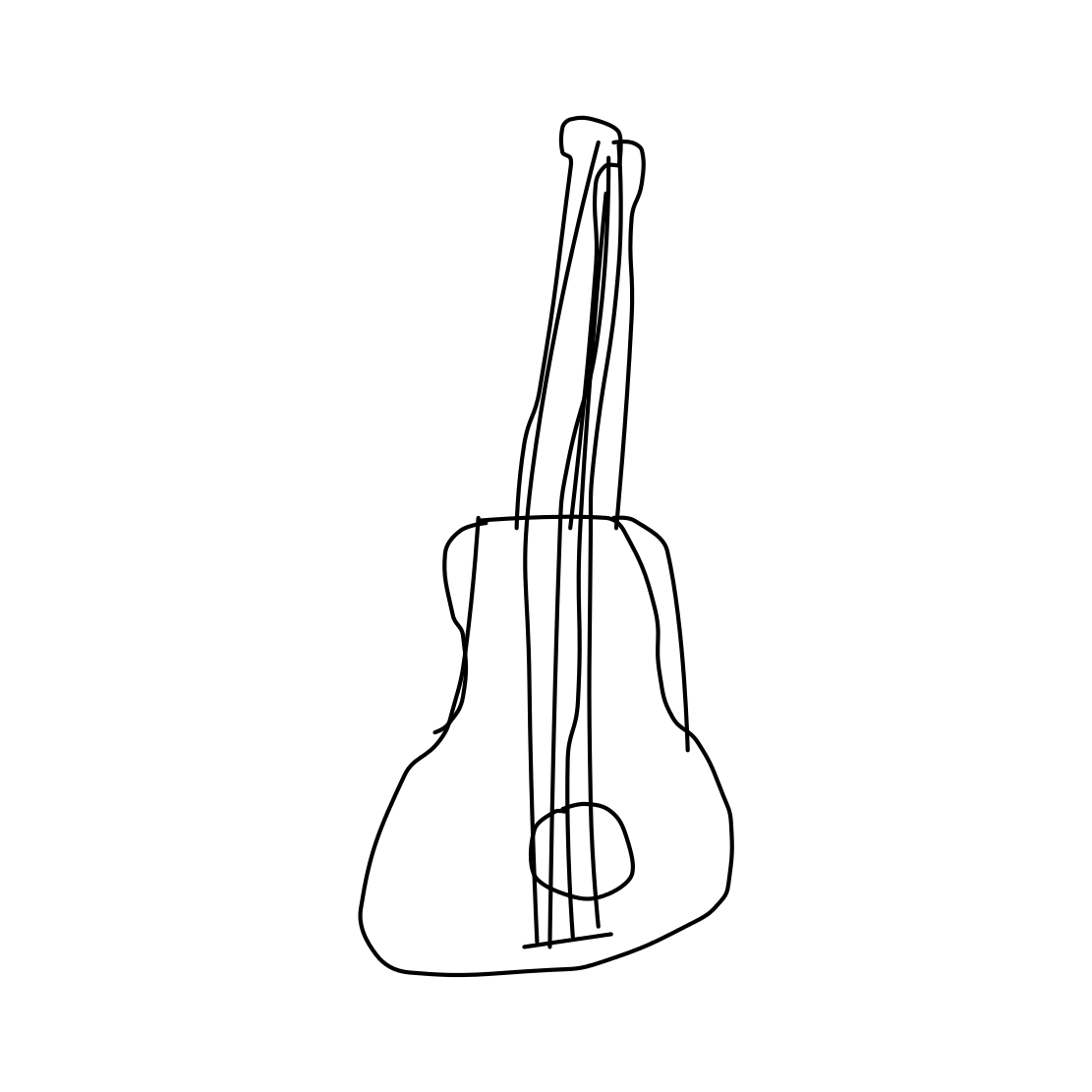}
    \includegraphics[width=0.12\textwidth]{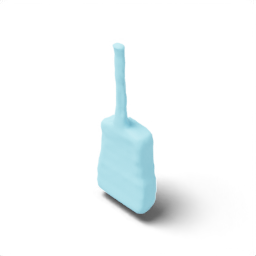}
    \includegraphics[width=0.12\textwidth]{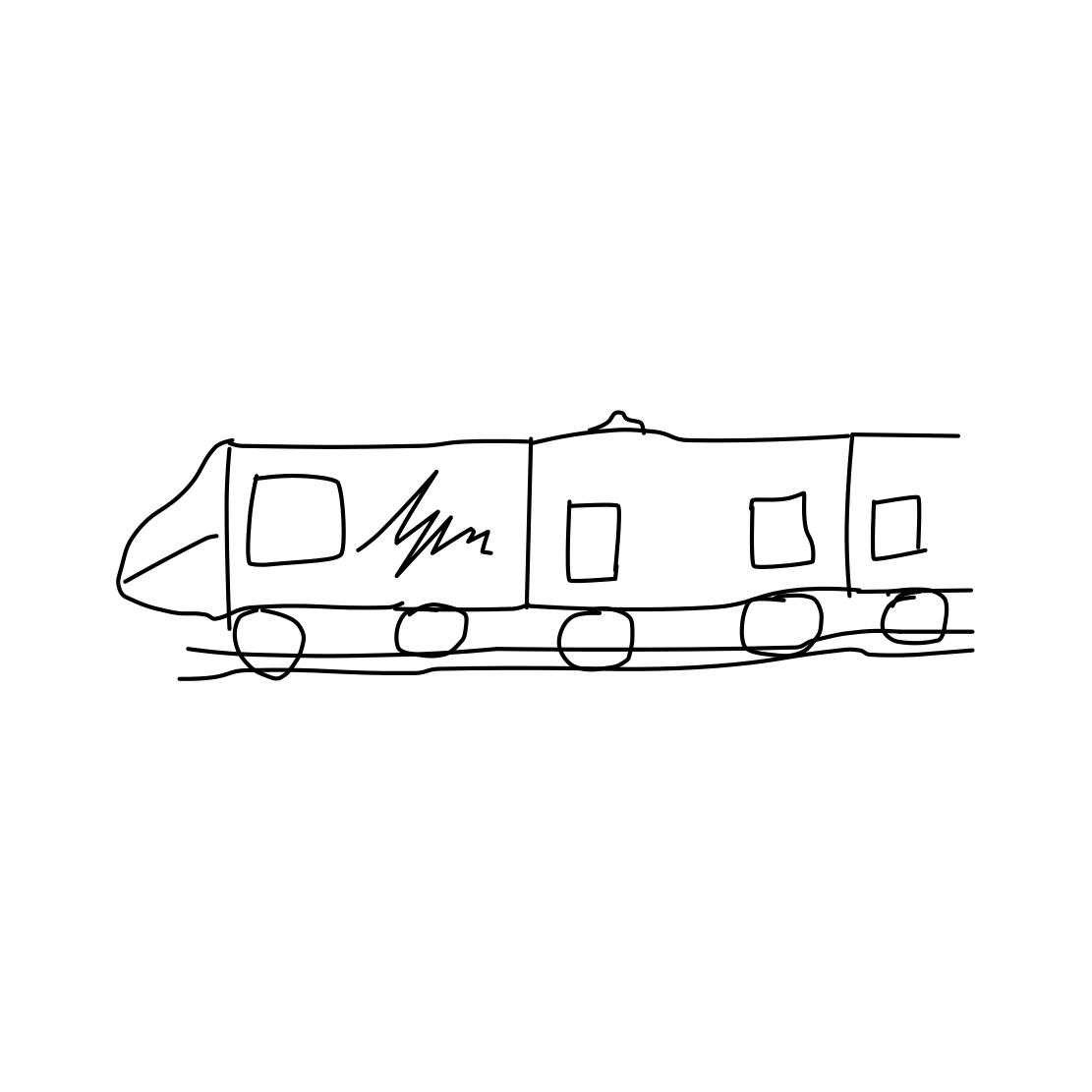}
    \includegraphics[width=0.12\textwidth]{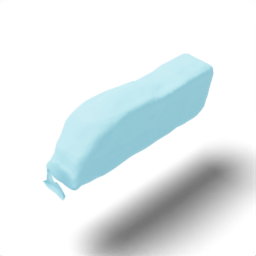}
    \includegraphics[width=0.12\textwidth]{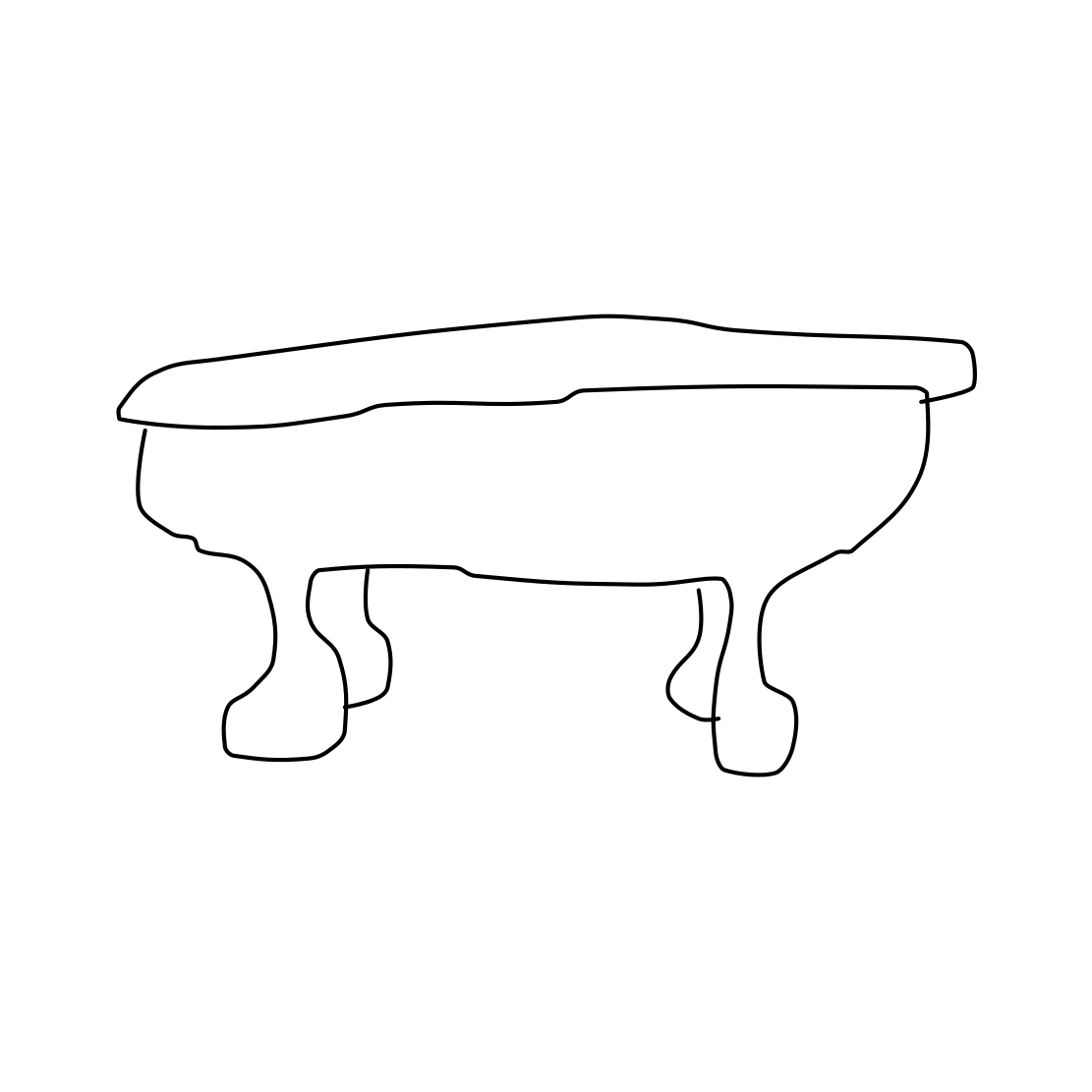}
    \includegraphics[width=0.12\textwidth]{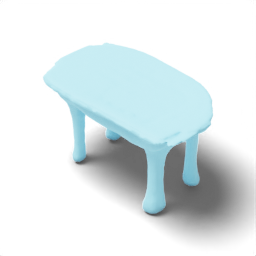}
    \\
    \includegraphics[width=0.12\textwidth]{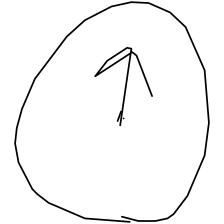}
    \includegraphics[width=0.12\textwidth]{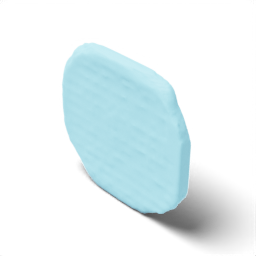}
    \includegraphics[width=0.12\textwidth]{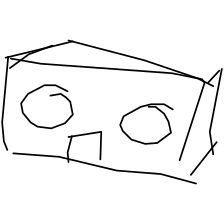}
    \includegraphics[width=0.12\textwidth]{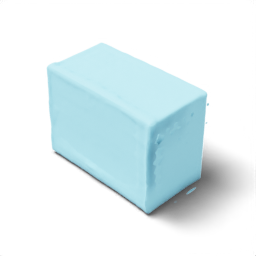}
    \includegraphics[width=0.12\textwidth]{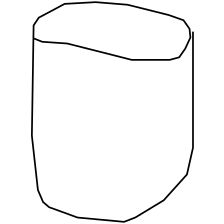}
    \includegraphics[width=0.12\textwidth]{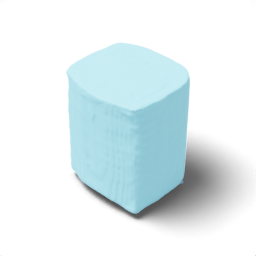}
    \includegraphics[width=0.12\textwidth]{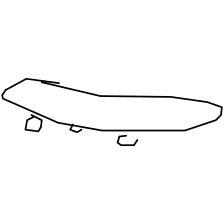}
    \includegraphics[width=0.12\textwidth]{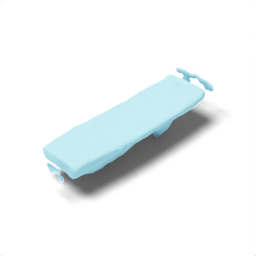}
    \\
    \includegraphics[width=0.12\textwidth]{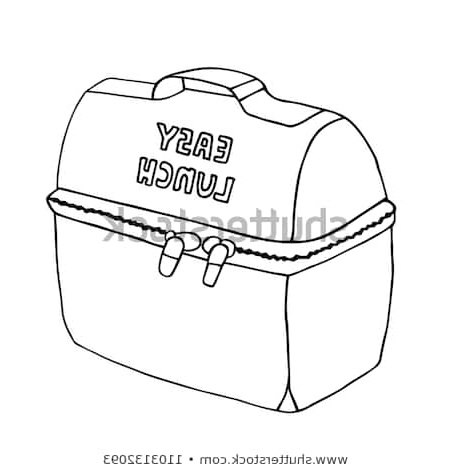}
    \includegraphics[width=0.12\textwidth]{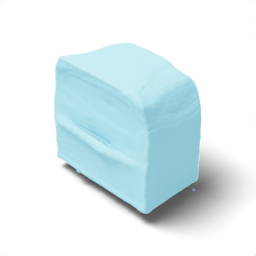}
    \includegraphics[width=0.12\textwidth]{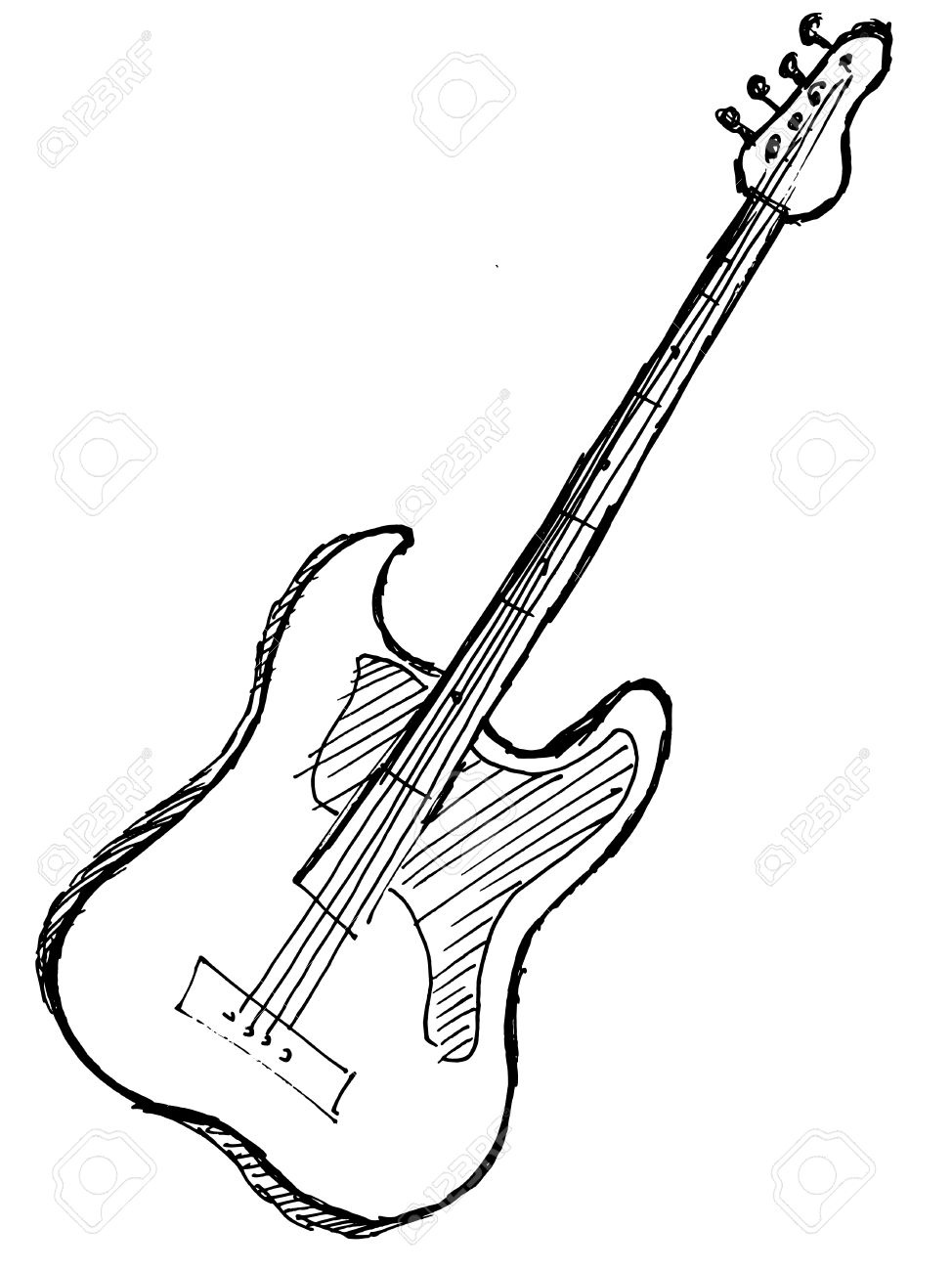}
    \includegraphics[width=0.12\textwidth]{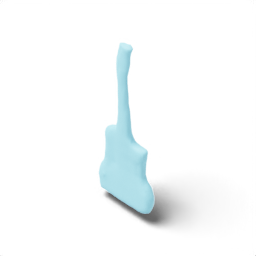}
    \includegraphics[width=0.12\textwidth]{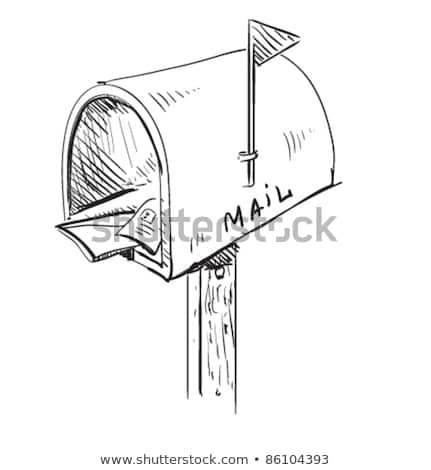}
    \includegraphics[width=0.12\textwidth]{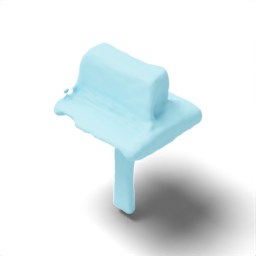}
    \includegraphics[width=0.12\textwidth]{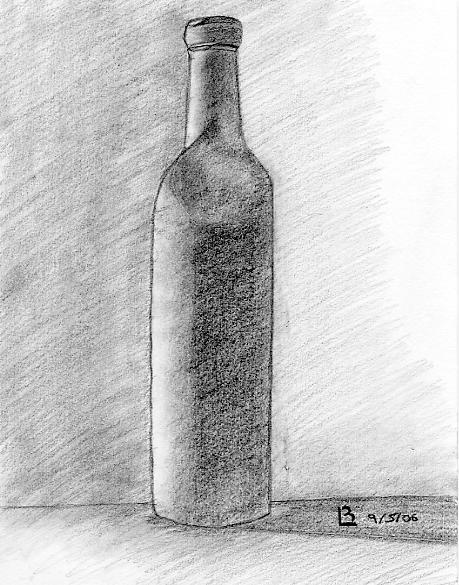}
    \includegraphics[width=0.12\textwidth]{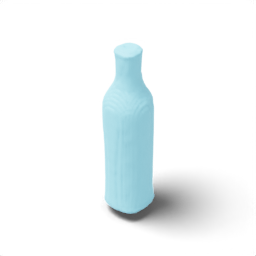}
    \\

    \includegraphics[width=0.12\textwidth]{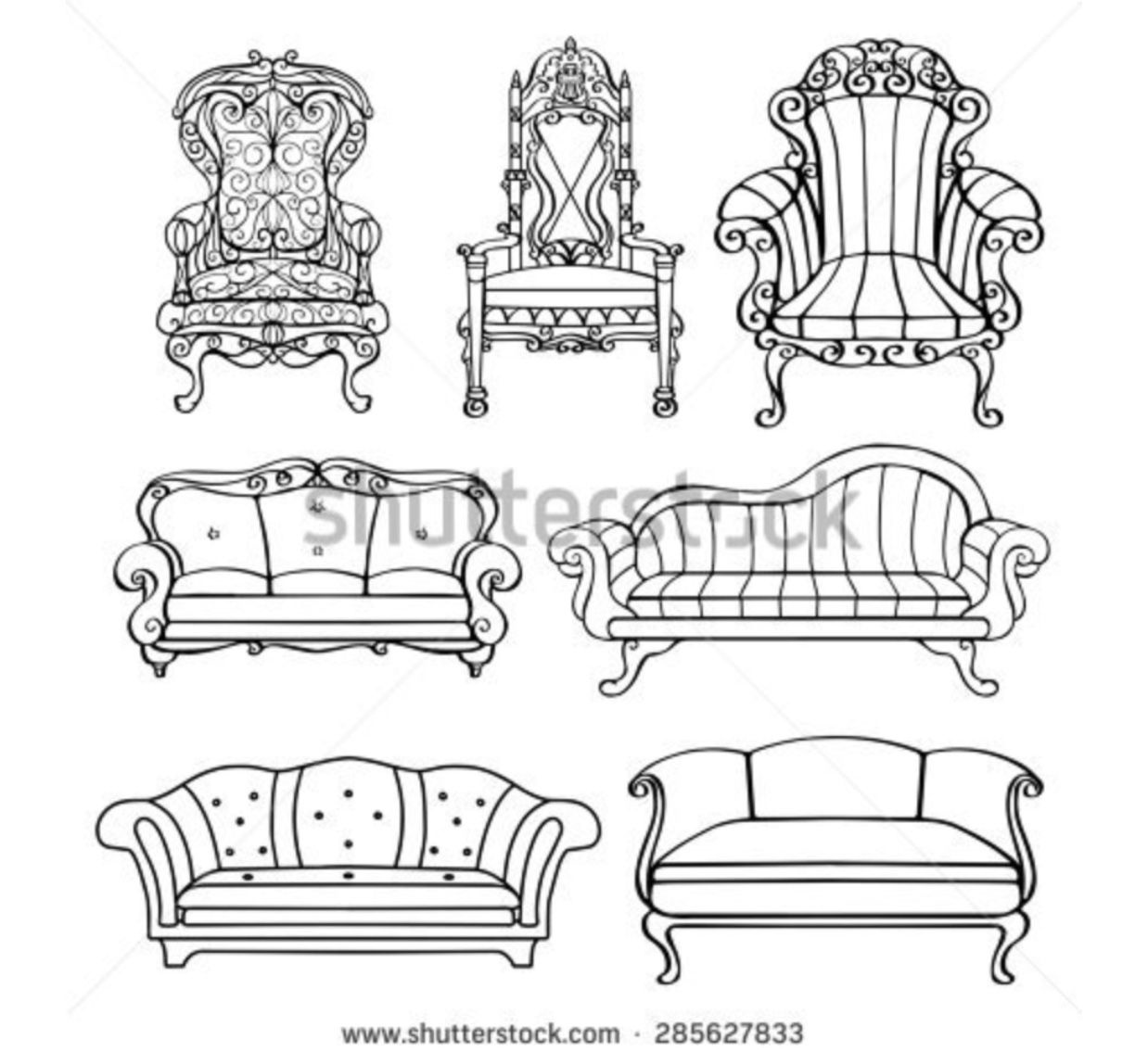}
    \includegraphics[width=0.12\textwidth]{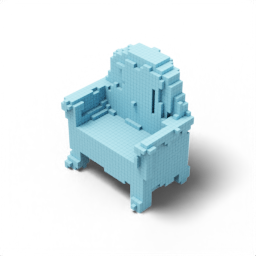}
    \includegraphics[width=0.12\textwidth]{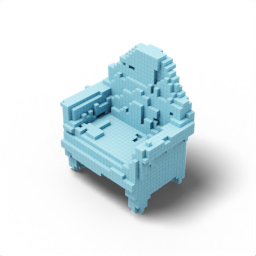}
    \includegraphics[width=0.12\textwidth]{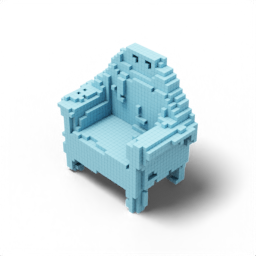}
    \includegraphics[width=0.12\textwidth]{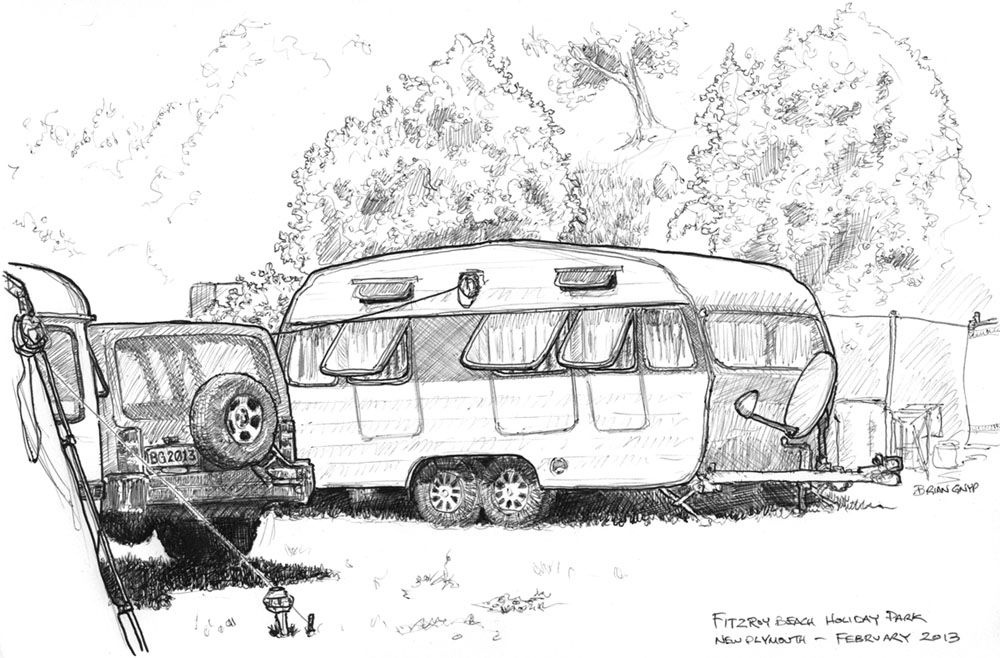}
    \includegraphics[width=0.12\textwidth]{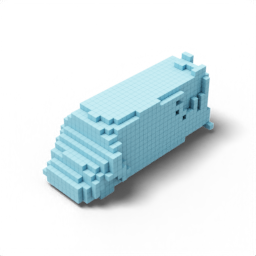}
    \includegraphics[width=0.12\textwidth]{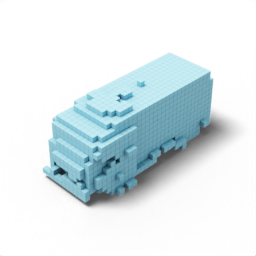}
    \includegraphics[width=0.12\textwidth]{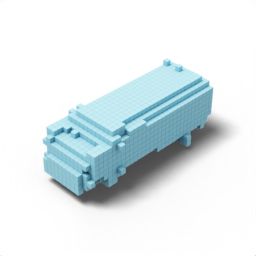}
    \\
    \includegraphics[width=0.12\textwidth]{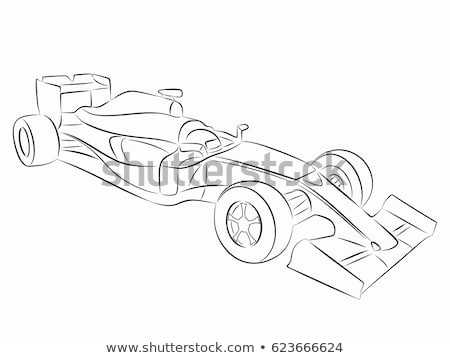}
    \includegraphics[width=0.12\textwidth]{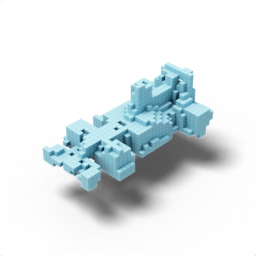}
    \includegraphics[width=0.12\textwidth]{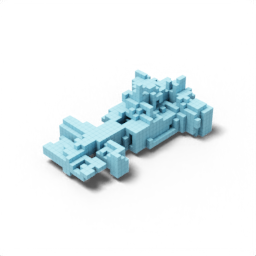}
    \includegraphics[width=0.12\textwidth]{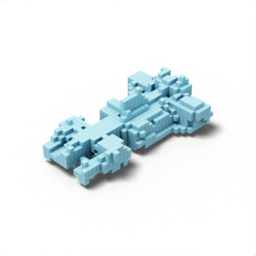}
    \includegraphics[width=0.12\textwidth]{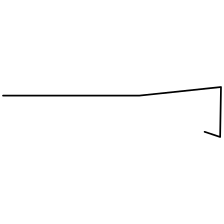}
    \includegraphics[width=0.12\textwidth]{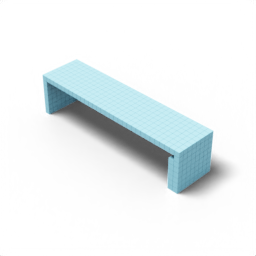}
    \includegraphics[width=0.12\textwidth]{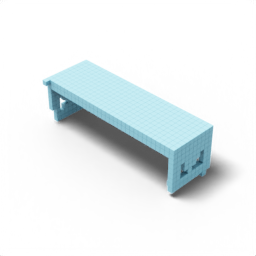}
    \includegraphics[width=0.12\textwidth]{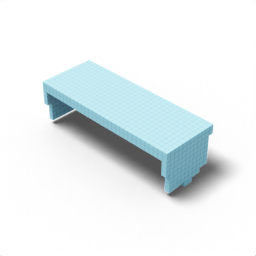}
    \\

    \caption{Additional outcomes of our $32^3$ voxel model on ShapeNet13 and implicit model on ShapeNet55 are presented. The initial row displays the outcomes of  $32^3$ voxel model  from QuickDraw, the second row from TU Berlin, the third row from ShapeNet Sketch, and the fourth row from Imagenet Sketch. The fifth row displays the outcomes of  implicit model from QuickDraw, the sixth row from TU Berlin and the seventh row from Imagenet Sketch. The last two rows exhibit several results obtained from a single sketch. }
    \label{fig:more_shapenet_results}
\end{figure*}

\begin{figure*}
\newcommand{\singlewidth}{0.24\textwidth}
\newcommand{\multiwidth}{0.49\textwidth}
    \includegraphics[width=\singlewidth]{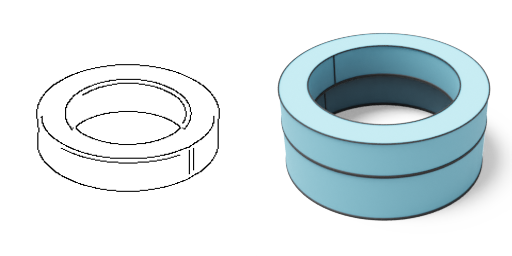}
    \includegraphics[width=\singlewidth]{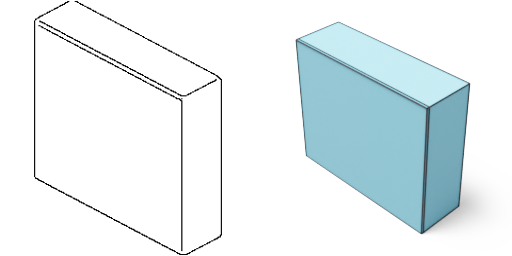}
    \includegraphics[width=\singlewidth]{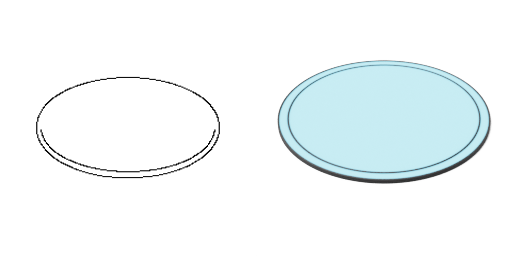}
    \includegraphics[width=\singlewidth]{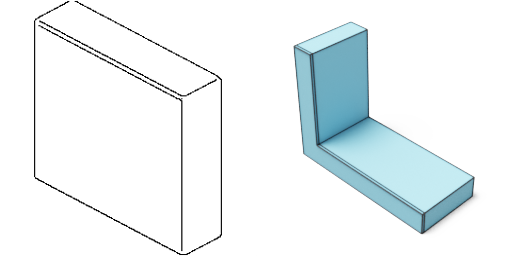}\\
    \includegraphics[width=\singlewidth]{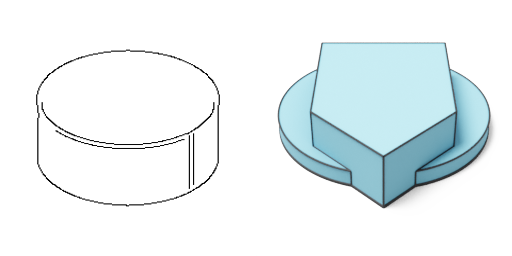}
    \includegraphics[width=\singlewidth]{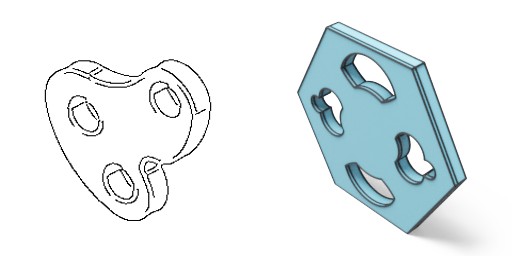}
    \includegraphics[width=\singlewidth]{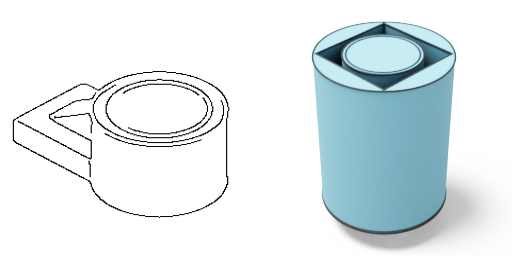}
    \includegraphics[width=\singlewidth]{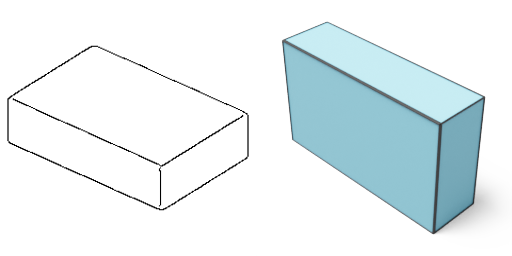}\\
    \includegraphics[width=\singlewidth]{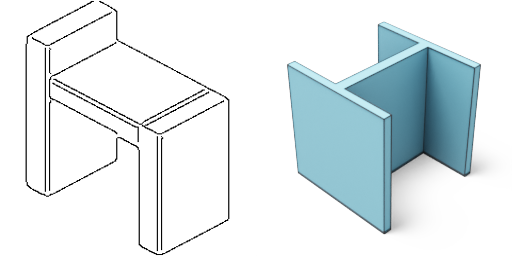}
    \includegraphics[width=\singlewidth]{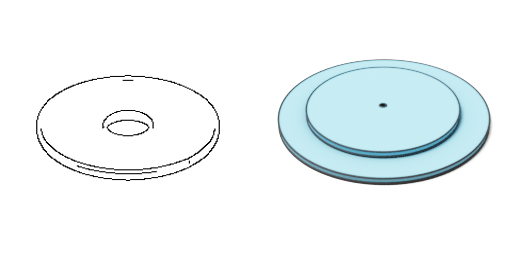}
    \includegraphics[width=\singlewidth]{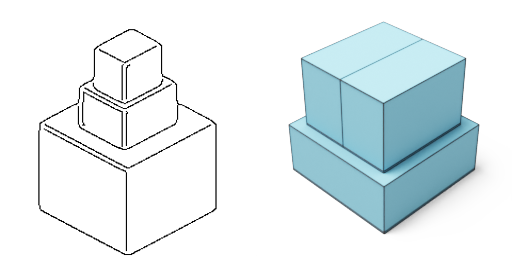}
    \includegraphics[width=\singlewidth]{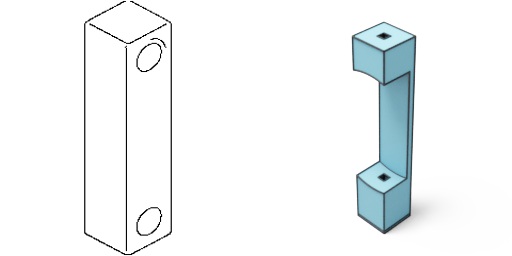}\\
    \includegraphics[width=\singlewidth]{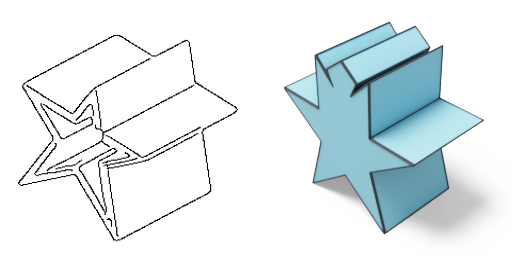}
    \includegraphics[width=\singlewidth]{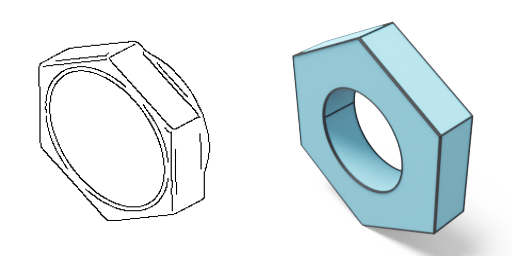}
    \includegraphics[width=\singlewidth]{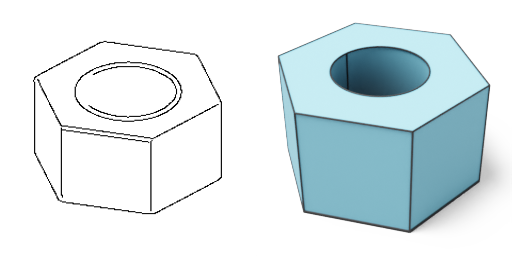}
    \includegraphics[width=\singlewidth]{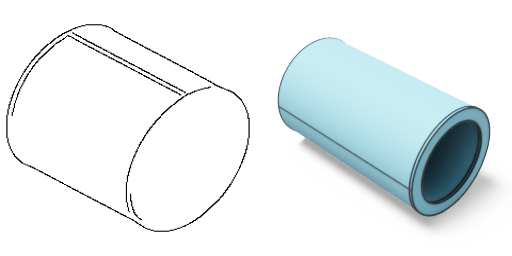}\\
    \includegraphics[width=\multiwidth]{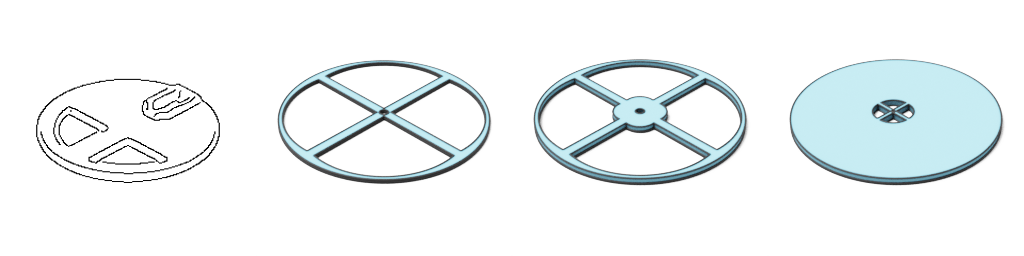}
    \includegraphics[width=\multiwidth]{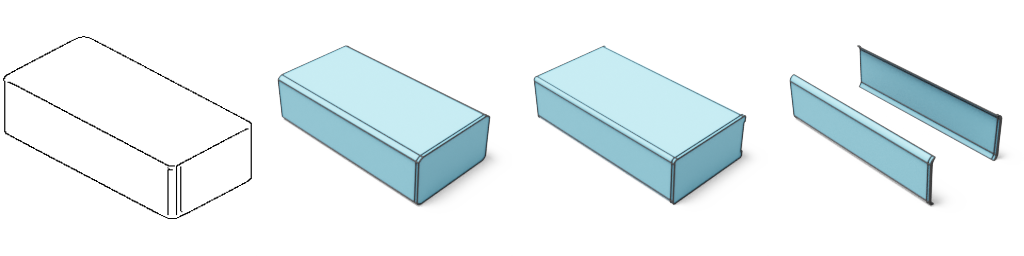}\\
    \includegraphics[width=\multiwidth]{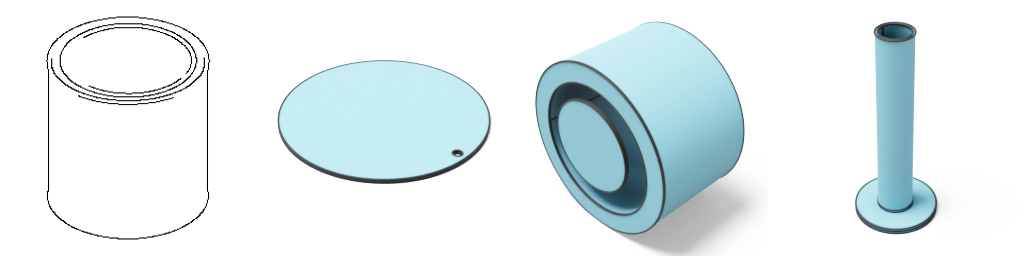}
    \includegraphics[width=\multiwidth]{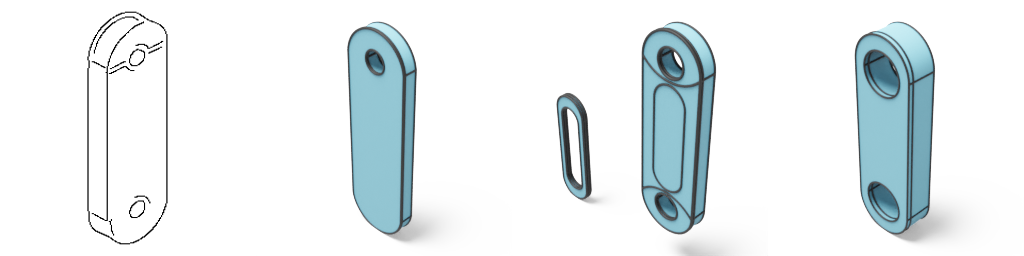}\\
    \includegraphics[width=\multiwidth]{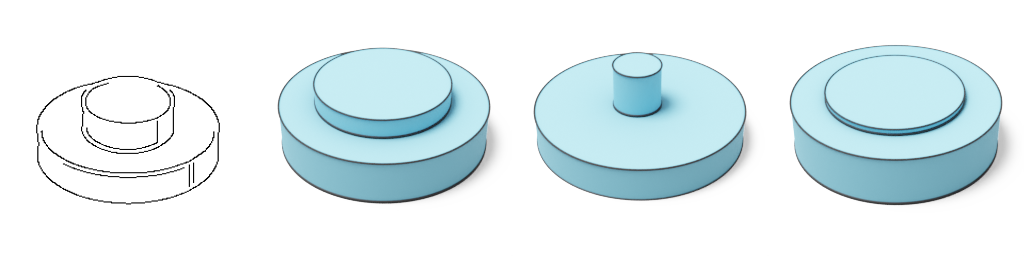}
    \includegraphics[width=\multiwidth]{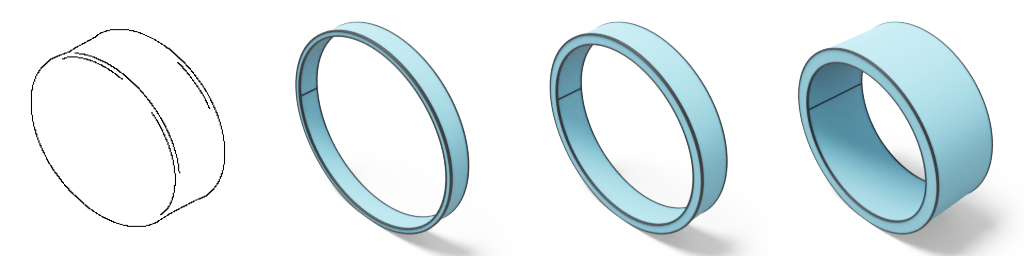}\\
    \includegraphics[width=\multiwidth]{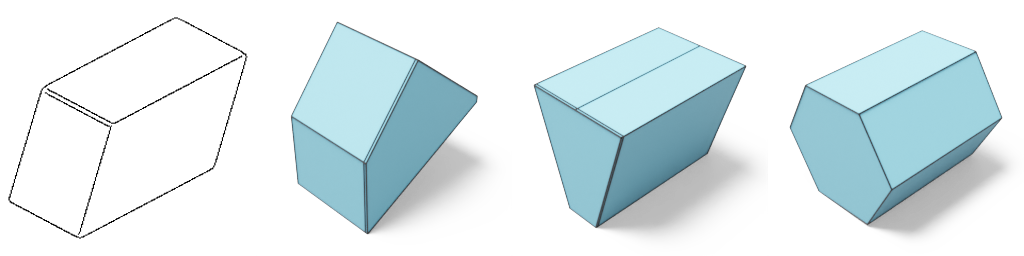}
    \includegraphics[width=\multiwidth]{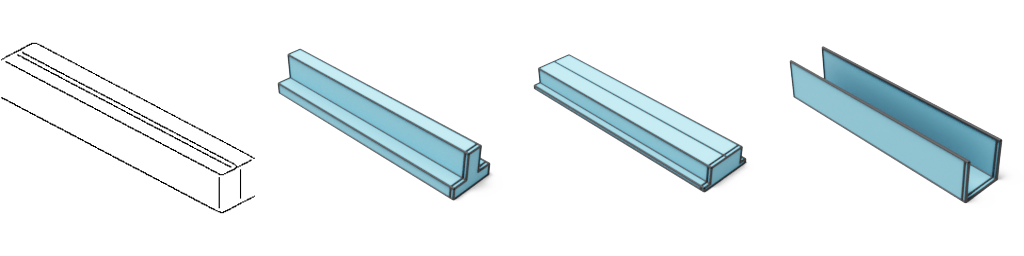}\\
    \caption{More results from our sketch-CAD model. The last four rows show multiple generation results for each sketch.}
    \label{fig:more_cad_results}
\end{figure*}

Additional results are provided in Figure \ref{fig:more_shapenet_results} and Figure \ref{fig:more_cad_results}.  

\end{document}